\documentclass[a4paper,10pt]{article}
\usepackage[utf8x]{inputenc}

\usepackage{graphicx}
\usepackage[colorlinks=true]{hyperref}
\hypersetup{urlcolor=blue, citecolor=red}
\usepackage{epstopdf} % needed if you have eps figures in a pdflatex manuscript
\usepackage{amsmath,amssymb,nicefrac}
\usepackage{algorithmicx}
\usepackage{subfigure}
\usepackage{authblk}
\usepackage{algorithm}% http://ctan.org/pkg/algorithms
\usepackage{algpseudocode}% http://ctan.org/pkg/algorithmicx
\newtheorem{mydef}{Definition}
\def \vec {\overrightarrow}
\def \Dset {\mathsf{D}}
\def \Gset {\mathsf{G}}
\newtheorem{theorem}{Theorem}

\graphicspath{{figures/}}

\title{A Tensor-Based Dictionary Learning Approach to \\
Tomographic Image Reconstruction}

\author[1]{Sara Soltani \thanks{ssol@dtu.dk}}
\author[2]{Misha E.~Kilmer \thanks{misha.kilmer@tufts.edu}}
\author[1]{Per Christian Hansen \thanks{pcha@dtu.dk}}
\affil[1]{Department of Applied Mathematics and Computer Science,
Technical University of Denmark,\authorcr DK-2800 Kgs. Lyngby, Denmark.}
\affil[2]{Department of Mathematics, Tufts University, \authorcr Medford, MA 02155, USA.}

\begin{document}

\maketitle
\let\thefootnote\relax\footnote{This work is part of the project HD-Tomo funded by Advanced Grant
No.\ 291405 from the European Research Council. Kilmer acknowledges support from NSF 1319653.}
\let\thefootnote\relax\footnote{\emph{2010 Mathematics Subject Classiffication}:\,Primary: 15A69; 
  Secondary: 65F22, 65K10.} % Optimization and variational techniques.}
\let\thefootnote\relax
 % 65F22	Ill-posedness, regularization
 % 15A69  	Multi-linear algebra, tensor products
 % 65K10  	Optimization and variational techniques

\begin{abstract}
We consider tomographic reconstruction using priors in the form of a
dictionary learned from training images.
The reconstruction has two stages:\
first we construct a tensor dictionary prior from
our training data, and then we pose the reconstruction problem
in terms of recovering the expansion coefficients in that dictionary.
Our approach differs from past approaches in that
a) we use a third-order tensor representation for our images
%rather than a more standard vectorized representation to define the dictionary priors,
and b) we recast the reconstruction problem using the tensor formulation.
The dictionary learning problem is presented as a non-negative tensor
factorization problem with sparsity constraints.
The reconstruction problem is formulated in a convex optimization framework
by looking for a solution with a sparse representation in the tensor dictionary.
%Our method for the tensor based dictionary learning and reconstruction
%is based on the notion of tensor-product introduced by Kilmer and Martin
%in their 2011 paper.
Numerical results show that our tensor formulation leads to very sparse
representations of both the training images and the reconstructions
due to the ability of representing repeated features compactly in the dictionary.
\end{abstract}

\section{Introduction}
Tomographic image reconstruction from noisy incomplete projection data
at few views or in a limited-angle setting is a common challenge
in computed tomography, yet classical methods such as filtered back-projection
(FBP) are well known to fail (see, e.g., \cite{Bian} and \cite{Frikel}).
To compute a robust solution it is necessary to incorporate adequate prior
information into the mathematical reconstruction formulation.
Total Variation (TV) regularization can be suited for edge-preserving
imaging problems in low dose and/or few-view data sets \cite{LaRoque};
but in the presence of noise
TV tends to oversmooth textures in natural images \cite{Strong}.

``Training images'' in the form of a carefully chosen set of images
can be used to represent samples of the prior\,---\,such as texture\,---\,for
the desired solution.
Clearly, such images must be reliable and application-specific.
A suitable way to incorporate the prior information from training images
and to extract the prototype features of such images is to form a dictionary
from the images such that they are reproducible from a limited (aka sparse)
linear combination of those images.
The dictionary is then used for representation of other images with
similar features.
Such ``sparse coding'' of natural images was introduced by
Olshausen and Field \cite{Olshausen}.

Recent development in the theory and computational tools for
sparse representation of signals and images (see, e.g., \cite{Elad}
and \cite{Bruckstein}) has enabled us to analyze massive training data.
Dictionary learning methods are now widely used to compute
basis elements and learn sparse representations of training signals and images
(see, e.g., \cite{Aharon} and \cite{Mairal}).
Likewise, sparse representation in terms of such
dictionaries has attracted increased interest in solving
imaging problems such as denoising \cite{Eladatel},
deblurring \cite{Liu}, and restoration \cite{MairalSapiro},
in addition to solving tomographic image reconstruction problems
\cite{Soltani,Xu}.

One common feature in the aforementioned papers on dictionary learning and
sparse representation in terms of these dictionaries is the reliance on the
(invertible) mapping of 2D images to vectors and subsequent use of a
linear algebraic framework:
Matrices are used for the dictionary representation (the columns represent
vectorized forms of image feature) and the use of a linear combination of the
columns of the dictionary gives the expression of
the image, in its vectorized form.
However, the training data itself can be more naturally represented as a
multidimensional array, called a tensor.
For example, a collection of $K$ gray-scale images
of size $M \times N$ could be arranged in an $M \times K \times N$ array,
also known as a third-order tensor.
Recent work in imaging applications such as facial recognition \cite{Hao}
and video completion \cite{ZhangTufts} has shown that using
the right kind of factorizations of particular tensor-based representations
of the data can have a distinct advantage over matrix-based counterparts.
For this reason, in this paper we will develop a fundamentally new approach
for both the dictionary learning and image reconstruction tasks
that is based on a particular type of tensor decomposition based around
the t-product introduced in \cite{Kilmer1}.

There are several different tensor factorizations and decompositions such
as CANDECOMP/PARAFAC (CP) \cite{Kiers} and Tucker decomposition \cite{Tucker}.
The use of different decompositions is driven by applications as well as
the properties of the decompositions.
For an extensive list of tensor decompositions,
their applications, and further references, see \cite{Kolda}.
Some recent works provide algorithms and analysis for tensor sparse coding
and dictionary learning based on different factorization strategies.
Caiafa and  Cichocki \cite{Caiafa} discuss multidimensional compressed
sensing algorithms using the Tucker decomposition.
Zubair and Wang \cite{Zubair} propose a tensor learning algorithm
based on the Tucker model with a sparsity constraint on the core tensor.
Tensor-based extensions of the method of optimal directions (MOD) \cite{Engan}
and the KSVD algorithm \cite{Aharon} have been studied in \cite{Roemer}
for separable multidimensional dictionaries.
An algorithm for tensor dictionary learning based on the CP decomposition,
called K-CPD, is presented in \cite{Duan}.

Recent work by Kilmer et al. \cite{Kilmer} sets up a new theoretical framework
which facilitates a straightforward extension of matrix factorizations to
third-order tensors based on a new tensor multiplication definition,
called the \emph{t-product}.
The motivation for our work is to use the t-product as a natural extension
for the dictionary learning problem and image reconstruction
in a third-order tensor formulation with the factorization based on
the framework in \cite{Kilmer1} and \cite{Kilmer}.

The goal of this paper is to re-visit the dictionary learning approach
in \cite{Soltani} for X-ray CT reconstruction, now using a tensor
formulation of the problem.
As we will show, the new formulation is not a trivial reformulation of the matrix-based
method, and we demonstrate that in the tensor formulation we obtain a
much sparser representation of both the dictionary and the reconstruction.

This paper is organized as follows.
We first establish basic notation and definitions in Section \ref{sec:not}.
In Section \ref{sec:back} we briefly discuss the dictionary learning problem and
the reconstruction problem using dictionaries.
In Section \ref{sec:TDL} we describe the tensor dictionary learning problem
and the corresponding algorithm based on the
alternating direction method of multipliers.
Then, in section \ref{sec:Rec} we utilize the learned tensor dictionary to
formulate the reconstruction problem as a regularized inverse problem.
Numerical results are presented in Section \ref{sec:Exp}
demonstrating the performance and parameter choice of our algorithm.

\section{Notations and Preliminaries on Tensors}
\label{sec:not}
In this section we present the definitions and notations
that will be used throughout this paper.
We exclusively consider the tensor definitions and the tensor product
notation introduced in \cite{Kilmer1} and \cite{Kilmer}.
Throughout the paper, a capital italics letter such as $A$ denotes a matrix
and a capital calligraphy letter such as $\mathcal{A}$ denotes a tensor.

\begin{figure} %[htp]
\centering
\subfigure{ \includegraphics[width=0.55\linewidth]{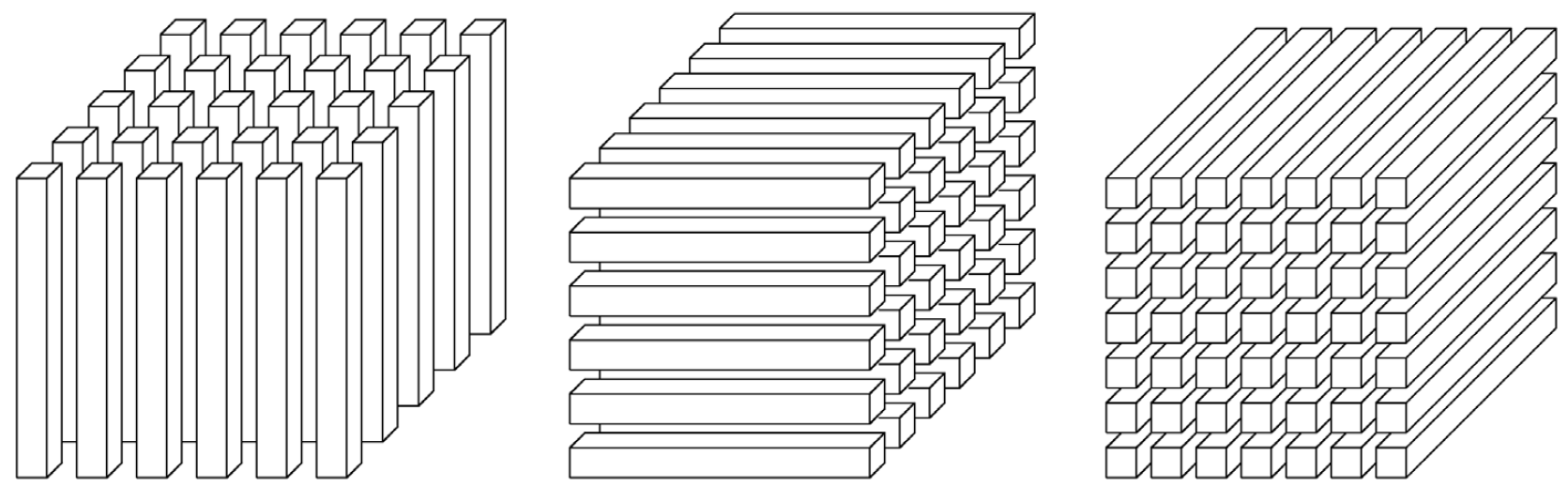}}
\subfigure{ \includegraphics[width=0.55\linewidth]{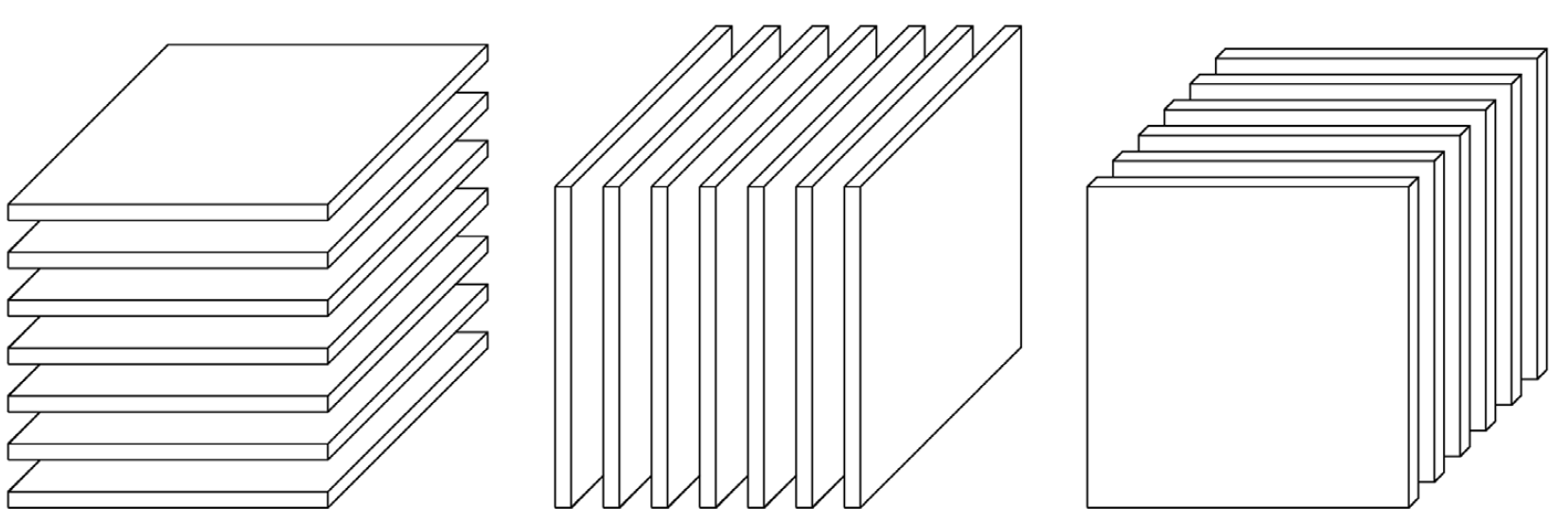}} \\
\caption{Different representations of a third-order tensor, from \cite{Kolda}.
Top left to right:\ column, row, and tube fibers.
Bottom left to right: horizontal, lateral, and frontal slices.}
\label{fig:FibersSlices}
\end{figure}

A \emph{tensor} is a multidimensional array of numbers.
The \emph{order} of a tensor refers to its dimensionality.
Thus, if $\mathcal{A} \in  \mathbb{R}^{l \times m \times n}$
then we say $\mathcal{A}$ is a third-order tensor.
A $1 \times 1 \times n$ tensor is called a \emph{tube fiber}.
A graphical illustration of a third-order tensor decomposed into its
tube fibers is given in the upper right image of Fig.~\ref{fig:FibersSlices}.
Thus, one way to view a third-order tensor is as a matrix of tube fibers.
In particular, an $\ell \times 1 \times n$ tensor is a vector of tube fibers.
To make this clear, we use the notation
$\vec{\mathcal{A}_j} = \mathcal{A}(:,j,:)$ to denote the $j$th ``column''
or \emph{lateral slice} of the third-order tensor
(see the middle figure of the bottom row of Fig.~\ref{fig:FibersSlices}).
The $k$th \emph{frontal slice}, which is an $\ell \times m$ matrix, is denoted by
$A^{(k)} \equiv \mathcal{A}(:, :,k)$.
Frontal slices and other decompositions of a third-order tensor are all shown in
Fig.~\ref{fig:FibersSlices}.

We can consider an $l \times 1 \times n$ tensor is a matrix oriented
into the third dimension.   It will therefore
be useful to use notation from \cite{Kilmer1} that allows us to easily move between
$l \times n$ matrices and their $l \times 1 \times n$ counterparts.
Specifically, the $\mathtt{squeeze}$ operation on
$\vec{\mathcal{X}} \in \mathbb{R}^{l\times 1 \times n}$ is identical to
the $\mathtt{squeeze}$ function in MATLAB:
  \[
    X = \mathtt{squeeze}(\vec{\mathcal{X}}) \quad \Rightarrow \quad
    X(i,k) =  \vec{\mathcal{X}}(i, 1, k).
  \]

The $\mathtt{vec}$ function unwraps the tensor $\mathcal{A}$ into a vector
of length $\ell mn$ by column stacking of frontal slices, i.e.,
in MATLAB notation:\ $\mathtt{vec}(\mathcal{A}) \equiv \mathcal{A}(:)$.
For the tensor $\mathcal{A}$ we define the \texttt{unfold} and
\texttt{fold} functions in terms of frontal slices:
  \begin{equation*}
    \mathtt{unfold}(\mathcal{A})= \begin{pmatrix}
     A^{(1)}  \\
     A^{(2)}  \\
     \vdots  \\
     A^{(n)}
    \end{pmatrix}\in \mathbb{R}^{ln \times m}, \qquad
    \mathtt{fold}(\mathtt{unfold}(\mathcal{A}))=\mathcal{A}.
  \end{equation*}
The block circulant matrix of size $\ell n \times nm$ that is generated
via $\mathtt{unfold}(\mathcal{A})$ is given as
  \begin{equation*}
     \mathtt{circ}(\mathcal{A})= \begin{pmatrix}
     A^{(1)} & A^{(n)} & A^{(n-1)} & \cdots & A^{(2)} \\
     A^{(2)} & A^{(1)} & A^{(n)} & \cdots & A^{(3)} \\
     \vdots  &         &         & \ddots &\vdots \\
     A^{(n)} & A^{(n-1)} & A^{(n-2)} & \cdots & A^{(1)}.
     \end{pmatrix} .
  \end{equation*}

\begin{mydef}
\label{def:tprod}
Let $\mathcal{B} \in \mathbb{R}^{l \times p \times n}$ and
$\mathcal{C} \in \mathbb{R}^{p \times m \times n}$.
Then the \textbf{t-product} from \cite{Kilmer1} is defined by
 \[
   \mathcal{A} = \mathcal{B}*\mathcal{C} \equiv
   \mathtt{fold} \bigl( \mathtt{circ}(\mathcal{B}) \,
   \mathtt{unfold}(\mathcal{C}) \bigr),
 \]
from which it follows that $\mathcal{A}$ is an $\ell \times m \times n$ tensor.
\end{mydef}
The t-product can be considered as a natural extension of the
matrix multiplication~\cite{Braman}.
In general the t-product is not commutative between two arbitrary tensors,
but it is commutative between tube fibers.

\begin{mydef}
Given $m$ tube fibers $\mathbf{c}_j \in \mathbb{R}^{1\times 1\times n}$,
$j=1,\ldots,m$ a \textbf{t-linear combination} \cite{Kilmer}
of the lateral slices $\vec{\mathcal{A}_j} \in \mathbb{R}^{\ell \times 1 \times n}$,
$j=1,\ldots,m$, is defined as
  \[
    \vec{\mathcal{A}_1}* \mathbf{c}_1+\vec{\mathcal{A}_2} *\mathbf{c}_2+
    \cdots +\vec{\mathcal{A}_m} *\mathbf{c}_m \equiv \mathcal{A}* \vec{\mathcal{C}\;},
  \]
where
  \[
    \vec{\mathcal{C}\;}=\begin{bmatrix}
    \mathbf{c}_1 \\ \vdots \\ \mathbf{c}_m \end{bmatrix} \in
    \mathbb{R}^{m \times 1 \times n} \ .
  \]
The multiplication $\mathbf{c}_j* \vec{\mathcal{A}}_j$ is not
defined unless $\ell =1$.
\end{mydef}

\begin{mydef}
The \textbf{identity tensor} $\mathcal{I}_{mmn}$ is the tensor whose first
frontal slice is the $m \times m$ identity matrix,
and whose other frontal slices are all zeros.
\end{mydef}
\begin{mydef}
An $m \times m \times n$ tensor $\mathcal{A}$ has an \textbf{inverse}
$\mathcal{B}$, provided that
 \[
   \mathcal{A} * \mathcal{B}= \mathcal{I}_{mmn} \quad \text{and} \quad
   \mathcal{B}*\mathcal{A} = \mathcal{I}_{mmn}.
 \]
\end{mydef}

\begin{mydef}
Following \cite{Kilmer1}, if $\mathcal{A}$ is $l \times m \times n$,
then the \textbf{transposed tensor} $\mathcal{A}^T$ is the $m \times l \times n$ tensor obtained by
transposing each of the frontal slices and then reversing the order of transposed frontal
slices $2$ through $n$.
\end{mydef}

\begin{mydef}
Let $a_{ijk}$ be the $i,j,k$ element of $\mathcal{A}$.
Then the \textbf{Frobenius norm} of the tensor $\mathcal{A}$ is
 \[
   \|\mathcal{A}\|_\mathrm{F} = \| \mathrm{vec}(\mathcal{A}) \|_2 =
   \sqrt{\sum_{i=1}^{l} \sum_{j=1}^{m} \sum_{k=1}^{n}a_{ijk}^2}.
 \]
\end{mydef}
We also use the following notation:
  \[
    \|\mathcal{A}\|_{\mathrm{sum}} = \| \mathrm{vec}(\mathcal{A}) \|_1 =
    \sum_{i,j,k}|a_{ijk}| , \qquad
    \| \mathcal{A} \|_{\max} = \| \mathrm{vec}(\mathcal{A}) \|_{\infty} =
    \max_{i,j,k} |a_{ijk}| .
  \]
If $A$ is a matrix then $\|A\|_{\mathrm{sum}}=\sum_{i,j}|a_{ij}|$.
Let $\sigma_i$, $i=1,\ldots,\min\{m,n\}$ denote the singular values of $A$.
The \emph{nuclear norm} (also known as the trace norm) is defined as
  \[
    \|A\|_*= \mathtt{trace}(\sqrt{A^TA})=\sum_{i=1}^{\min\{m,n\}}\sigma_i ,
  \]
where $A^T$ is the the transpose of $A$.

\section{The Dictionary Learning Problem}
\label{sec:back}
In classical dictionary learning, the problem is to find ``basis
elements'' and sparse representations of the
training signals/images.
That is, we want to write the input vectors
as a weighted linear combination of a small number of (unknown) basis vectors.

In practice, to find localized features of a training image (and to reduce the
computational work), training patches of smaller size are extracted.
Let the patches be of size $p \times r$ and let the matrix
$Y = [y_1, y_2,\ldots, y_t] \in  \mathbb{R}^{pr \times t}$
consist of $t$ training patches arranged as vectors/columns of length $pr$.
The resulting dictionary learning problem, based on this matrix formulation,
is then defined as follows: For the given data matrix
$Y$ and a given number of elements $s$, find
two matrices $D \in \mathbb{R}^{pr \times s}$ and
$H \in \mathbb{R}^{s \times t}$,
whose product approximates $Y$ as well as possible:\ $Y \approx DH$.
The columns of the \emph{dictionary matrix} $D$ (often with $s > pr$) represents
a ``basis'' of $s$ image patches, and the matrix $H$ contains
the representation of each training image patch approximated by the dictionary elements.
It is often assumed that, as $Y$ is non-negative, $D$ and $H$ should be constrained
to be non-negative as well.
Even with a non-negativity constraint, the decomposition is not unique~\cite{Donoho}.
Other constraints, such as sparseness,
statistical independence, and low complexity are often exploited in forming the basis
and the representation.
Different prior considerations lead to different
learning methods such as non-negative matrix factorization (NMF)
\cite{LeeSeung}, MOD \cite{Engan},
K-SVD \cite{Aharon}, the online dictionary learning method
\cite{Mairal} and many more (see, e.g., \cite{Elad} and \cite{Hastie}).

A framework for tomographic image reconstruction using matrix dictionary priors
was developed and described in \cite{Soltani}.
In tomography, a signal $b$ is measured from rays or signals passing through an object
of interest.
The discretized tomographic model is represented by an $m \times n$ matrix $A$.
Considering an unknown vector $x$ as vector of
pixels/voxels for the reconstructed image;
this yields a linear system $Ax \approx b$ with an ill-conditioned matrix~$A$.
Generally the image $x$ is not sparse but the situation
changes when we know that $x$ has a sparse representation in terms of a known basis.

Using a vectorized formulation in \cite{Soltani}, a global matrix dictionary $W$
is formed from the learned patch dictionary $D$, and the
linear tomographic problem is solved such that the vectorized image
$x$ is a conic combination of a small number of dictionary elements, i.e.,
$x = W\alpha$ and $\alpha$ is sparse, meaning that the solution lies in
the cone spanned by dictionary images
but that many of the weights are zero.
The simplicity of this approach is that once the basis elements have been determined,
the solution is linear in these new variables.

Images are naturally two-dimensional objects and we find that it is fundamentally
sound to work with them in their natural form.
For example we are looking for correlations image-to-image (not just pixel-to-pixel)
that let us reduce the overall redundancy in the data.
Therefore, we will consider a collection of training patches
as a third-order tensor, with each 2D image making up a slice of the data tensor.
It is natural to use higher-order tensor decomposition approaches,
which are nowadays frequently used in image analysis and signal processing
\cite{Cammoun,Caiafa,Cichocki,Hao,Kilmer}.
We describe this approach in more detail in the next section.
\section{Tensor Dictionary Learning}
\label{sec:TDL}

%\subsection{Using Tensors in Dictionary Learning}

%As mentioned in the introduction, two of the most popular decomposition or
%factorization models for multidimensional data are the CP and Tucker decompositions.
In recent years there has also been an increasing interest in
obtaining a non-negative \emph{tensor} factorization (NTF) (often based on CP and Tucker
decompositions) as a natural generalization of the NMF for a nonnegative data.
Similar to NMF, the sparsity of the representation has been empirically observed
in NTF based on CP and Tucker decompositions.
For NTF based on a subset of tensor decomposition methods,
we refer to \cite{Cichocki}.
Unlike the work in \cite{Cichocki}, we express the dictionary
learning problem in a third-order tensor framework based on the t-product.
This will be described in detail below, but the key is a
t-product-based NTF reminiscent of the NMF.

The NTF based on the t-product was proposed in \cite{HaoNNTFB}, where
preliminary work with MRI data showed the possibility that sparsity is
encouraged when non-negativity is enforced.
Here, we extend the work by incorporating sparsity constraints and we provide
the corresponding optimization algorithm.
Given the patch tensor dictionary $\mathcal{D}$, we
compute reconstructed images that have a sparse representation
in the space defined by the t-product and $\mathcal{D}$.
Thus, both the dictionary and the sparsity of the representation serve to
regularize the ill-posed problem.

\subsection{Tensor Factorization via t-Product}

%First we describe how to generate the \emph{data tensor} from
%which the dictionary tensor is obtained.
Let the third-order \emph{data  tensor} $\mathcal{Y} \in \mathbb{R}_+^{p\times t \times r}$
consist of $t$ training image patches of size $p \times r$, arranged as the
lateral slices of $\mathcal{Y}$, i.e.,
  \[
    \vec{\mathcal{Y}_j} = \mathcal{Y}(:,j,:), \quad \mathrm{for} \quad j=1,\ldots,t.
  \]
Our non-negative tensor decomposition problem, based on the t-product,
is the problem of writing the non-negative data tensor as a product
$\mathcal{Y} = \mathcal{D} * \mathcal{H}$
of two tensors $\mathcal{D} \in \mathbb{R}^{p\times s \times r}$
and $\mathcal{H} \in \mathbb{R}^{s \times t \times r}$.
The tensor $\mathcal{D}$ consists of $s$ dictionary 2D image patches
of size $p \times r$ arranged as the lateral slices of $\mathcal{D}$,
while $\mathcal{H}$ is the tensor of coefficients.
The main difference between NTF and NMF is that the $s\times t\times r$
tensor $\mathcal{H}$ has $r$ times more degrees of freedom in the representation
than the $s\times t$ matrix $H$.

The t-product from Definition \ref{def:tprod} involves unfolding and forming
a block circulant matrix of the given tensors.
Using the fact that a block circulant matrix can be block-diagonalized by the
Discrete Fourier Transform (DFT) \cite[\S 4.7.7]{Golub},
the t-product is computable in the Fourier domain \cite{Kilmer}.
Specifically, we can compute $\mathcal{Y} = \mathcal{D} * \mathcal{H}$
by applying the DFT along tube fibers of $\mathcal{D}$ and $\mathcal{H}$:
  \[
    \widehat{\mathcal{Y}}(:,:,k) =
    \widehat{\mathcal{D}}(:,:,k) \widehat{\mathcal{H}}(:,:,k), \qquad k = 1,\ldots,r ,
  \]
where $\,\widehat{ }\,$ denotes DFT; in MATLAB notation we apply
the DFT across the third dimension:\
$\widehat{\mathcal{Y}} = \mathtt{fft(} \mathcal{Y} \mathtt{,[ \ ],3)}$.
Working in the Fourier domain conveniently reduces the number of arithmetic
operations \cite{Hao}, and since the operation is
separable in the third dimension it allows for parallelism.

Although the representation of the training patches in the third-order tensor
resembles the matrix formulation,
it is not a re-formulation of the matrix problem packaged as tensors.
In fact, the tensor formulation gives a richer approach of
formulating the problem, as we now describe.

Recall that the $j$th patch $Y_j$ is the $j$th lateral slice of
$\mathcal{Y} = \mathcal{D}*\mathcal{H}$, i.e.,
$Y_j = \mathtt{squeeze}\bigl( \mathcal{Y}(:,j,:) \bigr)$.
Hence, as shown in \cite{HaoNNTFB},
  \begin{equation}
  \label{eq:rep}
    Y_j = \sum_{i=1}^{s}\mathtt{squeeze}\bigl(\mathcal{D}(:,i,:)\bigr)\,
    \mathtt{circ} \Bigl(\mathtt{squeeze} \bigl(\mathcal{H}(j,i,:)^T \bigr) \Bigr).
\end{equation}
In other words, the $j$th patch is a sum over all the lateral slices of $\mathcal{D}$,
each one ``weighted'' by multiplication with a circulant matrix derived
from a tube fiber of~$\mathcal{H}$.

We use a small example to show why this is significant.
Consider the $3 \times 3$ downshift matrix and the (column) circulant matrix
generated by the vector $v$:
  \[
    Z = \begin{pmatrix} 0 \ & 0 \ & 1 \\ 1 \ & 0 \ & 0 \\ 0 \ & 1 \ & 0 \end{pmatrix} ,
    \qquad
    C[v] = \mathtt{circ}(v) = \begin{pmatrix} v_1 & v_3 & v_2 \\ v_2 & v_1 & v_3 \\
      v_3 & v_2 & v_1 \end{pmatrix} .
  \]
Noting that
  \[
    C[v] = \sum_{k=1}^3 v_k Z^{k-1} = v_1 I +
    v_2 \begin{pmatrix} 0 \ & 0 \ & 1 \\ 1 \ & 0 \ & 0 \\ 0 \ & 1 \ & 0 \end{pmatrix} +
    v_3 \begin{pmatrix} 0 \ & 1 \ & 0 \\ 0 \ & 0 \ & 1 \\ 1 \ & 0 \ & 0 \end{pmatrix} .
  \]
it follows that
  \[
    D \, C[v] = \sum_{k=1}^3 v_k D Z^{k-1}.
  \]
Extrapolating to (\ref{eq:rep}), we obtain the following result.

\begin{theorem}
Let $Z$ denote the $n\times n$ downshift matrix.
With $D_i = \mathtt{squeeze} \bigl(\mathcal{D}(:,i,:)\bigr)$
and $h^{(ij)} = \mathtt{squeeze}(\mathcal{H}(j,i,:)^T)$,
the $j$th image patch is given by
  \begin{equation} \label{eq:decompose}
    Y_j = \sum_{i=1}^s  D_i C[h^{(ij)}]
    = \sum_{i=1}^s \left( h^{(ij)}_1 D_i + \sum_{k=2}^n h^{(ij)}_k D_i Z^{k-1} \right) .
  \end{equation}
\end{theorem}

To show the relevance of this result we note that the product $D_i Z^{k-1}$ is
$D_i$ with its columns cyclically shifted left by $k-1$ columns.
Assuming that $D_i$ represents a ``prototype'' element/feature in the image,
we now have a way of also including shifts of that prototype in our dictionary
\emph{without} explicitly storing those shifted bases in the dictionary.
Note that if $h_k^{(ij)} = 0$, $k=2,\ldots,n$ then $Y_j$ is a (standard)
linear combination of matrices $D_i$; this shows that our new approach
effectively subsumes the matrix-based approach from \cite{Soltani},
while making the basis richer with the storage of only a few entries
of a circulant matrix rather than storing extra basis image patches!

\subsection{Formulation of the Tensor-Based Dictionary Learning Problem}

One is usually not interested in a perfect factorization of the
data because over-fitting can occur, meaning that the learned parameters do
fit well the training data, but have a bad generalization performance.
This issue
is solved by making a priori assumptions on the dictionary and coefficients.

Based on the approximate decomposition $\mathcal{Y} \approx \mathcal{D}*\mathcal{H}$,
we consider the generic tensor-based dictionary learning problem
(similar to the matrix formulation from \cite{Soltani}):
 \begin{equation}
    \min_{\mathcal{D},\mathcal{H}} \quad
    \mathcal{L}_{\mathrm{dic}}(\mathcal{Y},\mathcal{D}*\mathcal{H}) +
    \Phi_{\mathrm{dic}}(\mathcal{D}) +
    \Phi_{\mathrm{rep}}(\mathcal{H}) .
  \end{equation}
The misfit of the factorization approximation is measured by the
loss function $\mathcal{L}_{\mathrm{dic}}$, (e.g., the Frobenius norm).
Different priors on the dictionary $\mathcal{D}$
and the representation tensor $\mathcal{H}$ are %
controlled by the regularization functions
$\Phi_{\mathrm{dic}}(\mathcal{D})$ and $\Phi_{\mathrm{rep}}(\mathcal{H})$.

NTF itself results in a sparse representation.
Imposing sparsity-inducing norm constraints on the representation
allows us to further enforce sparsity of the representation
of the training image, i.e., the training patches being represented
as a combination of a small number of dictionary elements.
At the same time this alleviates the non-uniqueness drawback of the NTF.

Therefore, we pose the tensor dictionary learning problem as
a non-negative sparse coding problem \cite{Hoyer}:
  \begin{equation}
  \label{eq:diclear}
    \min_{\mathcal{D}, \mathcal{H}}\  \nicefrac{1}{2}
    \|\mathcal{Y}-\mathcal{D}*\mathcal{H}\|_\mathrm{F}^2 +
    \lambda \|\mathcal{H}\|_{\mathrm{sum}} +
    I_{\mathsf{D}}(\mathcal{D}) +
    I_{\mathbb{R}_+^{s \times t \times r}}(\mathcal {H}).
\end{equation}
Here $\Dset$ is a closed set defined below,
$I_{Z}$ denotes the indicator function of a set $Z$, and $\lambda \geq 0$ is a
regularization parameter that controls the sparsity-inducing penalty
$\|\mathcal{H}\|_{\mathrm{sum}}$.
If we do not impose bound constraints on the dictionary elements, then the
dictionary and coefficient tensors $\mathcal{D}$ and $\mathcal{H}$ can be
arbitrarily scaled, because for any $\xi > 0$ we have
$\|\mathcal{Y}-(\xi\mathcal{D})*(\nicefrac{1}{\xi}\mathcal{H})\|^2_{\mathrm{F}}=
\|\mathcal{Y}-\tilde{\mathcal{D}}*\tilde{\mathcal{H}}\|^2_{\mathrm{F}}$.
We define the compact and convex set $\Dset$ such that $\mathcal{D} \in \Dset$
prevents this inconvenience:
  \begin{equation}
    \Dset \equiv  \{ \mathcal{D} \in \mathbb{R}_+^{p \times s \times r} ~|~
    \| \mathcal{D}(:,i,:) \|_{\mathrm{F}} \leq \sqrt{pr}
    , \ i=1,\ldots,s \} .
  \end{equation}
When $r = 1$ then \eqref{eq:diclear}
collapses to the standard non-negative sparse coding problem.

\subsection{The Tensor-Based Dictionary Learning Algorithm}

The optimization problem \eqref{eq:diclear} is non-convex,
while it is convex with respect to each variable $\mathcal{D}$ or $\mathcal{H}$
when the other is fixed.
Computing a local minimizer can be done using the
Alternating Direction Method of Multipliers (ADMM)~\cite{Boyd}, which is
a splitting method from the augmented Lagrangian family.
We therefore consider an equivalent form of \eqref{eq:diclear}:
  \begin{align} \label{eq:EquDicLear}
    \begin{array}{ll}
      \mbox{minimize}_{\mathcal{D}, \mathcal{H}, \mathcal{U}, \mathcal{V}}
      & \nicefrac{1}{2}\, \|\mathcal{Y}-\mathcal{U}*\mathcal{V}\|_{\mathrm{F}}^2 +
        \lambda \, \| \mathcal{H} \|_{\mathrm{sum}} +
        I_{\mathbb{R}_+^{s \times t \times r}}(\mathcal{H}) +
        I_{\Dset}(\mathcal{D}) \\
      \mbox{subject to}
      & \mathcal{D} = \mathcal{U} \quad \mathrm{and} \quad \mathcal{H} = \mathcal{V},
    \end{array}
  \end{align}
where $\mathcal{D}, \mathcal{U} \in  {\mathbb{R}}^{ p \times s \times r}$ and
$\mathcal{H}, \mathcal{V} \in {\mathbb{R}}^{s \times t \times r}$.
The augmented Lagrangian for \eqref{eq:EquDicLear} is
  \begin{align}
  \label{eq:Lag}
    \begin{split}
      L_{\rho}(\mathcal{D}, \mathcal{U} , \mathcal{H} , \mathcal{V}, \varLambda,\bar{\varLambda})\
        & = \ \nicefrac{1}{2} \|  \mathcal{Y}-\mathcal{U}*\mathcal{V} \|_{\mathrm{F}}^2 +
        \lambda \, \| \mathcal{H} \|_{\mathrm{sum}} +
        I_{\mathbb{R}_+^{s \times t \times r}}(\mathcal{H}) +
        I_{\Dset}(\mathcal{D}) \\
      & \qquad + \varLambda^T \odot (\mathcal{D} - \mathcal{U}) +
        \bar{\varLambda}^T \odot(\mathcal{H} - \mathcal{V}) \\
      & \qquad + \rho \bigl( \nicefrac{1}{2}\| \mathcal{D} - \mathcal{U} \|_{\mathrm{F}}^2 +
        \nicefrac{1}{2}\| \mathcal{H} - \mathcal{V} \|_{\mathrm{F}}^2 \bigr) ,
    \end{split}
  \end{align}
where $\varLambda \in {\mathbb{R}}^{p \times s \times r}$ and
$\bar{\varLambda} \in {\mathbb{R}}^{s \times t \times r}$ are Lagrange multiplier tensors,
$\rho~>~0$ is the quadratic penalty parameter, and
$\odot$ denotes the Hadamard (entrywise) product.

The objective function becomes separable by introducing the auxiliary variables
$\mathcal{U}$ and $\mathcal{V}$.
The alternate direction method is obtained by minimizing $L_{\rho}$ with respect to
$\mathcal{D}$, $\mathcal{H}$, $\mathcal{U}$, $\mathcal{V}$ one at a time
while fixing the other variables at their most recent values and updating the
Lagrangian multipliers $\varLambda$ and $\bar{\varLambda}$.
If $\mathsf{P}_{\Dset}$ is the metric projection on $\Dset$
(which is computed using
Dykstra's alternating projection algorithm \cite{Combettes}),
then the ADMM updates are given by:
  \begin{subequations}
  \label{update}
    \begin{alignat}{2}
    \mathcal{D}_{k+1} &= \min_{\mathcal{D} \in \Dset} L_\rho
      (\mathcal{D},\mathcal{H}_k,\mathcal{U}_k,\mathcal{V}_k,{\varLambda}_k,{\bar{\varLambda}}_k)
      = P_{\Dset}( \mathcal{U}_k - \rho^{-1}\varLambda_k) \\
    \mathcal{V}_{k+1} &= \min_{\mathcal{V}}
      L_\rho(\mathcal{D}_{k},\mathcal{H}_{k},\mathcal{U}_{k},\mathcal{V},
      {\varLambda}_k,{\bar{\varLambda}}_k) \\
    \notag &= \bigl(\mathcal{U}_k^T*\mathcal{U}_k+\rho \mathcal{I}\bigr)^{-1}*
      \bigl( \mathcal{U}_k^T*\mathcal{Y}+\bar{\varLambda}_k + \rho \mathcal{H}_k \bigr) \\
    \mathcal{H}_{k+1} &= \min_{\mathcal{H} \in \mathbb{R}_+^{s \times t \times r}} L_\rho
      (\mathcal{D}_{k+1},\mathcal{H},\mathcal{U}_k,\mathcal{V}_{k+1},
      {\varLambda}_k,{\bar{\varLambda}}_k) \\
    \notag & = \mathsf{P}_+ \bigl( \mathsf{S}_{\lambda/\rho} (\mathcal{V}_{k+1}-\rho^{-1}
      \bar{\varLambda}_k ) \bigr) \\
    \mathcal{U}_{k+1} &= \min_{\mathcal{U}} L_\rho
      (\mathcal{D}_{k+1},\mathcal{H}_k,\mathcal{U},\mathcal{V}_{k+1},
      {\varLambda}_k,{\bar{\varLambda}}_k) \\
    \notag &= \bigl( \mathcal{Y}*\mathcal{V}_{k+1}^T+\varLambda_k +
      \rho \mathcal{D}_{k+1} \bigr) * \bigl( \mathcal{V}_{k+1}*\mathcal{V}_{k+1}^T +
      \rho \mathcal{I} \bigr)^{-1} \\
    \varLambda_{k+1} & = \varLambda_{k}+ \rho (\mathcal{D}_{k+1}-\mathcal{U}_{k+1}) \\
    \bar{\varLambda}_{k+1} & = \bar{\varLambda}_{k}+ \rho
      (\mathcal{H}_{k+1}-\mathcal{V}_{k+1} ) .
    \end{alignat}
  \end{subequations}
Here $\mathsf{P}_+(\Theta)_{i,j} = \max\{\theta_{i,j},0\}$
and $\mathsf{S}_{\lambda/\rho}$ denotes soft thresholding.
The updates for $\mathcal{U}_{k+1}$ and $\mathcal{V}_{k+1}$
are computed in the Fourier domain.

The KKT-conditions for \eqref{eq:Lag} can be expressed as
  \[
    \mathcal{D} = \mathcal{U}, \quad \mathcal{H} = \mathcal{V},
  \]
  \[
    \varLambda = -(\mathcal{Y}-\mathcal{D}*\mathcal{H})*\mathcal{H}^T,
    \quad \bar{\varLambda} = -\mathcal{D}^T*(\mathcal{Y}-\mathcal{D}*\mathcal{H}),
  \]
  \[
    -\varLambda \in \partial \Phi_{\mathrm{dic}}(\mathcal{D}), \quad
    -\bar{\varLambda} \in \partial \Phi_{\mathrm{rep}}(\mathcal{H}),
  \]
where $\partial f(\mathcal{X})$ denotes the sub-differential of $f$ at $\mathcal{X}$.
The KKT conditions are used to formulate stopping criteria for the ADMM algorithm,
and we use the following conditions:
  \begin{subequations}
  \label{stcrit}
    \begin{alignat}{2}
    \frac{\|\mathcal{D}-\mathcal{U}\|_{\max}}{\max(1,\|\mathcal{D}\|_{\max})} \leq
     \epsilon, & \quad
     \frac{\|\mathcal{H}-\mathcal{V}\|_{\max}}{\max(1,\|\mathcal{H}\|_{\max})}
     \leq  \epsilon, \\
    \frac{\|\bar{\varLambda}-\mathcal{D}^T*(\mathcal{D}*\mathcal{H}-\mathcal{Y})\|_{\max}}
      {\max(1,\|\bar{\varLambda}\|_{\max})} \leq  \epsilon,  & \quad
      \frac{\|\varLambda-(\mathcal{D}*\mathcal{H}-\mathcal{Y})*\mathcal{H}^T\|_{\max}}
      {\max(1,\|\varLambda\|_{\max})} \leq  \epsilon,
    \end{alignat}
  \end{subequations}
where $\epsilon>0$ is a given tolerance.
Algorithm \ref{alg1} summarizes the algorithm to solve \eqref{eq:diclear}.
Note that satisfaction of the KKT conditions produces a local minimum;
this is not a guarantee of convergence to the global optimum.

\begin{algorithm}
 \caption{Tensor Dictionary Learning Algorithm}
 \label{alg1}
 \begin{algorithmic}
   \State {\bf Input}: Tensor of training image patches
     $\mathcal{Y} \in \mathbb{R}_+^{p \times t \times r}$, number of dictionary images $s$,
     tolerances $\rho,\epsilon >0$.
   \State {\bf Output}: Tensor dictionary
     $\mathcal{D}_k \in \mathbb{R}_+^{p \times s \times r}$,
     tensor representation $\mathcal{H}_k \in \mathbb{R}_+^{s \times t \times r}$.
   \State {\bf Initialization}:
     Let the lateral slices of $\mathcal{U}$ be randomly selected training patches,
     let $\mathcal{V}$ be the identity tensor,
     let $\mathcal{H}=\mathcal{V}$, and let ${\varLambda},{\bar{\varLambda}}$ be
     zero tensors of appropriate sizes.
   \For{$k~=~1,\ldots$}
     \State{Update $\mathcal{D}_k,\mathcal{H}_k,\mathcal{U}_k,\mathcal{V}_k,{\varLambda}_k,{\bar{\varLambda}}_n$ by means of \eqref{update}.
     \If {all stopping criteria \eqref{stcrit} are met}
     \State{Exit.}
     \EndIf
  \EndFor}
 \end{algorithmic}
\end{algorithm}

Under rather mild conditions the ADMM method can be shown to converge for all
values of the algorithm parameter $\rho$ in the Lagrange function $L_{\rho}$ \eqref{eq:Lag},
cf.~\cite{Boyd}.
Small values of $\rho$ lead to slow convergence; larger values give faster convergence
but puts less emphasis on minimizing the residual for the NTF.
For the convergene properties of ADMM and the impact of the parameter $\rho$
see \cite{Ghadimi} and the references therein.
\section{Tomographic Reconstruction with Tensor Dictionary}
\label{sec:Rec}
Recall that a linear tomographic problem
is often written $Ax  \approx b$ with $A \in \mathbb{R}^{m \times n}$,
where the vector $x$ represents the unknown $M \times N$ image,
the vector $b$ is the inaccurate/noisy data, and the
matrix $A$ represents the forward tomography model.
Since we assume that the vector $x$ represents an image of absorption coefficients
we impose a nonnegativity constraint on the solution.

Without loss of generality we assume that the
size of the image is a multiple of the patch sizes in the dictionary.
We partition the image into $q=(M/p)(N/r)$ non-overlapping patches
of size $(M/p)\times(N/r)$, i.e.,
$X_j\in \mathbb{R}^{p \times r}$ for $j=1,\ldots,q$.

In the matrix-based formulation of the reconstruction problem \cite{Soltani},
once the patch dictionary is formed we write the image patches we want to recover
(subvectors of the reconstructed image $x$) as conic combinations
of the patch dictionary columns.
The inverse problem then becomes one of recovering the expansion coefficients
subject to non-negativity constraints (which produces a nonnegative $x$
because the dictionary elements are nonnegative).

Here we define a similar reconstruction problem in our tensor-based formulation.
We arrange all the patches $X_j$ of the reconstructed image as
lateral slices of a $p \times q \times r$ tensor $\mathcal{X}$, i.e.,
  \[
    X_j = \mathtt{squeeze}(\vec{\mathcal{X}_j}) , \qquad
    \vec{\mathcal{X}_j} = \mathcal{X}(:j,:) , \qquad j=1,\ldots,q .
  \]
Moreover, we assume that there exists a $s \times q \times r$
coefficient tensor $\mathcal{C}$ such that the image patches
can be written as t-linear combinations of the patch dictionary elements, i.e.,
  \begin{equation}
    \mathcal{X} = \mathcal{D} * \mathcal{C} \qquad \Leftrightarrow \qquad
    \vec{\mathcal{X}_j} =  \mathcal{D} * \vec{\mathcal{C}_j} , \quad
    j=1,\ldots,q ,
  \end{equation}
where the tube fibers of $\vec{\mathcal{C}_j} = \mathcal{C}(:,j,:)$
can be considered as the expansion coefficients.
In other words, we restrict our solution so that it is a t-linear combination
of the dictionary images.

Then, similar to \eqref{eq:rep}, each patch $X_j$ in the reconstruction
can be built from the matrices $\mathtt{squeeze}(\vec{\mathcal{D}_i})$,
$i=1\ldots,s$:
  \begin{equation}
  \label{eq:BlcokRep}
    X_j = \mathtt{squeeze} \bigl( \mathcal{D}*\vec{\mathcal{C}_j} \bigr) =
    \sum_{i=1}^{s}\mathtt{squeeze} \bigr( \vec{\mathcal{D}_i} \bigl)
    \, \mathtt{circ} \Bigl( \mathtt{squeeze} \bigr(
    \vec{\mathcal{C}_j}(i,1,:)^T \bigl) \Bigl) .
\end{equation}
Since the circulant matrices are not scalar multiples of the identity matrix,
$X_j$ is not a simple
linear combination of the matrices $\mathtt{squeeze}(\vec{\mathcal{D}_i})$.

Thus, we want to find a tensor $\mathcal{C}$ such that
$\mathcal{X} = \mathcal{D}*\mathcal{C}$ solves the reconstruction problem,
and to ensure a nonnegative reconstruction, we enforce
non-negativity constraints on $\mathcal{C}$.
Then we write the vectorized image as $x = \Pi \mathtt{vec}(\mathcal{D}*\mathcal{C})$,
where the permutation matrix $\Pi$ ensures the correct shuffling of the
pixels from the patches.
Then our generic reconstruction problem takes the form
  \begin{equation}
  \label{eq:recgen}
    \min_{{\mathcal{C}}} \
    \mathcal{L}_{\mathrm{rec}}
    \bigl( A \Pi \mathtt{vec}(\mathcal{D}*\mathcal{C}),b \bigr) +
    \Phi_{\mathrm{sp}}({\mathcal{C}}) +
    \Phi_{\mathrm{im}}(\mathcal{D}*{\mathcal{C}}) , \qquad \mathcal{C} \geq 0 .
  \end{equation}
The data fidelity is measured by the loss function $\mathcal{L}_{\mathrm{rec}}$,
and regularization is imposed via $\Phi_{\mathrm{sp}}$ which enforces
a \underline{sp}arsity prior on ${\mathcal{C}}$,
and $\Phi_{\mathrm{im}}$ which enforces an \underline{im}age prior on the reconstruction.
By choosing these three functions to be convex,
we can solve \eqref{eq:recgen} by means of convex optimization methods.

Our patches are non-overlapping because overlapping patches tend to produce blurring
in the overlap regions of the reconstruction.
Non-overlapping patches may
give rise to block artifacts in the reconstruction, because the objective in
the reconstruction problem does not penalize jumps across the values
at the boundary of neighboring patches.
To mitigate this type of jumps, we add an image penalty term
$\Phi_{\mathrm{im}}(\mathcal{D}*\mathcal{C}) =
\delta^2 \psi(\Pi  \mathtt{vec}(\mathcal{D}*\mathcal{C}))$
that discourages such artifacts, where $\delta$ is a regularization parameter,
and the function $\psi$ is defined by
 \begin{equation}
  \label{eq:psi}
    \psi(z) =  \frac{1}{M(M/p-1)+N(N/r-1)} \, \frac{1}{2} \, \| L z \|_2^2.
  \end{equation}
The matrix $L$ is a defined such that $Lz$ is a vector with finite-difference
approximations of the vertical/horizontal derivatives across the patch boundaries.
The denominator is the
total number of pixels along the boundaries of the patches in the image.

We consider two different ways to impose a sparsity prior on $\mathcal{C}$
in the form $\Phi_{\mathrm{sp}}({\mathcal{C}}) = \mu\, \varphi_\nu(\mathcal{C})$,
$\nu = 1,2$, where $\mu$ is a regularization parameter and
  \begin{equation}
  \label{eq:regterms}
    \varphi_1(\mathcal{C}) = \nicefrac{1}{q} \| \mathcal{C} \|_{\mathrm{sum}},
    \qquad \varphi_2(\mathcal{C}) = \nicefrac{1}{q} \bigl( \| \mathcal{C} \|_{\mathrm{sum}}
    + \| C \|_* \bigr) ,
  \end{equation}
in which the $sq \times r$ matrix $C$ is defined as
  \[
    C = \begin{pmatrix} \mathtt{squeeze}\bigl( \vec{\mathcal{C}_1} \bigr) \\
    \vdots \\ \mathtt{squeeze}\bigl( \vec{\mathcal{C}_q} \bigr) \end{pmatrix} .
  \]
The first prior $\varphi_1$ corresponds to a standard sparsity prior in
reconstruction problems.
The second prior $\varphi_2$, which tends to produce a sparse and
low-rank $C$, is inspired by a similar use in compressed sensing~\cite{M. Golbabaee}.

To summarize, we consider a reconstruction problem of the form
  \begin{align}
  \label{eq:trec}
    \begin{array}{ll} \displaystyle
    \mbox{minimize}_{{\mathcal{C}}} \
      & \frac{1}{2m} \| A \Pi \mathtt{vec}(\mathcal{D}*\mathcal{C}) - b \|_2^2
      +  \mu \varphi_{\nu} \bigl( \mathcal{C} \bigr)
      + \delta^2 \psi \bigl( \Pi  \mathtt{vec}(\mathcal{D}*\mathcal{C}) \bigr) \\[2mm]
    \mbox{subject to} \
      & {\mathcal{C}} \geq 0 ,
    \end{array}
 \end{align}
where $\mu$ and $\delta$ are regularization parameters.
We note that
\eqref{eq:trec} is a convex but non-differentiable optimization problem.
It is solved using the software package TFOCS (Templates for First-Order Conic Solvers)
\cite{Becker}.
The implementation details are included in the Appendix.

We note that imposing the non-negativity constraint on the solution implies that
each image patch $X_j$ belongs to a closed set defined by
  \begin{equation}
  \label{eq:cone}
    \Gset = \{ \mathcal{D} * \vec{\mathcal{Z}}~|~\vec{\mathcal{Z}}
    \in \mathbb{R}_+^{s \times 1 \times r} \} \ \subseteq \
    \mathbb{R}_+^{p \times 1 \times r}.
  \end{equation}
The set $\Gset$ is a cone, since for any $\vec{\mathcal{V}} \in \Gset$
and any nonnegative tube fiber $\mathbf{c} \in \mathbb{R}^{1 \times 1 \times r }$
the product $\vec{\mathcal{V}}*\mathbf{c}$ belongs to $\Gset$.
Clearly, if the dictionary $\mathcal{D}$ contains the standard basis that
spans $\mathbb{R}_+^{p \times 1 \times r}$ then $\Gset$
is equivalent to the entire nonnegative orthant $\mathbb{R}_+^{p \times 1 \times r}$,
and any image patch $X_j$ can be reconstructed by a t-linear combination
of dictionary basis images.
However, in the typical case where $\Gset$ is a proper subset of
$\mathbb{R}_+^{p \times 1 \times r}$ then not all nonnegative images
have an exact representation in $\Gset$, leading to an approximation error.
\section{Numerical Experiments}
\label{sec:Exp}
\begin{figure}%[htp]
\centering
\subfigure{ \includegraphics[width=0.3\linewidth]{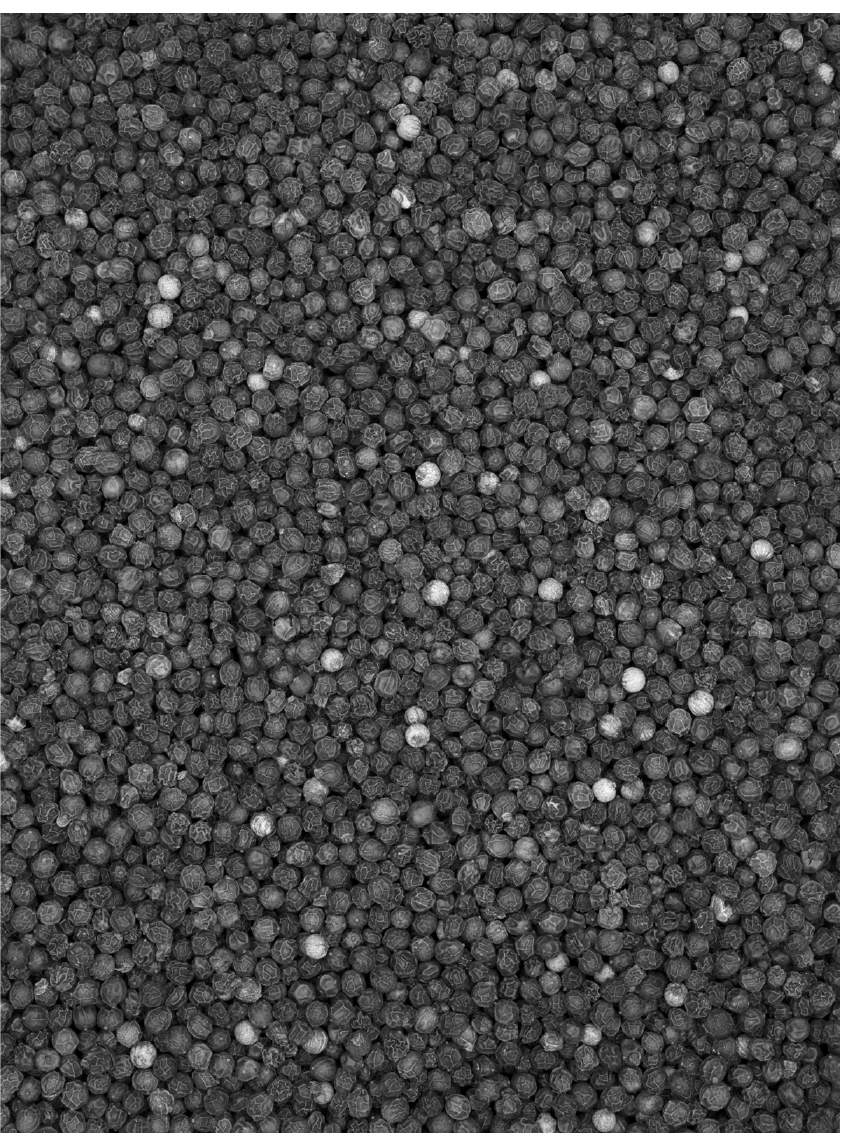}}
\subfigure{ \includegraphics[width=0.3\linewidth]{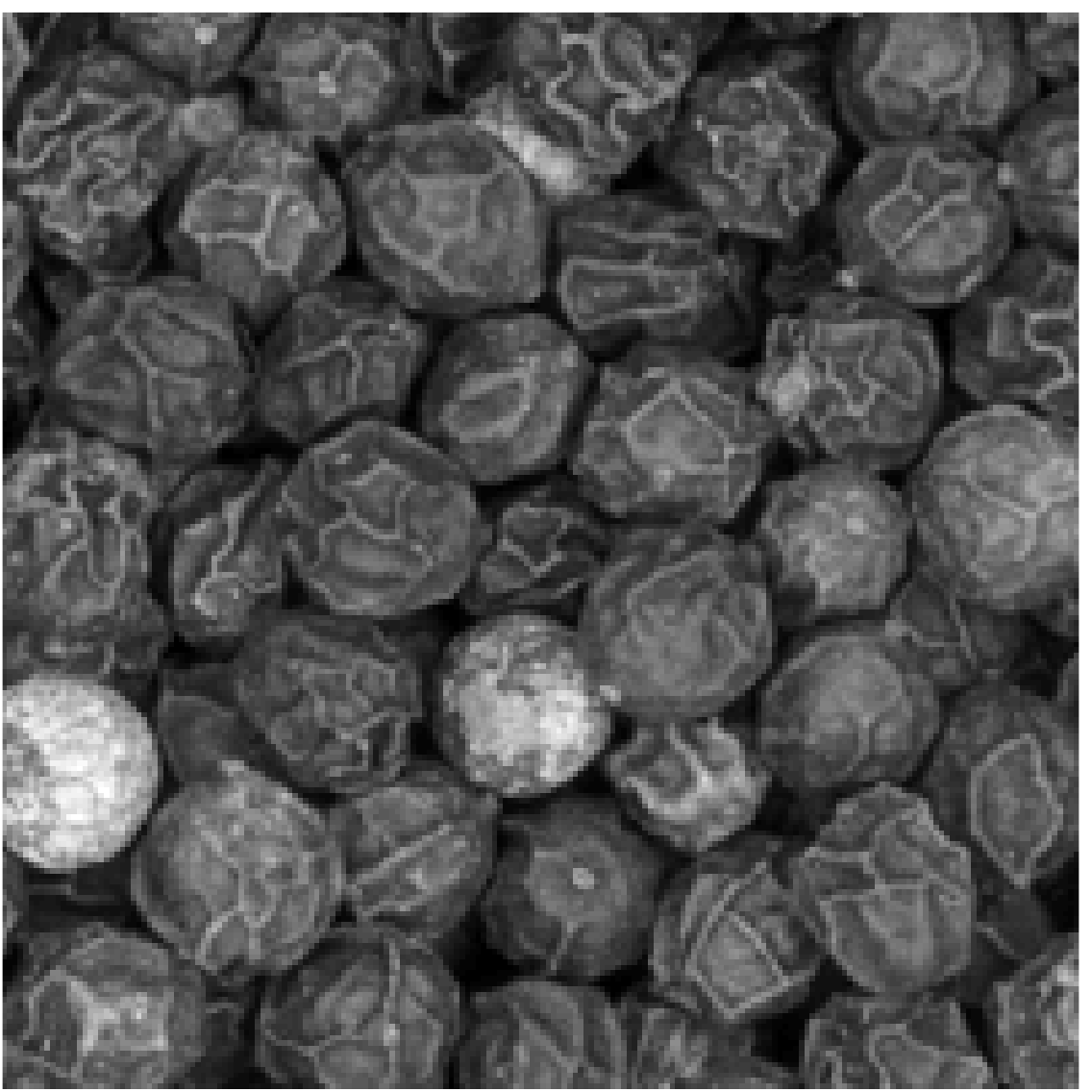}}
\caption{Left:\ the high-resolution image.
Right:\ the $200 \times 200$ exact image $x^{\mathrm{exact}}$.}
\label{fig:Images}
\end{figure}

We conclude with computational tests to examine the tensor formulation.
All experiments are run in MATLAB (R2014a) on a 64-bit Linux system.
The reconstruction problems are solved using the software package TFOCS version 1.3.1
\cite{Becker} and compared with results from the matrix-based approach~\cite{Soltani}.
We use a $1600 \times 1200$ high-resolution photo of peppers;
from this image we extract the $p\times r$ training image patches, as well as
the $200\times 200$ ground-truth or exact image $x^{\mathrm{exact}}$,
see Fig.~\ref{fig:Images}.
The exact image is not contained in
the training set, so that we avoid committing an inverse crime.
All the images are gray-level and scaled in the interval $[0, 1]$.

\subsection{Dictionary Learning Experiments}

Patch sizes should be sufficiently large to capture the desired structure
in the training images, but the computational
cost of the dictionary learning increases with the patch size.
A study of the patch size $p \times r$ and number $s$ of elements
in \cite{Soltani} shows that a reasonably large patch size gives a good trade-off
between the computational work and the approximation error by the dictionary,
and that the over-representation factor $s/(pr)$ can be smaller for larger patches.
For these reasons, we have chosen $p=r=10$ and (unless otherwise noted) $s=300$
for both the dictionary learning and tomographic reconstruction studies.
We extract 52,934 patches from the high-resolution image and apply
Algorithm~\ref{alg1} to learn the  dictionary.
The tensor dictionary $\mathcal{D}$ and the coefficient tensor $\mathcal{H}$
are $10\times 300 \times 10$ and $300 \times 52934 \times 10$, respectively.

\begin{figure}%[htp]
\centering
\includegraphics[width=1\linewidth]{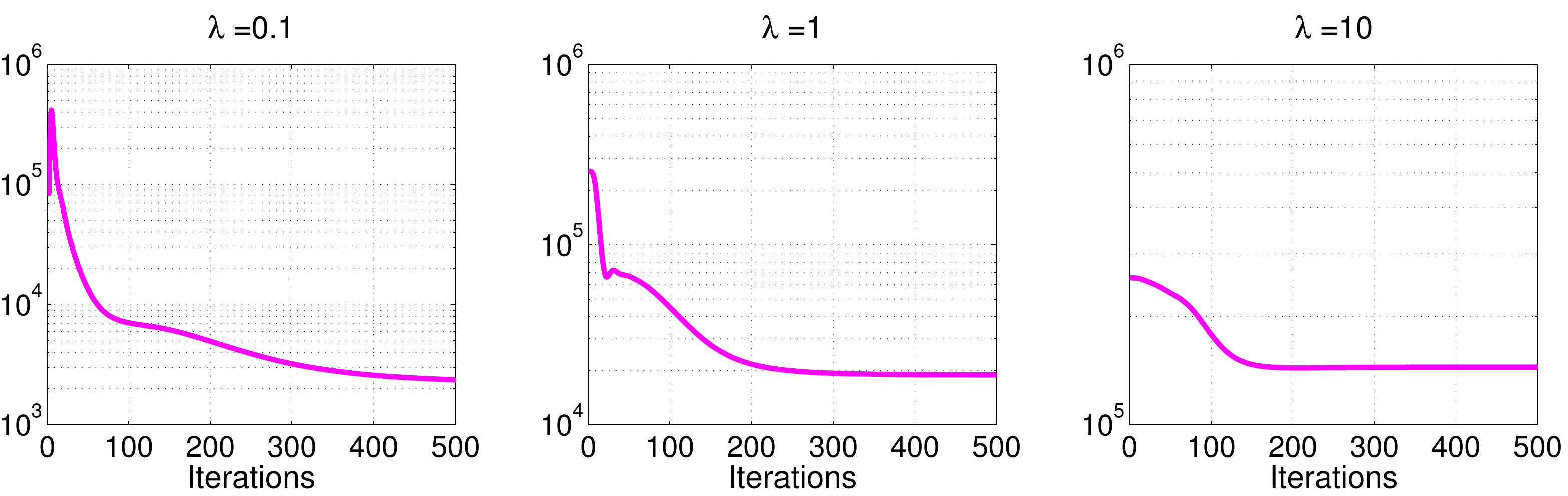}
\caption{Convergence of Algorithm \ref{alg1} for $\lambda=0.1$, 1, and 10.
We plot $\nicefrac{1}{2}\|\mathcal{Y}-\mathcal{D}*\mathcal{H}\|^2_\mathrm{F}+
\lambda\|\mathcal{H}\|_\mathrm{sum}$ versus the number of iterations.
Note the different scalings of the axes.}
\label{fig:conv}
\end{figure}

Convergence plots for $\lambda=0.1$, 1, and 10 are shown in Fig.~\ref{fig:conv}.
For $\lambda=10$ we put emphasis on minimizing the sparsity
penalty, and after about 200 iterations we have reached convergence where the
residual term dominates the objective function.
For $\lambda = 0.1$ we put more emphasis on minimizing the
residual term, and we need about 500 iterations to converge; now the objective
function is dominated by the sparsity penalty.

\begin{figure}%[htp]
\centering
\includegraphics[width=0.45\linewidth]{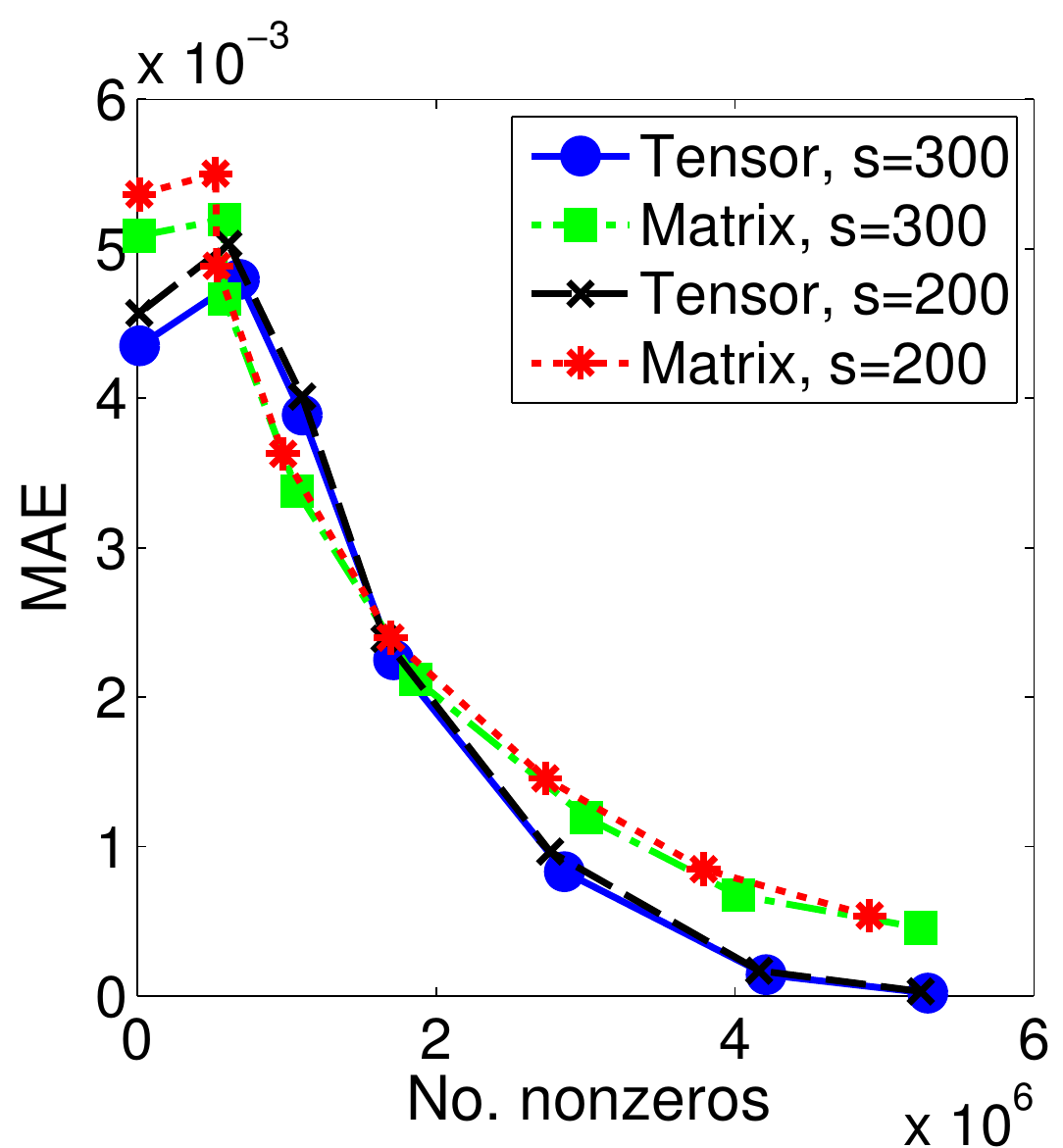}
\caption{The mean approximation error MAE \eqref{eq:MAE} for the tensor and matrix
  formulations versus the number of nonzeros of $\mathcal{H}$ and $H$, respectively,
  as functions of~$\lambda$ (small $\lambda$ give a larger number of nonzeros).}
\label{fig:DicPlots}
\end{figure}

Next we consider the approximation errors mentioned in the previous section.
Following \cite{Soltani}, a way to bound these errors
is to consider how well we can approximate the exact image $x^{\mathrm{exact}}$
with patches in the cone $\Gset$ \eqref{eq:cone} defined by the dictionary.
Consider the $q$ approximation problems for all blocks $X^{\mathrm{exact}}_j$,
$j=1,2,\ldots,q$, of the exact image:
  \begin{equation*} \textstyle
    \min_{\vec{\mathcal{C}_j}} \nicefrac{1}{2} \bigl\|
    \mathcal{D}*\vec{\mathcal{C}_j} - X^{\mathrm{exact}}_j \bigr\|_{\mathrm{F}}^2
    \qquad \mathrm{s.t.} \qquad \vec{\mathcal{C}_j} \geq 0.
  \end{equation*}
If $\vec{\mathcal{C}_j}^{\star}$ denotes the solution to the $j$th problem,
then $\mathrm{vec}(\mathcal{D}*\vec{\mathcal{C}_j}^\star)$ is the best
approximation in $\Gset$ of the $j$th block $X^{\mathrm{exact}}_j$.
We define the \emph{mean approximation error} as
  \begin{equation*}
  \label{eq:MAE}
    \mathrm{MAE} = \frac{1}{pqr} \sum_{j=1}^q \bigl\|
    \mathcal{D}*\vec{\mathcal{C}_j}^\star - X^{\mathrm{exact}}_j \bigr\|_{\mathrm{F}} .
  \end{equation*}
The MAE is defined analogously for the matrix formulation in \cite{Soltani}.
Figure \ref{fig:DicPlots} shows how these MAEs vary with the number of nonzeros of
$\mathcal{H}$ and $H$, as a function of $\lambda$, for both $s=200$ and $s=300$.
This plot shows that for a given number of nonzeros in $\mathcal{H}$ or $H$
we obtain approximately the same mean approximation error.
In other words despite the fact that the $s \times t \times r$ tensor $\mathcal{H}$
has $r$ times more degrees of freedom in the representation than the
$s \times t$ matrix $H$, we do not need more nonzero values to represent
our training images.
Hence the number of nonzeros is not a decisive factor.

\begin{figure}%[htp]
\centering
\includegraphics[width=0.45\linewidth]{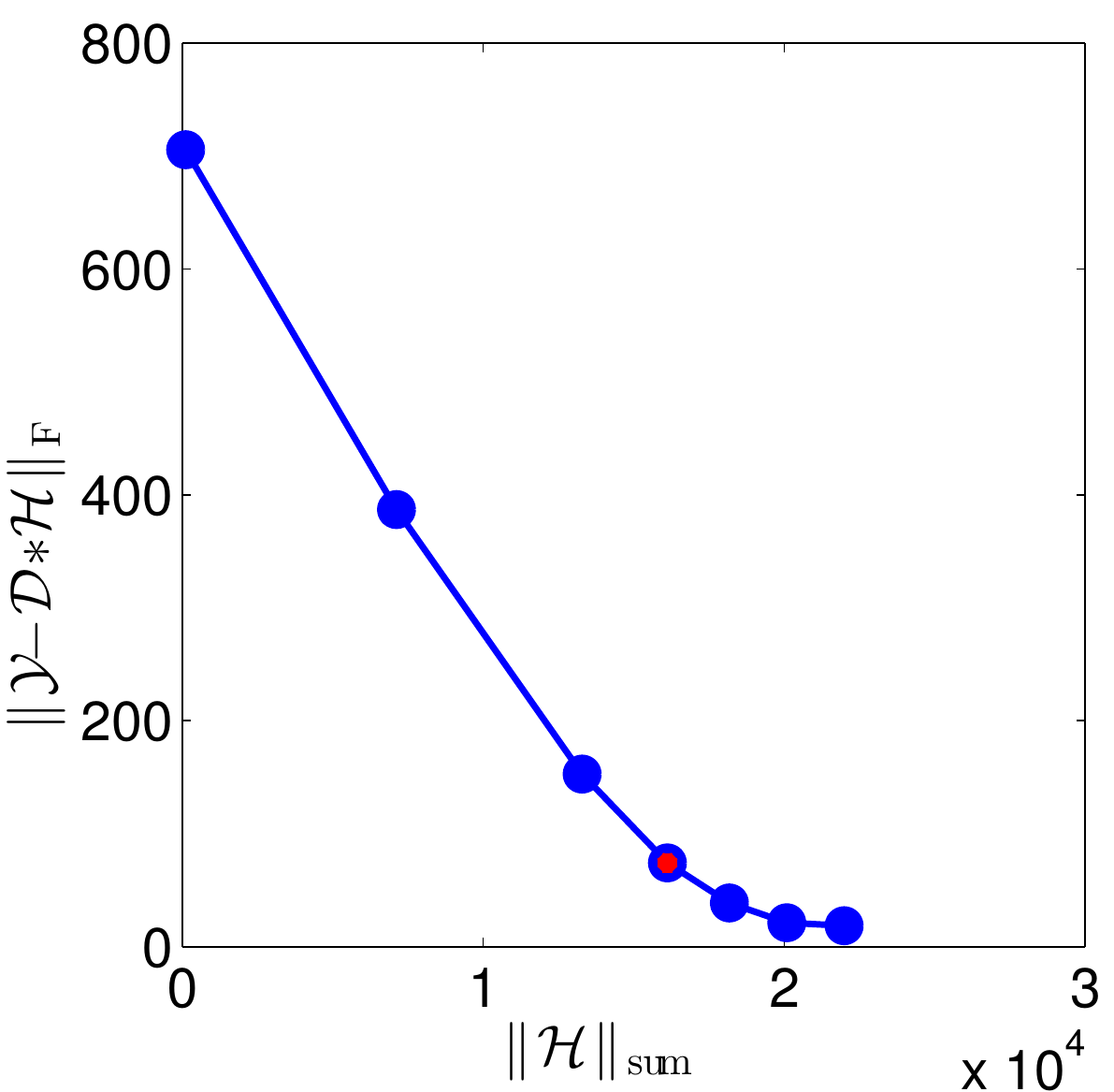}
\caption{A trade-off curve for the tensor dictionary learning problem;
  the red dot denotes the value $\lambda=3.1623$ that minimizes
  $\|\mathcal{H}\|_{\mathrm{sum}}^2 +
  \| \mathcal{Y} - \mathcal{D} * \mathcal{H} \|_{\mathrm{F}}^2$.}
\label{fig:X}
\end{figure}

In Fig.~\ref{fig:DicPlots} we note that for large enough $\lambda$ both $\mathcal{H}$ and $H$
consist entirely of zeros, in which case the dictionaries $\mathcal{D}$ and $D$
are solely determined by the constraints.
Hence, as $\lambda$ increases the MAE settles at a value that is almost independent
on~$\lambda$.

To determine a suitable value of the regularization parameter $\lambda$ in \eqref{eq:diclear}
we plot the residual norm $\|\mathcal{Y}-\mathcal{D}*\mathcal{H}\|_{\mathrm{F}}$
versus $\|\mathcal{H}\|_{\mathrm{sum}}$ for various $\lambda \in [0.1,\,100]$
in Fig. \ref{fig:X}.
We define the optimal parameter to be the one that minimizes
$\|\mathcal{H}\|_{\mathrm{sum}}^2 + \| \mathcal{Y} - \mathcal{D} * \mathcal{H} \|_{\mathrm{F}}^2$,
which is obtained for $\lambda=3.1623$, and we use this value throughout
the rest of our experiments.

\begin{figure}%[htp]
\centering
\includegraphics[width=0.45\linewidth]{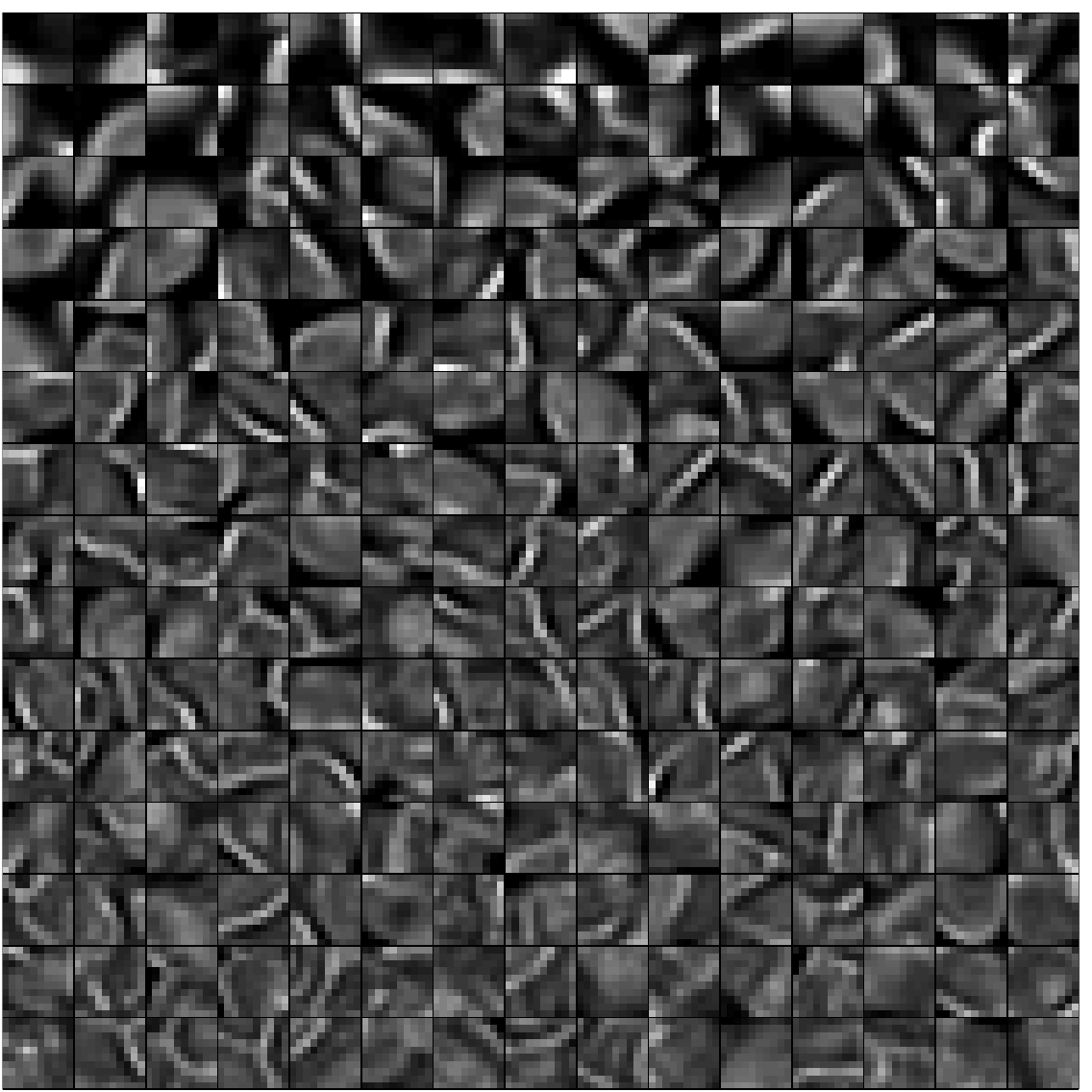} \quad
\includegraphics[width=0.45\linewidth]{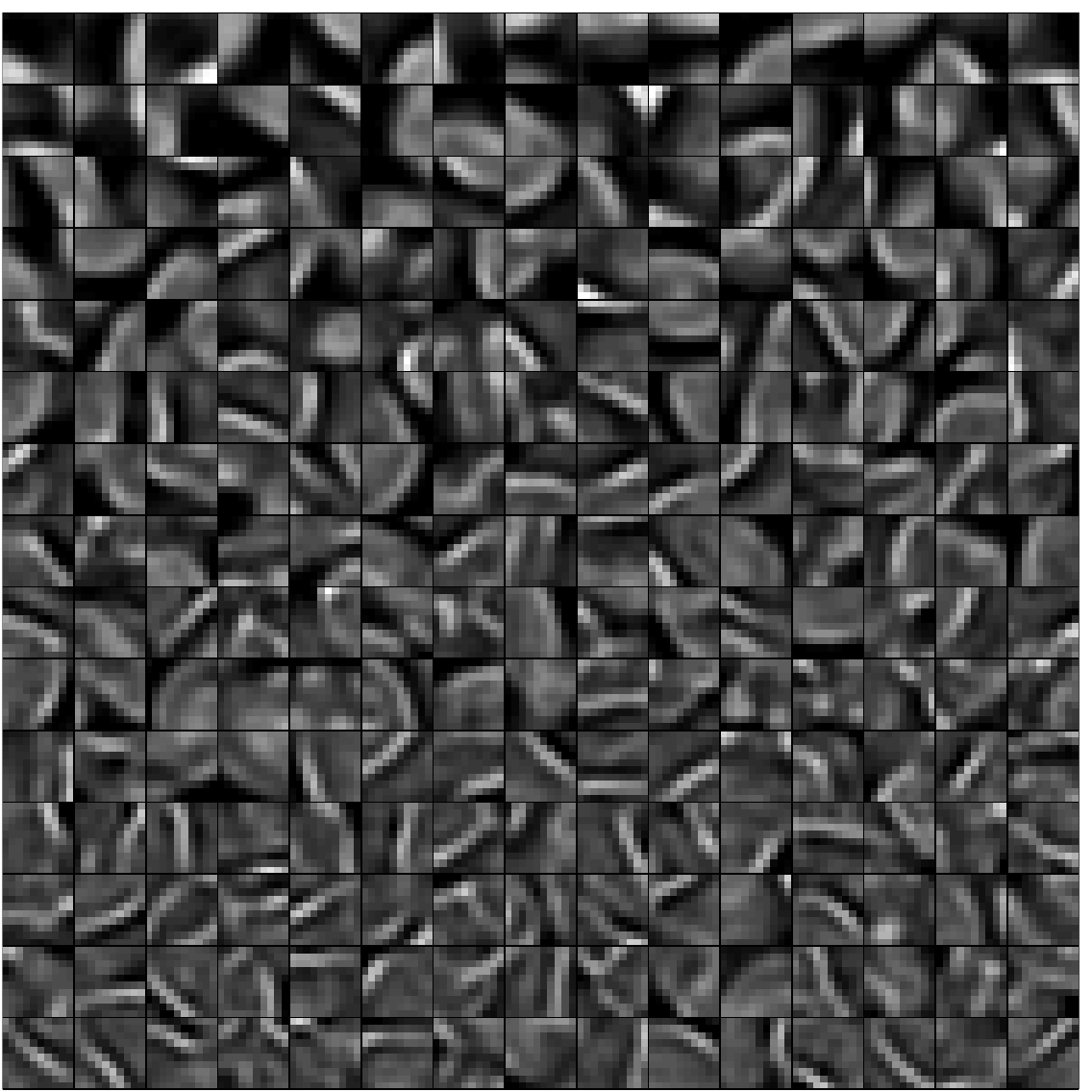}
\caption{Examples of dictionary elements/images from the tensor formulatio (left)
and the matrix formulation (right) with $10 \times 10$ patches and $\lambda=3.1623$
and $s=300$.}
\label{fig:DicImages}
\end{figure}

Figure \ref{fig:DicImages} shows examples of tensor and matrix dictionary
elements/images, where lateral slices of the tensor dictionary and
columns of the matrix dictionary are represented as images.
The dictionary images are sorted according to increasing variance.
The tensor and matrix dictionary images are different but they are visually similar.

\begin{figure}%[htp]
\centering
\includegraphics[width=0.40\linewidth]{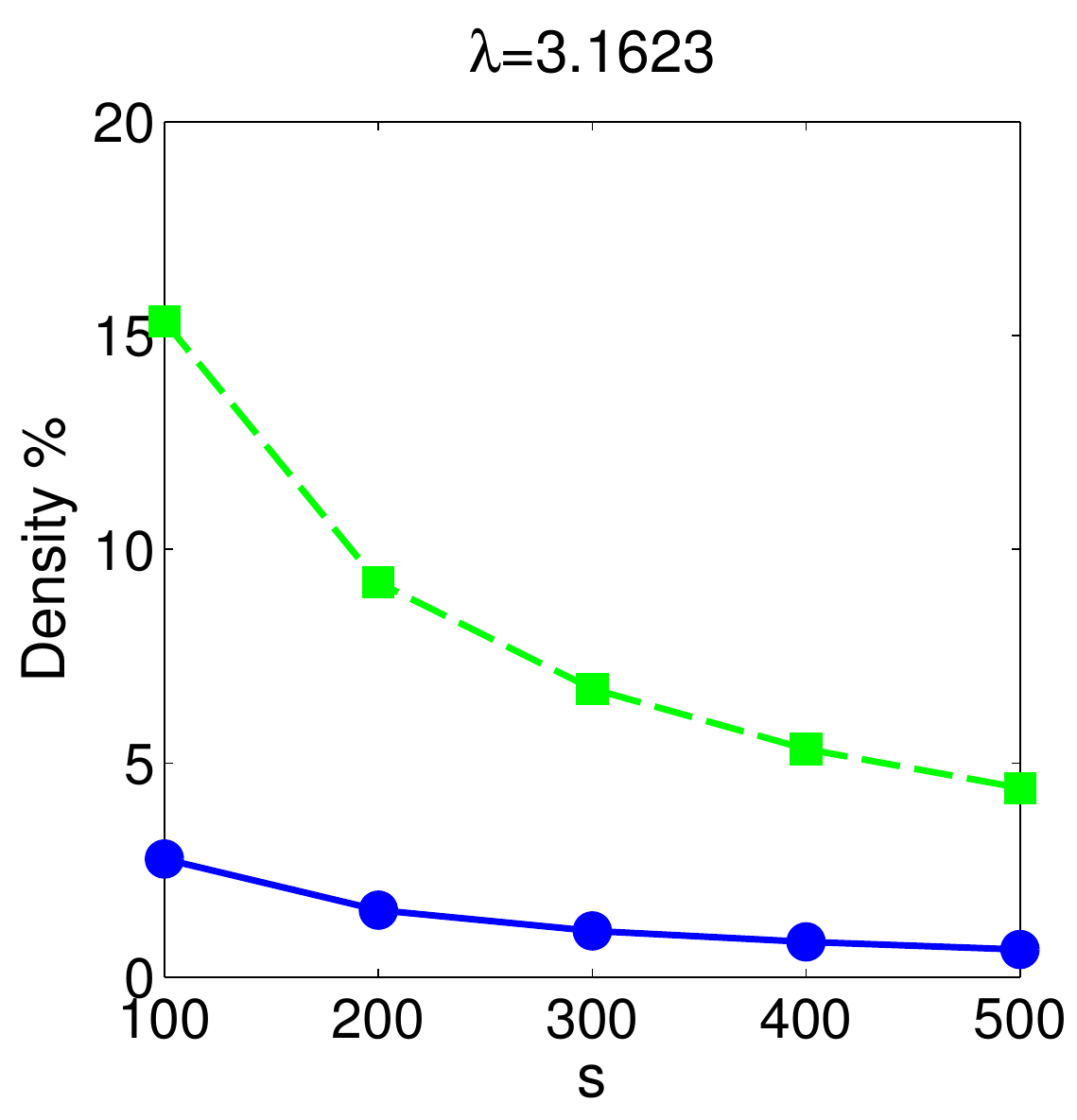} \quad
\includegraphics[width=0.40\linewidth]{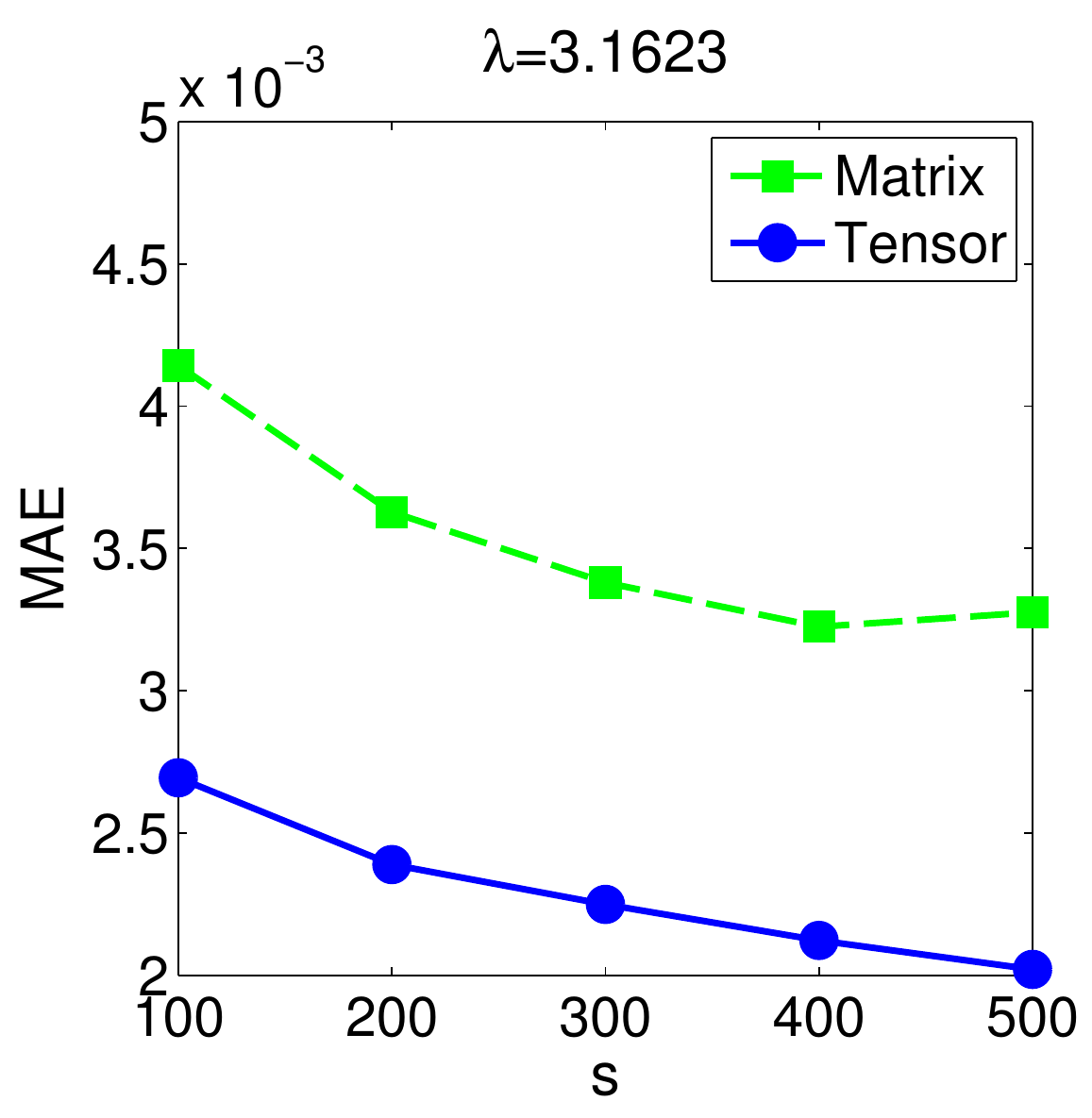}
\caption{Dependence of the dictionary on the number of dictionary elements~$s$,
for both the tensor and matrix formulations.
Left:\ the density of $\mathcal{H}$ and $H$.
Right:\ the MAE associated with the dictionaries.}
\label{fig:densityMAEs}
\end{figure}

We conclude these experiments with a study of how the number $s$ of dictionary
elements influences the dictionary, for the fixed $\lambda = 3.1623$.
Specifically, Fig.~\ref{fig:densityMAEs} shows how the density and the MAE
varies with $s$ in the range from 100 to 500.
As we have already seen for $s=300$ the density of $\mathcal{H}$ is consistently
much lower than that of $H$, and it is also less dependent
on $s$ in the tensor formulation.
We also see that the MAE for the tensor formulation is consistently lower
for the tensor formulation:\ even with $s=400$ dictionary elements in
the matrix formulation we cannot achieve the tensor formulation's low MAE
for $s=100$.
These results confirm our intuition that the tensor formulation is better
suited for sparsely representing the training image, because due to the
ability of capturing repeating fectures we can use a much smaller dictionary.

\subsection{Reconstruction Experiments}
\label{sec:recExp}

In this section we present numerical experiments for 2D tomographic
reconstruction in few-projection and noisy settings.
We perform two different experiments to analyze our algorithm:\
first we examine the role of different regularization terms
and then we study the reconstruction quality in different tomography scenarios.
We also present results using a more realistic test problem.

Recall from Section \ref{sec:Rec} that our reconstruction
model has the form $A\, x \approx b$ with $A\in\mathbb{R}^{m \times n}$.
We consider parallel-beam geometry and the test
problem is generated by means of the function \texttt{paralleltomo}
from \textsc{AIR Tools} \cite{Hansen}.
The exact data is generated by the forward model
$b^{\mathrm{exact}} = Ax^{\mathrm{exact}}$, to which we add white Gaussian noise.

The accuracy of the reconstruction is measured by the relative 2-norm error
  \[
    \mathrm{RE} = \|x^{\mathrm{exact}}-x\|_2 \, / \, \|x^{\mathrm{exact}}\|_2 .
  \]
We also report the structural similarity index measure (SSIM) \cite{Wang} (recall that
a larger SSIM means a better reconstruction).
We remind that the error is due to the combination of the approximation error, the error
from the data, and the regularization error.

The parameters $\delta$ and $\mu$ in the reconstruction problem \eqref{eq:trec}
both play a role in terms of regularization;
to simplify \eqref{eq:trec} we set $\tau = \mu / q$.
As described in \cite{Soltani}, a nonnegative constraint in the reconstruction
problem plays an extra role of regularization and therefore
the reconstruction is not very sensitive to the regularization parameters
$\delta$ and $\tau$, hence they are chosen from a few numerical experiments
such that a solution with the smallest error is obtained.

We compare our method with filtered back projection (FBP),
Tikhonov regularization, and total variation (TV).
The FBP solution is computed using MATLAB's $\mathtt{iradon}$ function with
the ``Shepp-Logan'' filter.
The Tikhonov solution solves the problem
  \[
    \min_x \|Ax-b\|_2^2+\lambda_{\mathrm{Tikh}}\|x\|^2_2,
  \]
where $\lambda_{\mathrm{Tikh}}$ is the Tikhonov regularization parameter.
The TV regularization problem has the form
  \begin{equation*}
    \min \ \nicefrac{1}{2}\, \|A\, x-b\|_2^2 + \lambda_{\mathrm{TV}}
    \sum_{i=1}^n \bigl\| D^{\mathrm{fd}}_i x \bigr\|_2 , \qquad
    0 \leq x \leq 1 ,
  \end{equation*}
where $D^{\mathrm{fd}}_i$
computes a finite-difference approximation of the gradient at each pixel,
and $\lambda_{\mathrm{TV}}$ is the TV regularization parameter.
We solve this problem with the software \textsc{TVReg} \cite{Jensen}.
The Tikhonov and TV regularization parameters are chosen to yield the smallest
reconstruction error.

The computational bottleneck of the objective function evaluation in solving
\eqref{eq:trec} is calculating $\mathcal{D}*\mathcal{C}$, where
$\mathcal{D} \in \mathbb{R}^{p \times s \times r}$
and $\mathcal{C} \in \mathbb{R}^{s \times q \times r}$.
Recall that the computation is done in the Fourier domain, and since
$\log(r) < q,p $ the computational complexity of the t-product is
$O(sqpr+s(p+q)r\log(r)) = O(sqpr)$ \cite{Hao}.
In the matrix formulation \cite{Soltani} the computational bottleneck
is the matrix multiplication $D \, \mathtt{reshape}\big(\alpha,s,q)$
where $D \in \mathbb{R}^{pr \times s}$ and $\alpha \in \mathbb{R}^{sq\times 1}$,
also with complexity~$O(sqpr)$.
This gives the tensor formulation an advantage, since we can use a much smaller
$s$ here, say, 2--3 times smaller than in the matrix formulation.

Since  computation times vary between different computers, and since we
did not pay specific attention to efficiency, we report the number of objective function
evaluations returned by TFOCS\@.
We stop the iterations when the relative
change in the iteration vector is less than $10^{-7}$.
For the comparison to be fair, the starting point in all the computations
is the zero vector/matrix of appropriate size.

\subsubsection{Study of Regularization Terms}
\label{sec:expr1}

We solve the reconstruction problem using the exact image shown in Fig. \ref{fig:Images}.
Moreover, we use $10\times10$ patches, $s = 300$, and $\lambda = 3.1623$.
For the problems in this section we use $N_{\mathrm{p}} = 25$ projections,
$N_{\mathrm{r}} = 283$ rays per projection, and 1\% noise.
We compare two different regularization terms in the reconstruction problem
\eqref{eq:trec}.
The 1-norm (sparsity) regularization $\| \mathcal{C} \|_{\mathrm{sum}}$
is similar to the 1-norm regularization in the dictionary learning
problem~\eqref{eq:diclear}.
The regularization term $\| \mathcal{C} \|_{\mathrm{sum}} + \| C \|_{*}$
results in coefficient tensors that are simultaneously low rank and sparse.

\begin{figure}%[htp]
\centering
\subfigure[FBP]{ \includegraphics[width=0.3\linewidth]{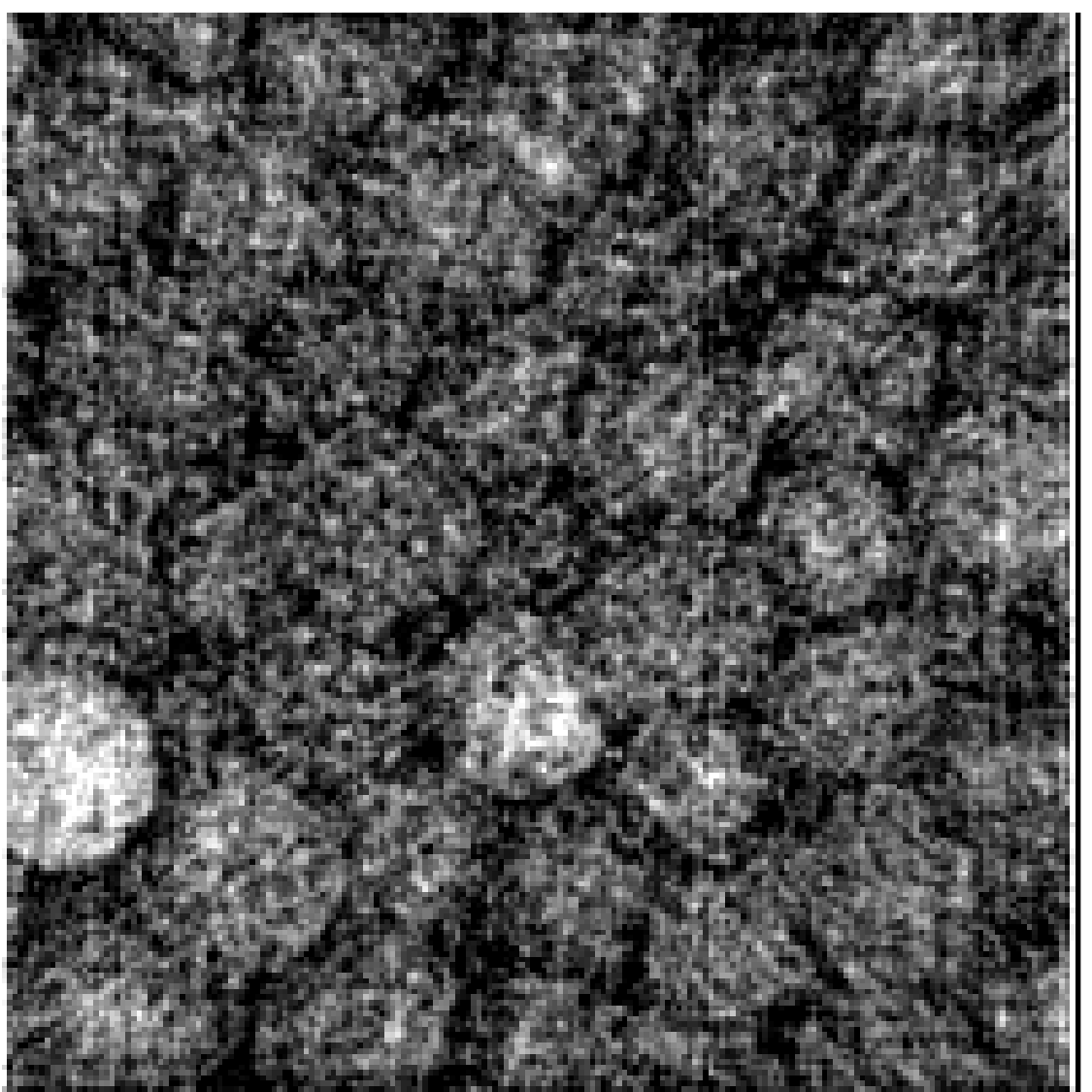}}
\subfigure[Tikhonov]{ \includegraphics[width=0.3\linewidth]{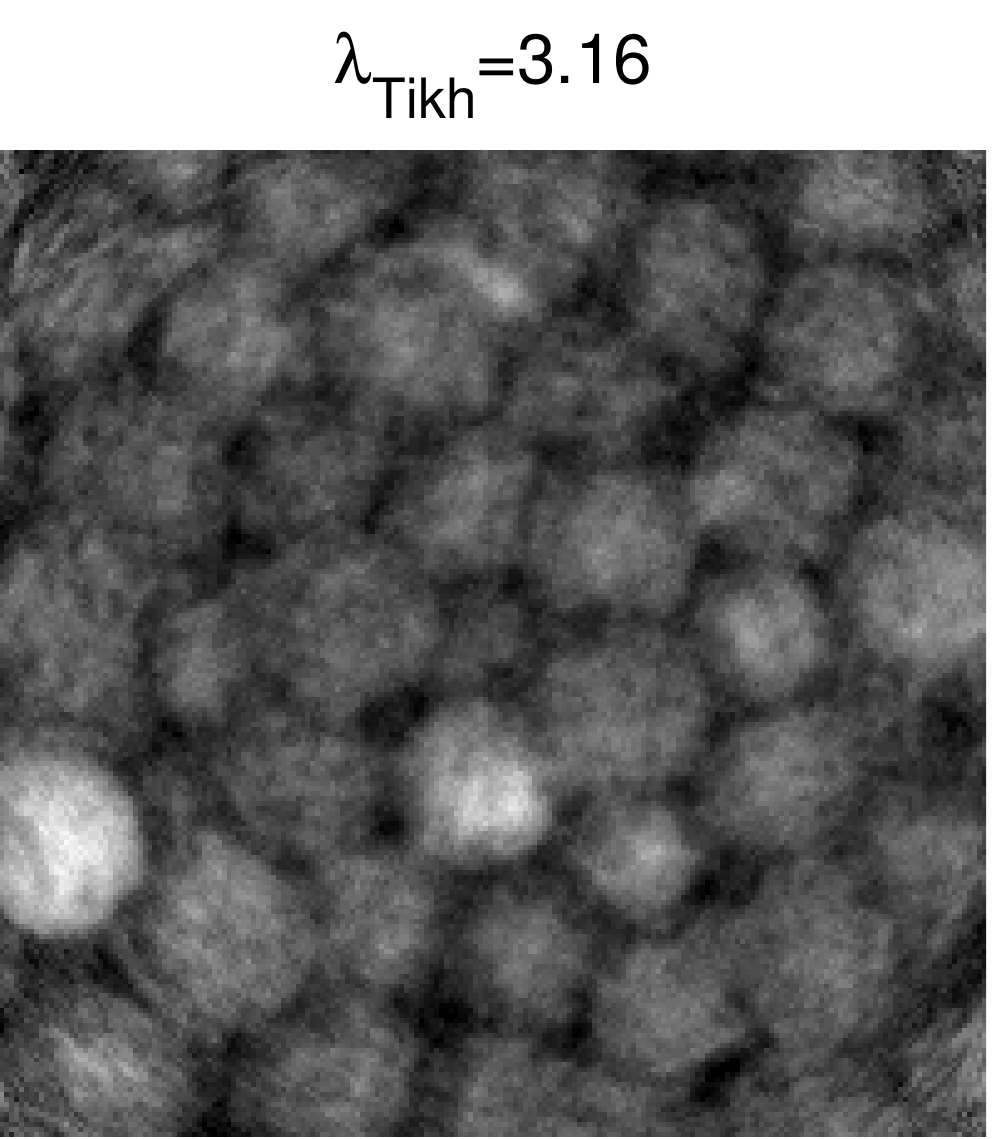}}
\subfigure[TV]{ \includegraphics[width=0.3\linewidth]{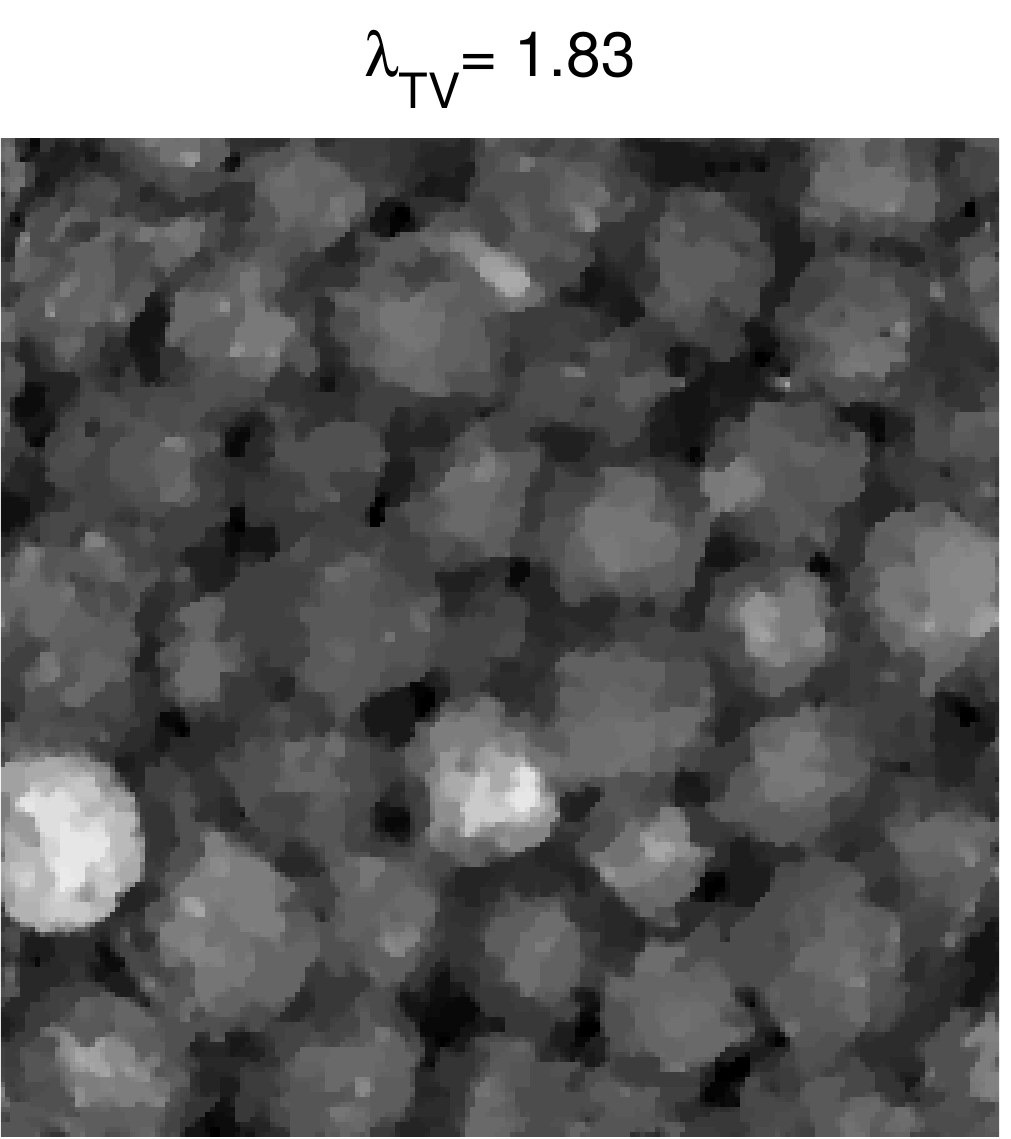}}\\
\subfigure[Matrix formulation]{ \includegraphics[width=0.3\linewidth]{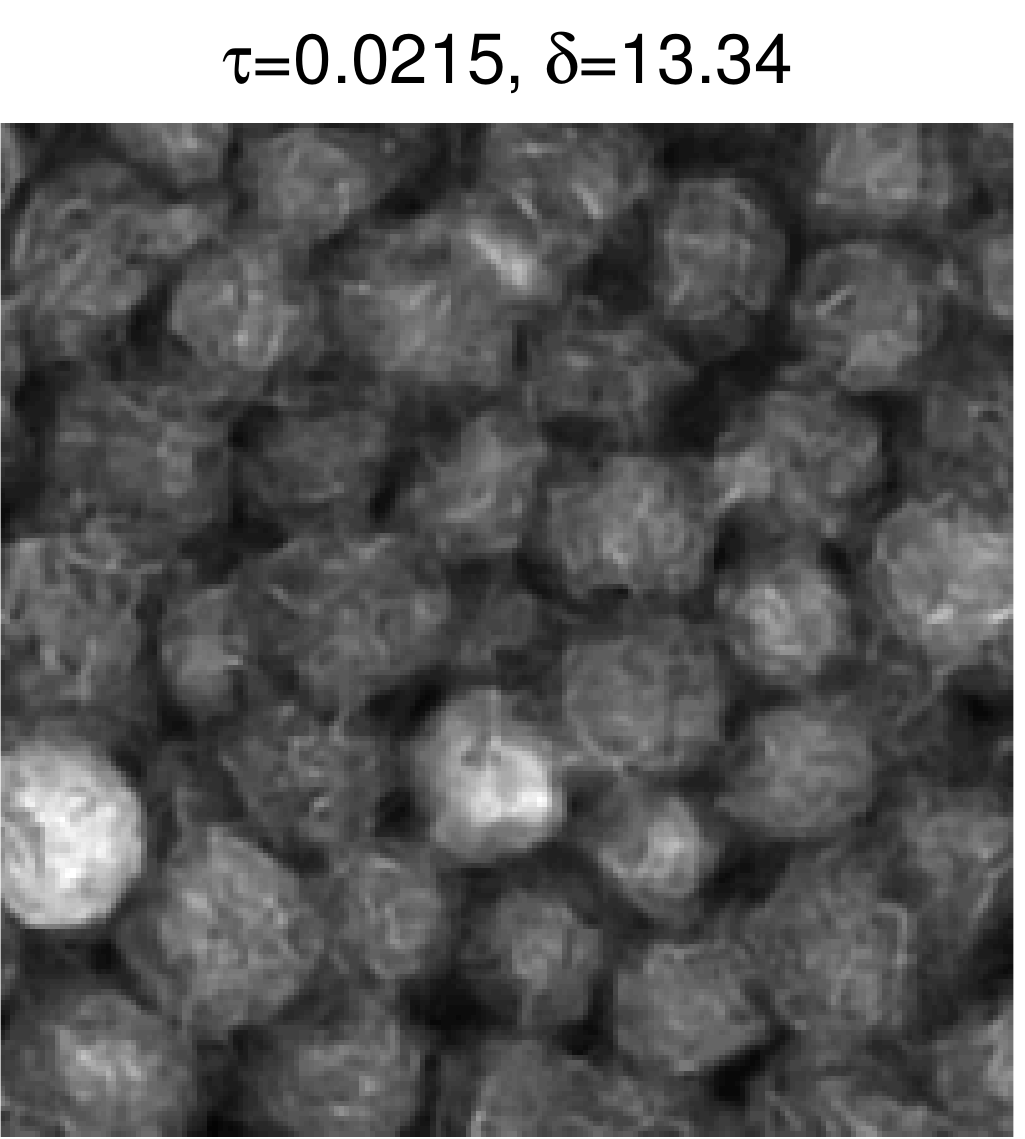}}
\subfigure[Tensor:\ $\|\mathcal{C}\|_{\mathrm{sum}}$]{ \includegraphics[width=0.3\linewidth]{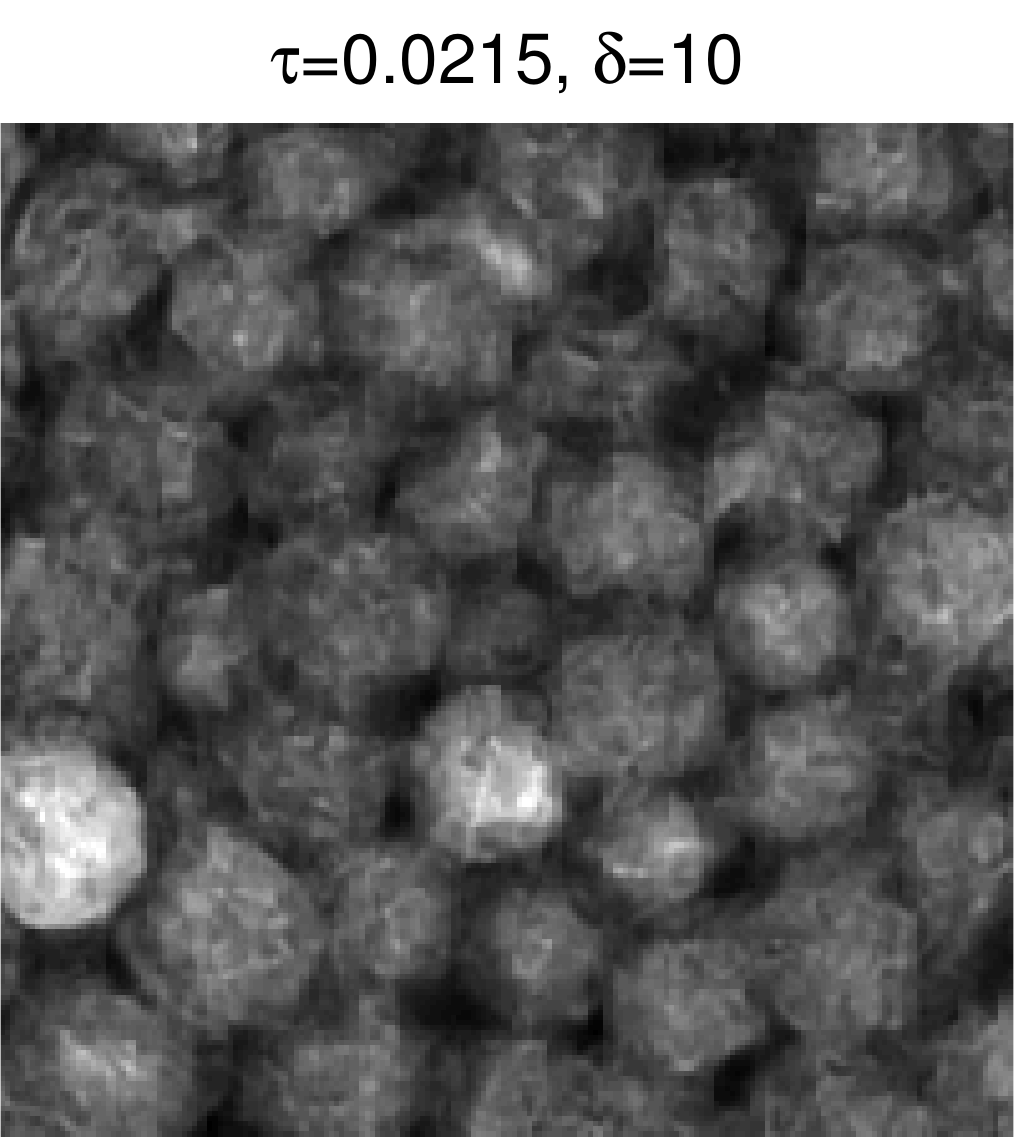}}
\subfigure[Tensor:\ $\|\mathcal{C}\|_{\mathrm{sum}}+\|C\|_{*}$]{ \includegraphics[width=0.3\linewidth]{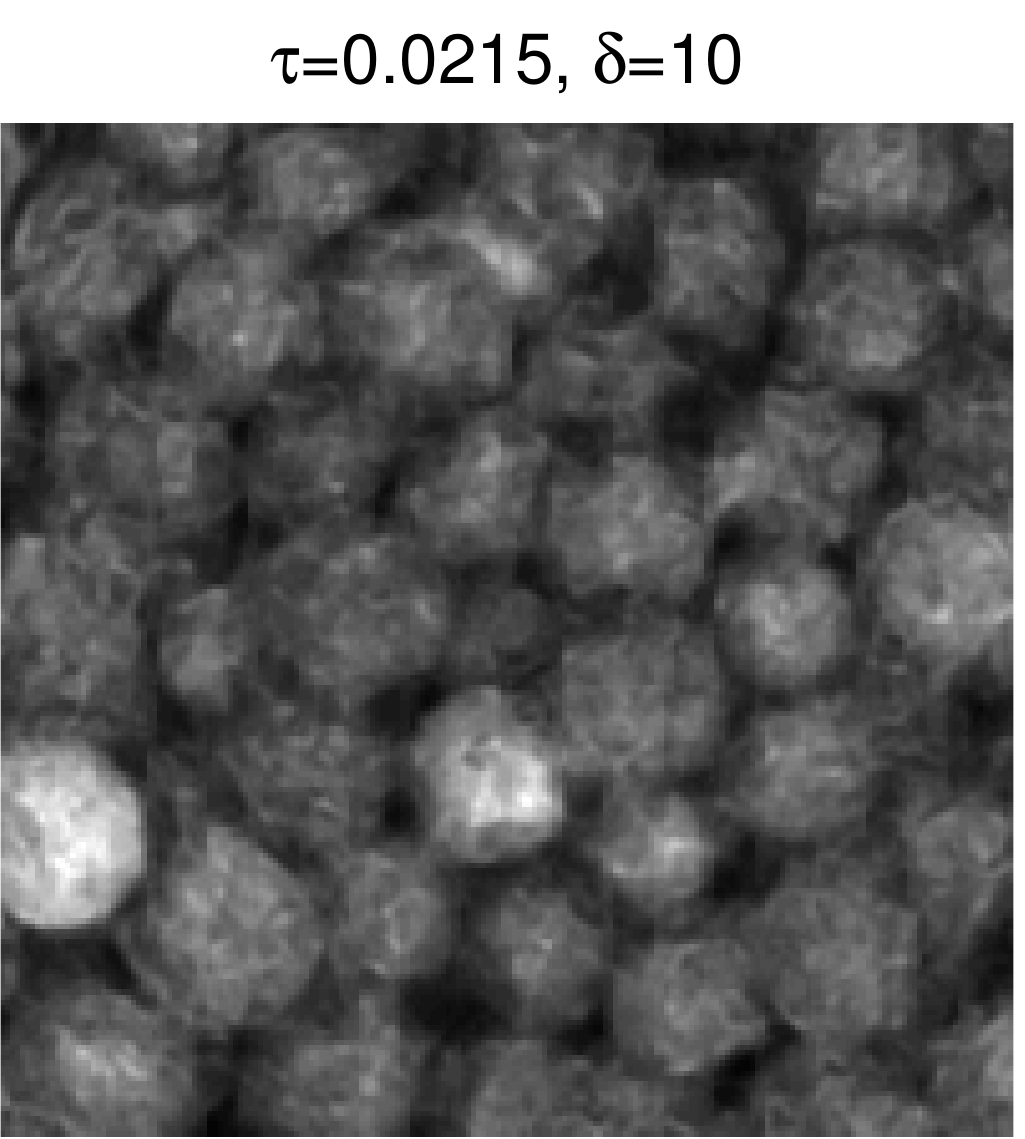}}
\caption{Comparison of the best solutions computed by different reconstruction methods.
Subfigures (e) and (f) correspond to our new tensor formulation with two different
regularization terms; we used $\lambda=3.1623$.}
\label{fig:ImRes1}
\end{figure}

\begin{table}%[htp]
\caption{Comparison of the best solutions computed by different reconstruction methods.
The bold numbers indicate the lowest iteration number, density of $\mathcal{C}$
and compression percentages, and highest SSIM measure.}
\label{tab:1}
\begin{tabular}{llllll}
\hline\noalign{\smallskip}
Method & Itr.$\#$ & Density$\%$ & Compr.\% & RE\% & SSIM  \\
\hline
FBP:\ filtered back projection & - & - & - & 54.81 & 0.2981 \\
Tikhonov regularization:\ & - & - & - & 21.99 & 0.5010 \\
TV:\ total variation & - & - & - & 21.37 & 0.4953 \\
Matrix formulation \cite{Soltani} & 36843 & 12.53 & 5.31 & 22.00 & 0.4903 \\
Tensor alg., $\|\mathcal{C}\|_{\mathrm{sum}}$ reg. & 48787 &  {\bf 5.30} & {\bf 0.67} &
  22.21 & 0.4890 \\
Tensor alg., $\|\mathcal{C}\|_{\mathrm{sum}}+\|C\|_*$ reg. & {\bf 8002} & 10.27 &
  3.26 & {\bf 21.55}  & {\bf 0.5061} \\
\hline
\end{tabular}
\end{table}

We compare the tensor reconstruction solution with the solutions obtained
by the matrix formulation \cite{Soltani} as well as FBP, Tikhonov regularization,
and TV\@.
The reconstructions are shown in Fig.~\ref{fig:ImRes1}.
The corresponding relative errors, SSIM, and densities of $\mathcal{C}$
as well as the number of objective function evaluation are listed in Table \ref{tab:1}.
The table also lists the compressibility, defined as the percentage of coefficients
which have values larger than $10^{-4}$.
Both the density and the compressibility show that we obtain very
sparse representations of the reconstructed image.

The FBP, Tikhonov, and TV methods fail to produce desirable reconstructions,
although the 2-norm reconstruction error for the TV solution is slightly smaller than
that for our solutions.
The RE and SSIM do not tell the full story, and
using a dictionary clearly improves recovering the \emph{texture} of the image.
The reconstructed images in Fig.~\ref{fig:ImRes1} are similar across the
matrix and tensor formulations;
however, the results in Table \ref{tab:1} show that the tensor-formulation solution
is more than 5 times more compressed and more than 2 times sparser than
the matrix-formulation solution.
Imposing both sparsity and low-rank regularization
$\|\cdot\|_{\mathrm{sum}}+\|\cdot\|_*$ produces a marginally more accurate solution
with a denser representation.

\subsubsection{More Challenging Tomographic Reconstructions}
\label{sec:expr2}

\begin{figure}%[htp]
\begin{minipage}{0.25\textwidth}
  {\scriptsize (a) $N_{\mathrm{p}}=50$ \\ angles in $[0^{\circ},180^{\circ}]$ \\
  noise $1\%$}
\end{minipage}
\begin{minipage}{0.75\textwidth}
  \subfigure{\includegraphics[width=0.32\linewidth]{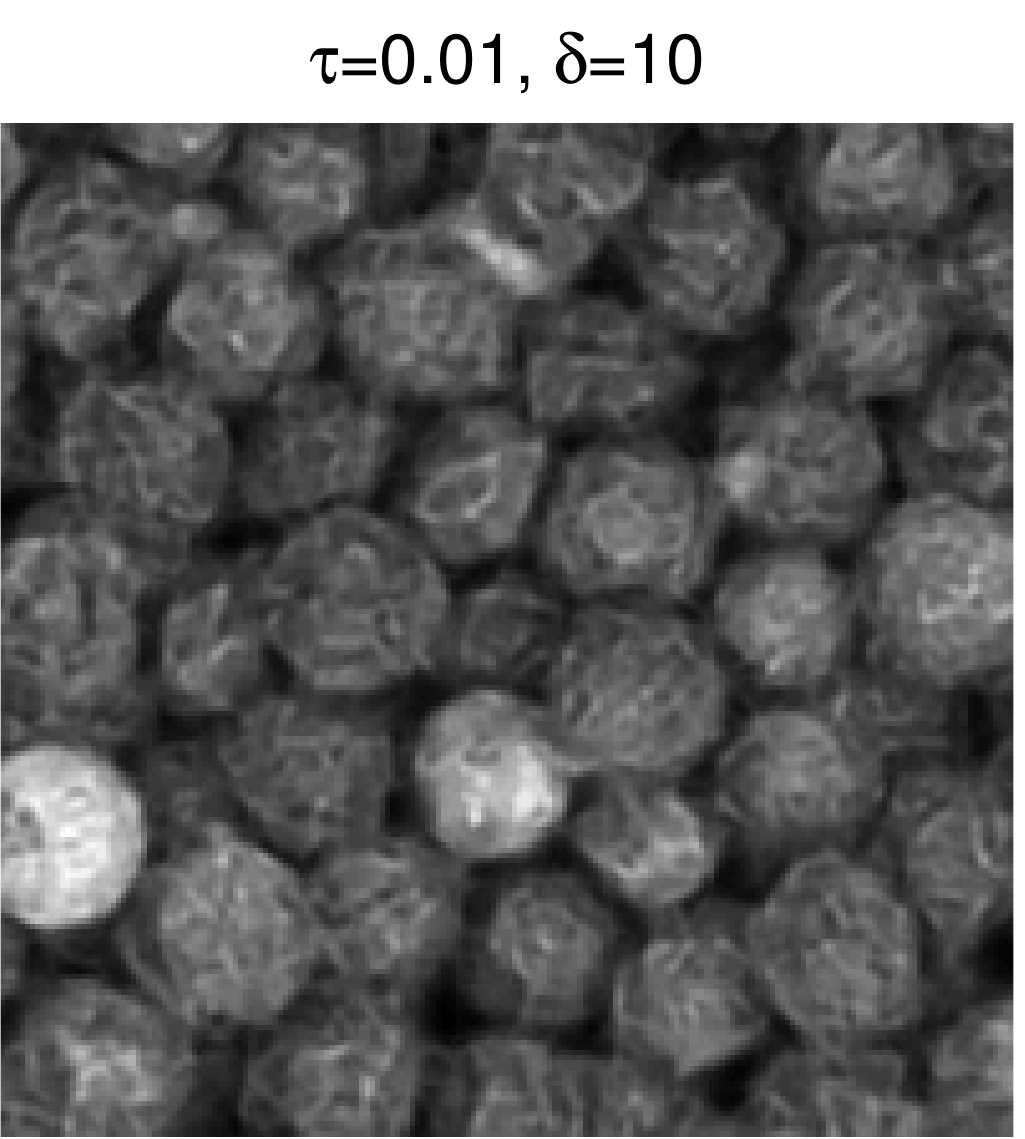}}
  \subfigure{\includegraphics[width=0.32\linewidth]{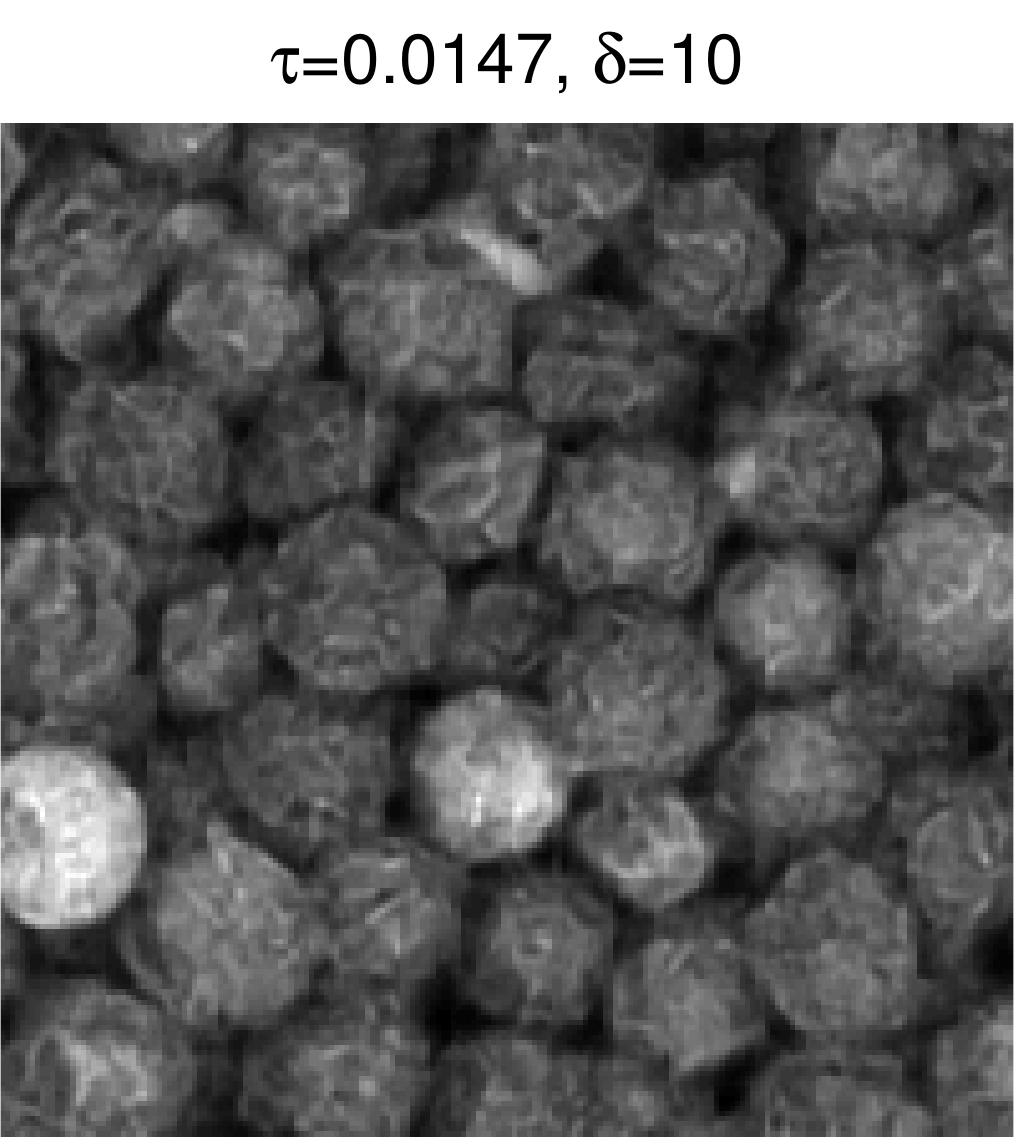}}
  \subfigure{\includegraphics[width=0.32\linewidth]{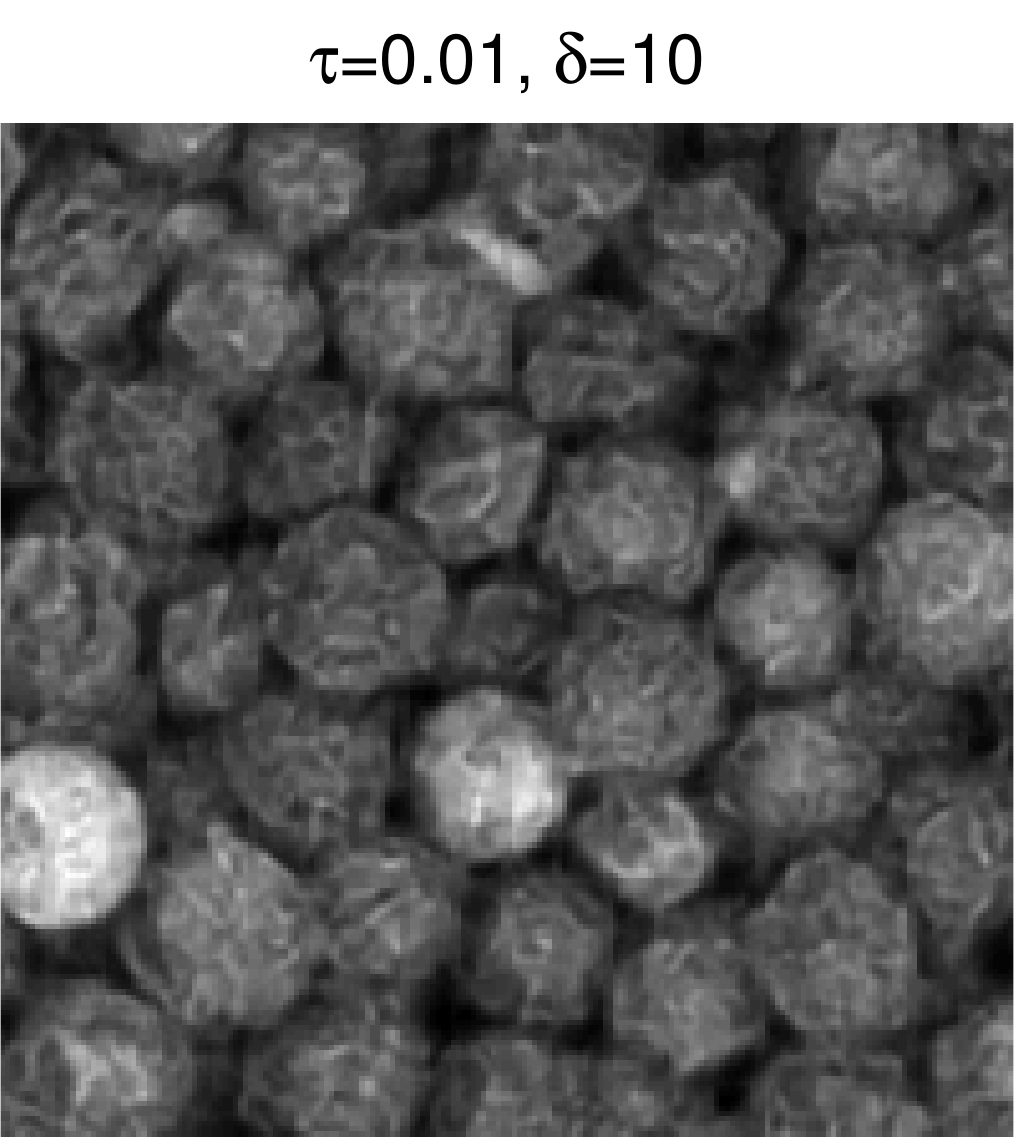}}
\end{minipage}
\begin{minipage}{0.25\textwidth}
  {\scriptsize (b) $N_{\mathrm{p}}=50$ \\ angles in $[0^{\circ},120^{\circ}]$ \\
  noise $1\%$}
\end{minipage}
\begin{minipage}{0.75\textwidth}
  \subfigure{\includegraphics[width=0.32\linewidth]{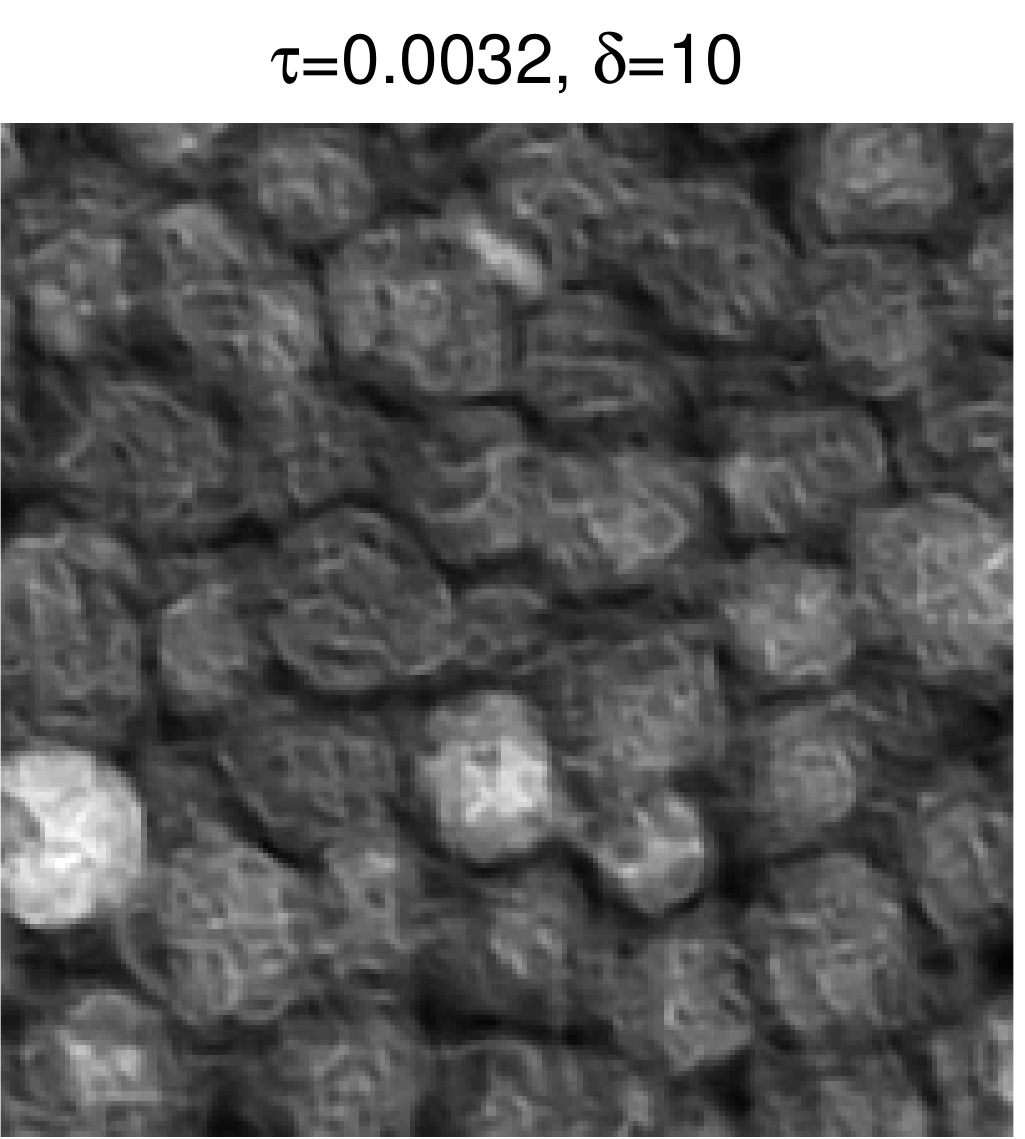}}
  \subfigure{\includegraphics[width=0.32\linewidth]{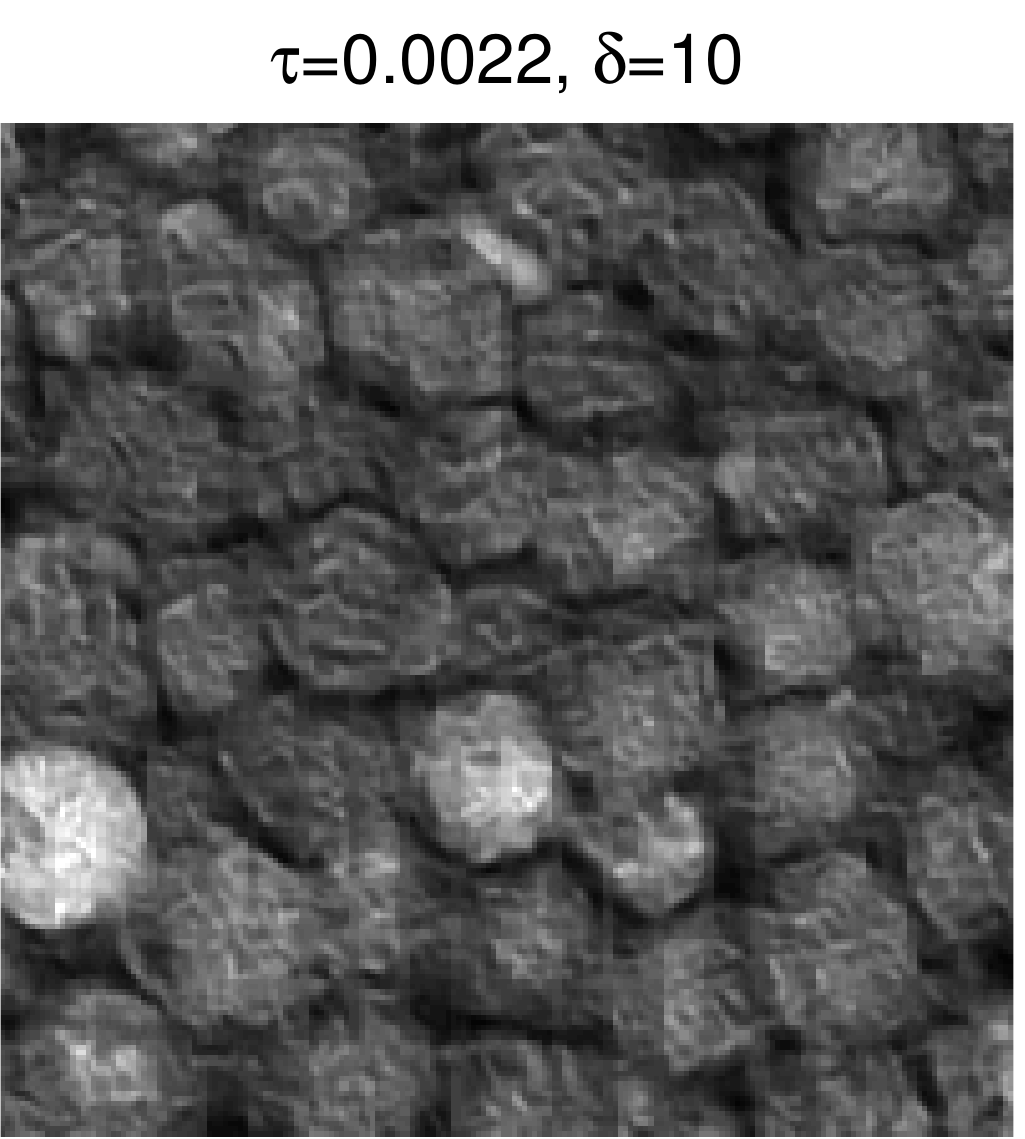}}
  \subfigure{\includegraphics[width=0.32\linewidth]{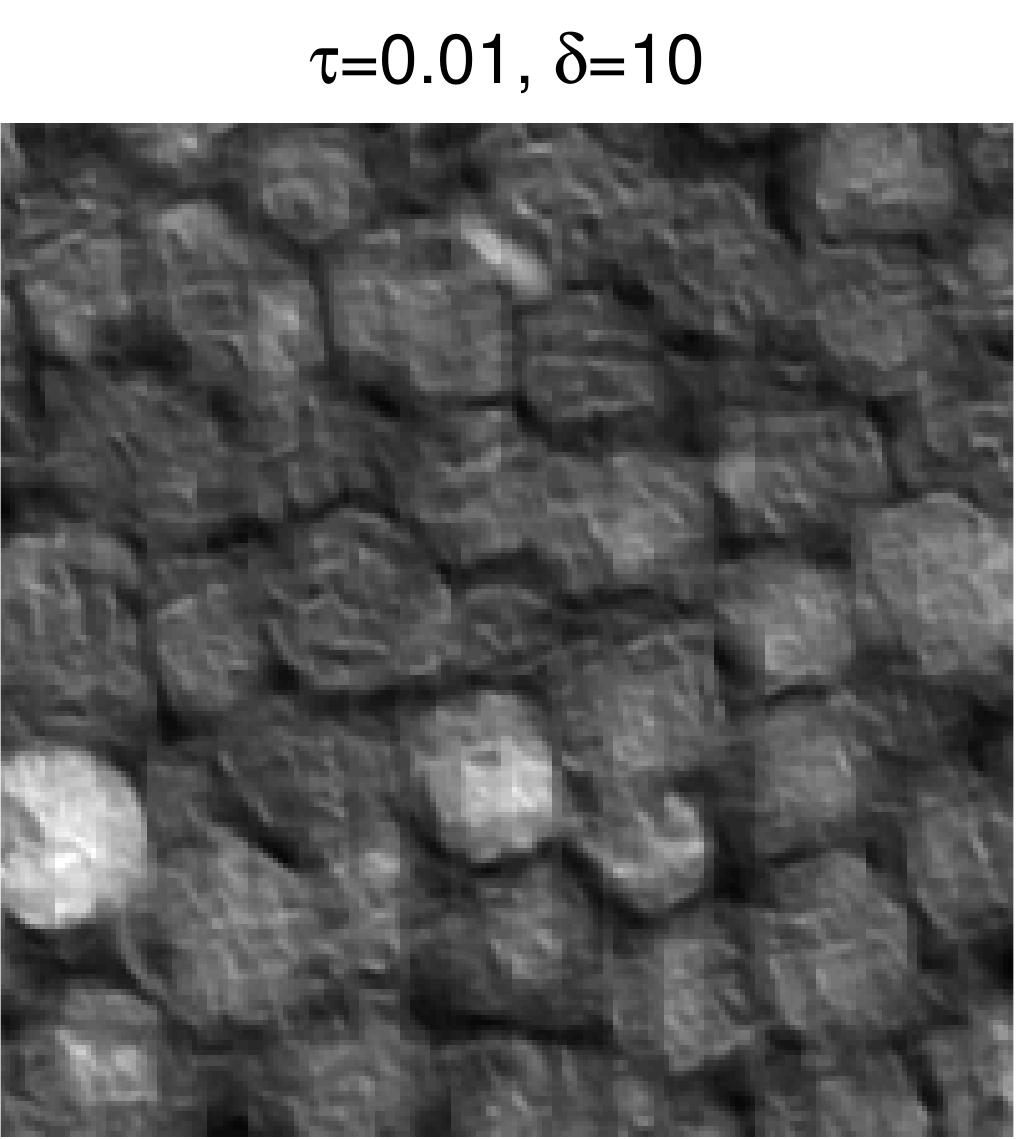}}
\end{minipage}
\begin{minipage}{0.25\textwidth}
  {\scriptsize (c) $N_{\mathrm{p}}=25$ \\ angles in $[0^{\circ},120^{\circ}]$ \\
  noise $1\%$}
\end{minipage}
\begin{minipage}{0.75\textwidth}
  \subfigure{\includegraphics[width=0.32\linewidth]{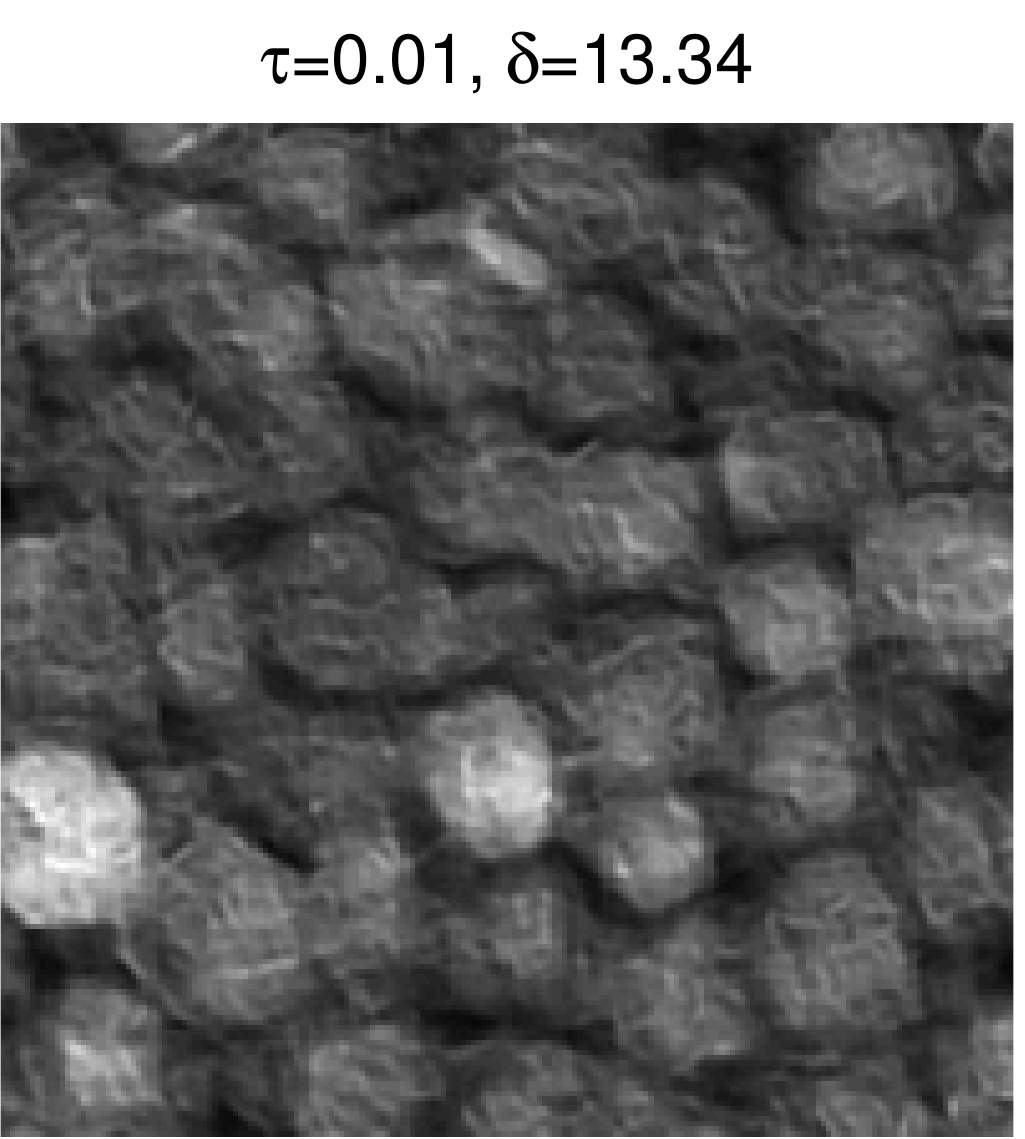}}
  \subfigure{\includegraphics[width=0.32\linewidth]{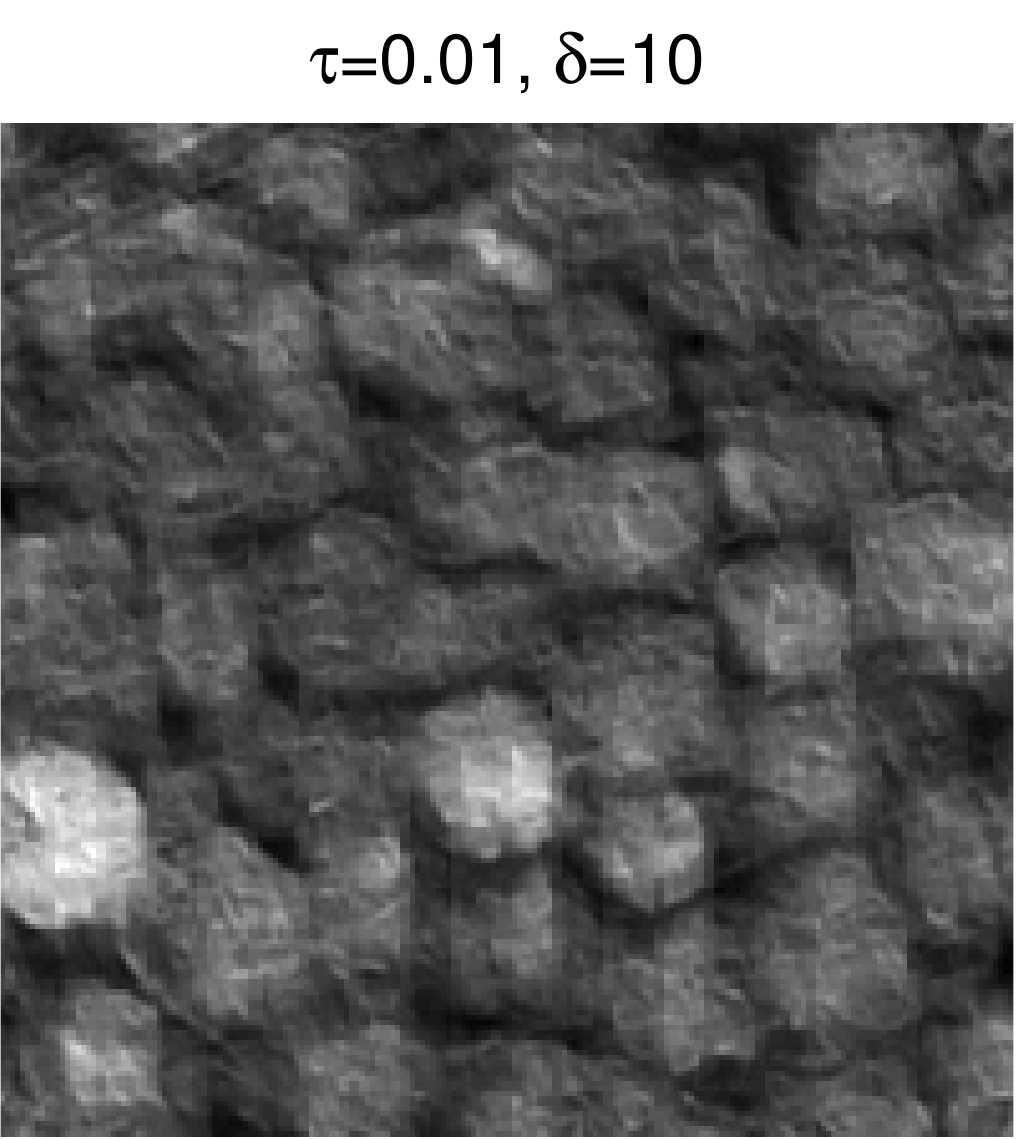}}
  \subfigure{\includegraphics[width=0.32\linewidth]{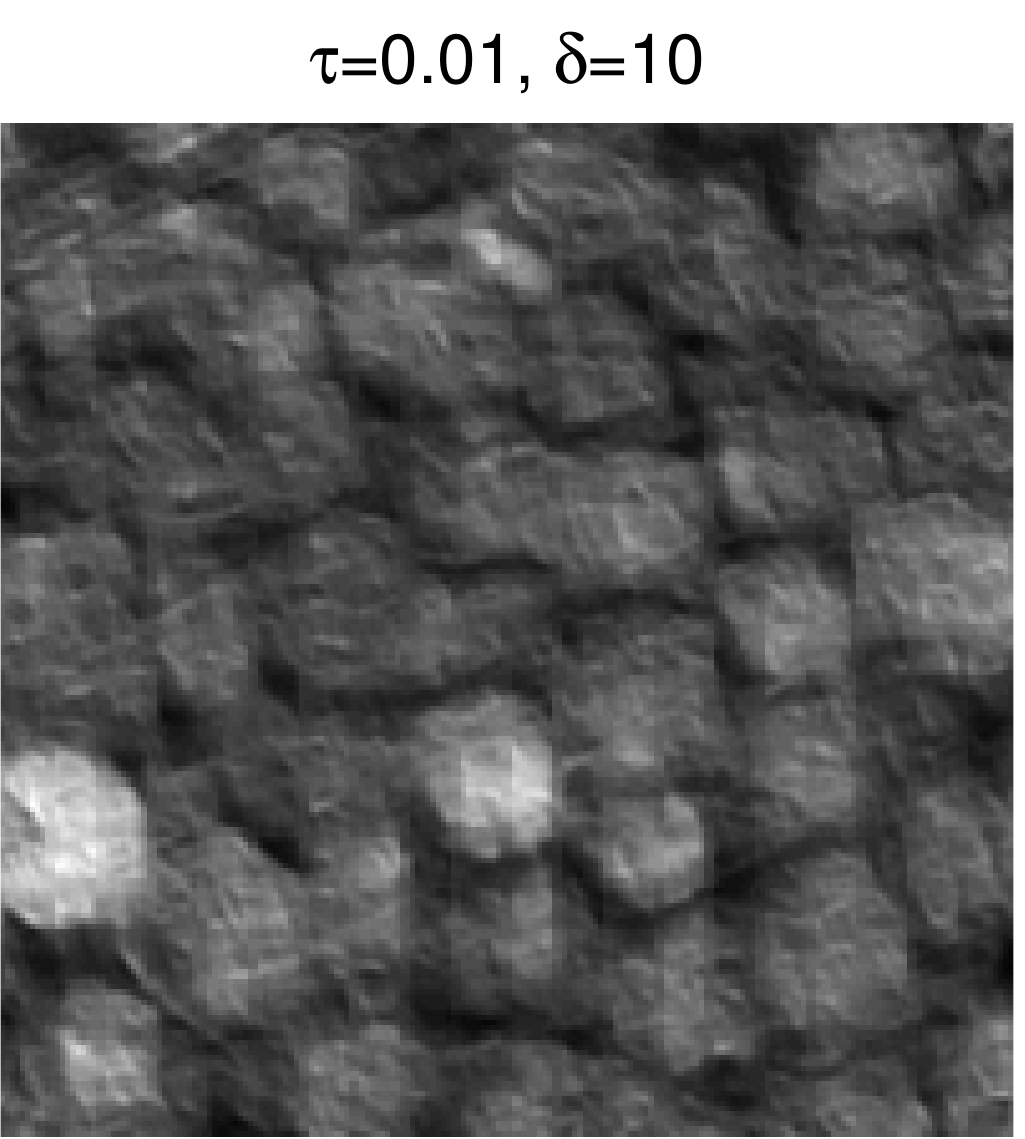}}
\end{minipage}
\begin{minipage}{0.25\textwidth}
  {\scriptsize (d) $N_{\mathrm{p}}=50$ \\ angles in $[0^{\circ},180^{\circ}]$
  noise $5\%$}
\end{minipage}
\begin{minipage}{0.75\textwidth}
  \subfigure{\includegraphics[width=0.32\linewidth]{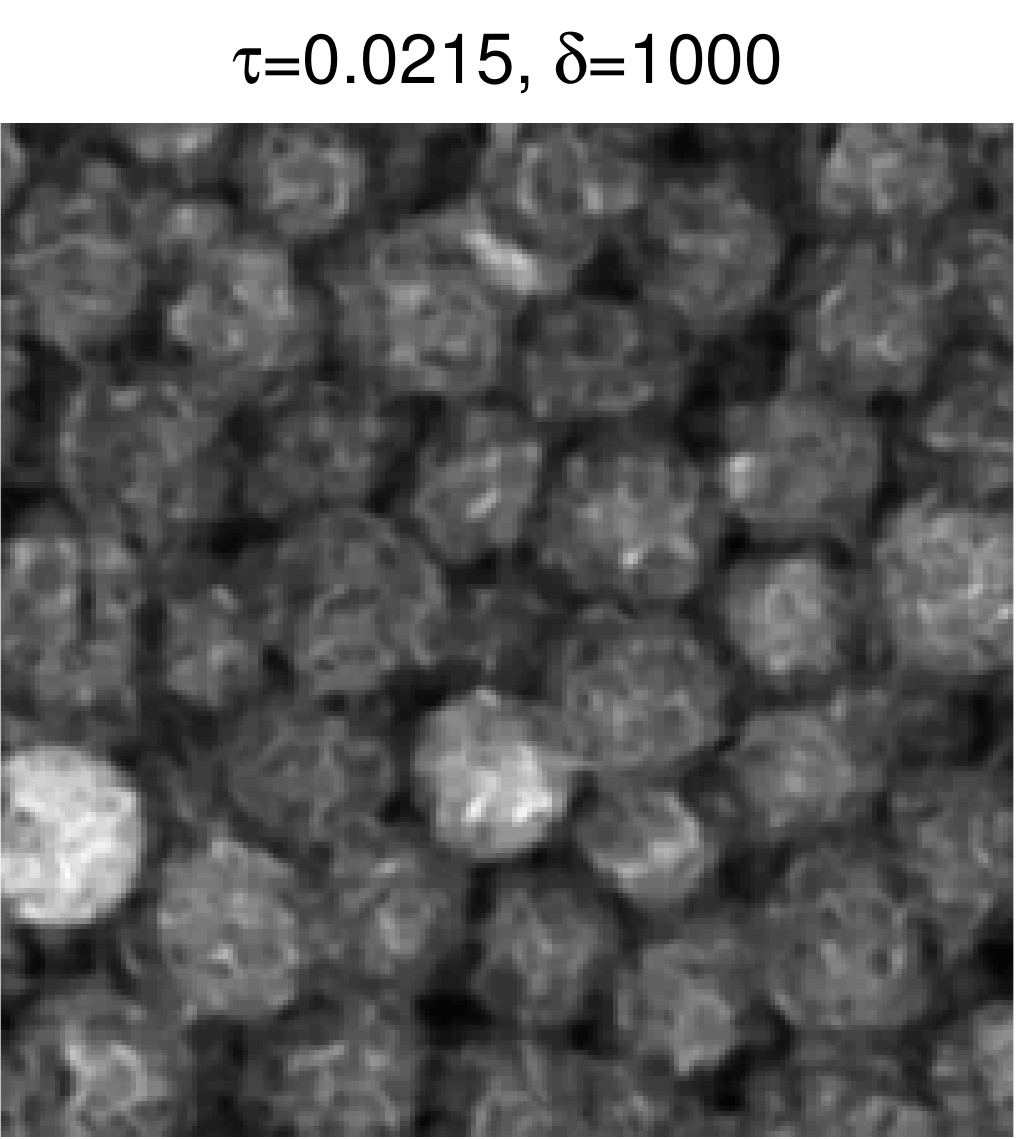}}
  \subfigure{\includegraphics[width=0.32\linewidth]{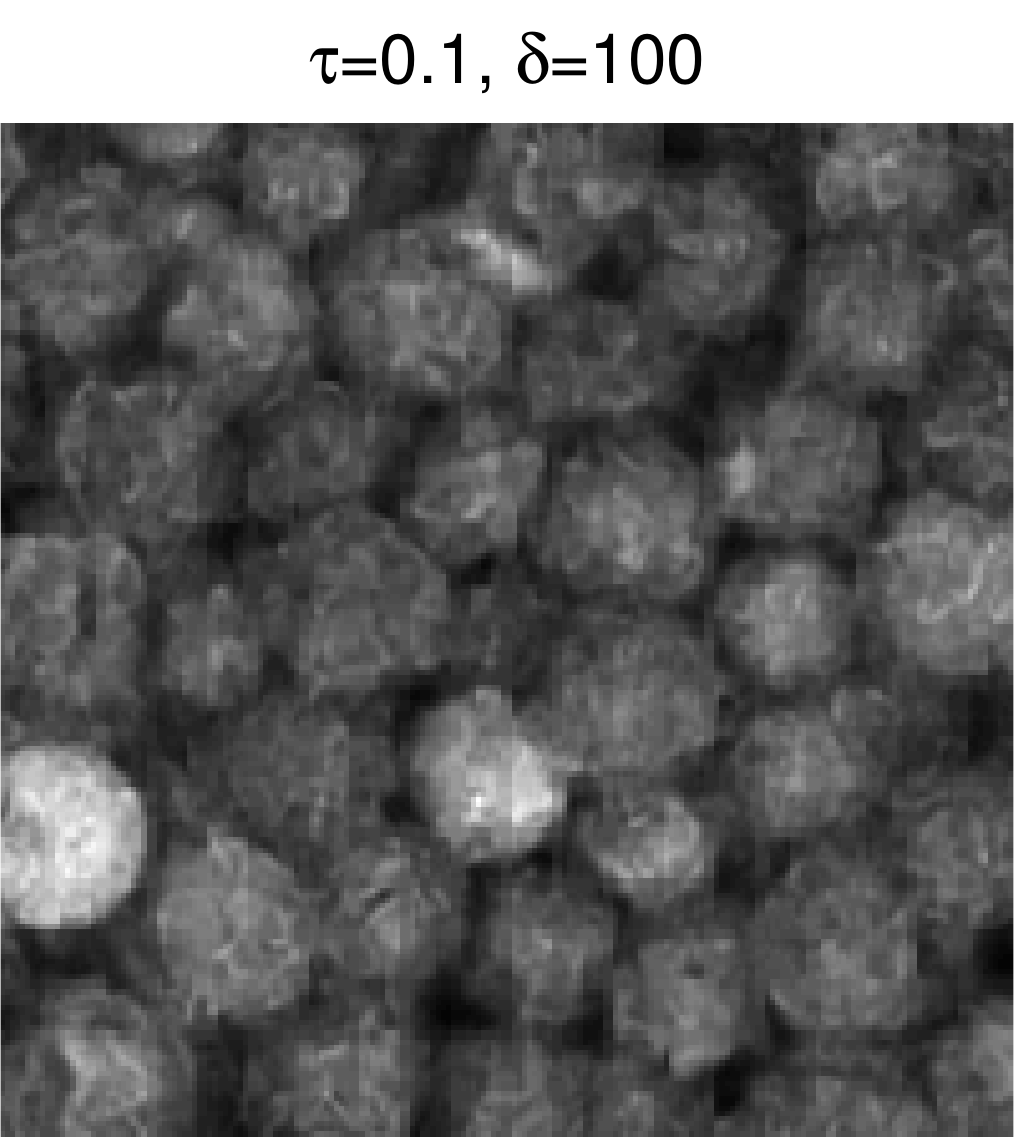}}
  \subfigure{\includegraphics[width=0.32\linewidth]{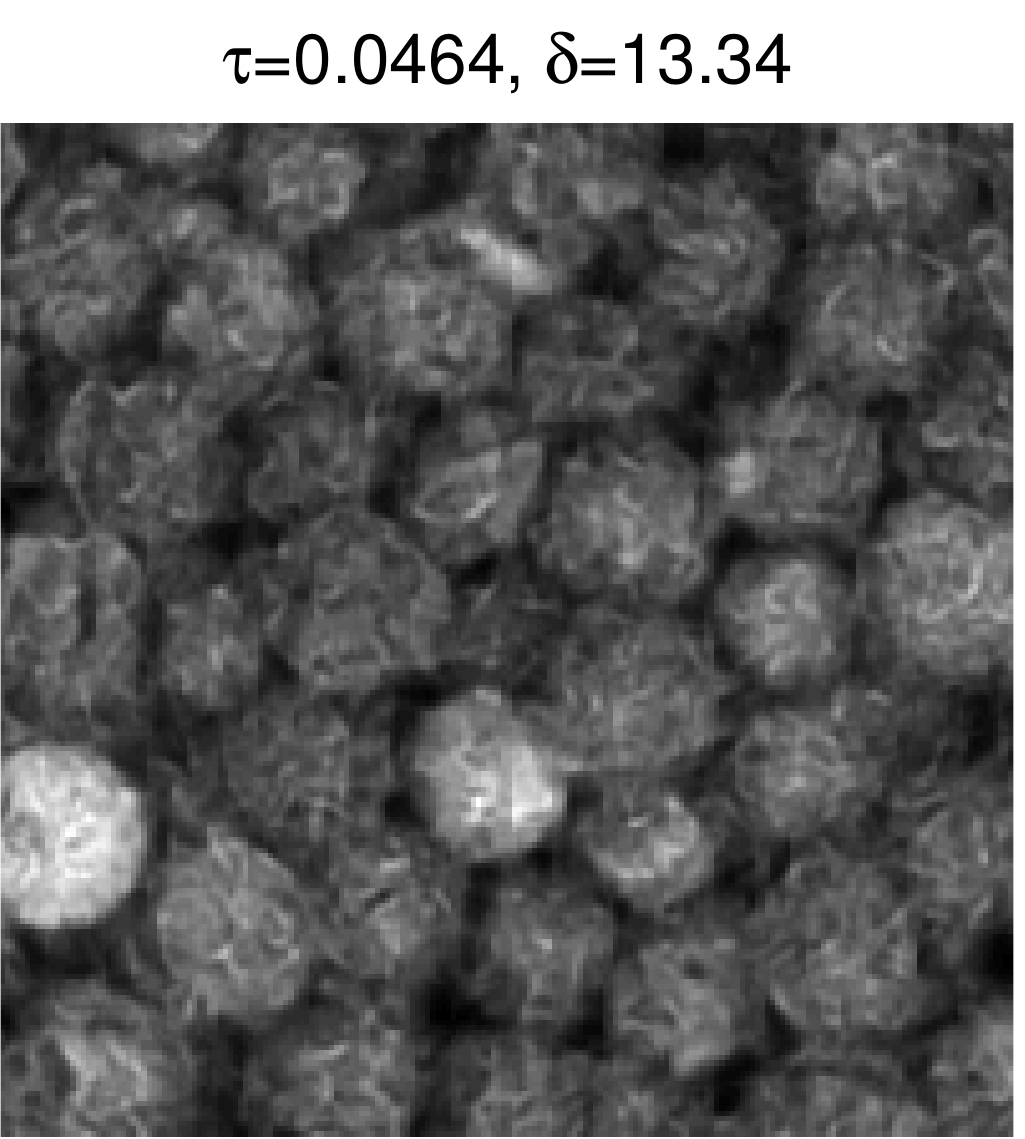}}
\end{minipage}
\begin{minipage}{0.25\textwidth}
  {\scriptsize (e) $N_{\mathrm{p}}=25$ \\ angles in $[0^{\circ},180^{\circ}]$
  noise $5\%$}
\end{minipage}
\begin{minipage}{0.75\textwidth}
  \subfigure{\includegraphics[width=0.32\linewidth]{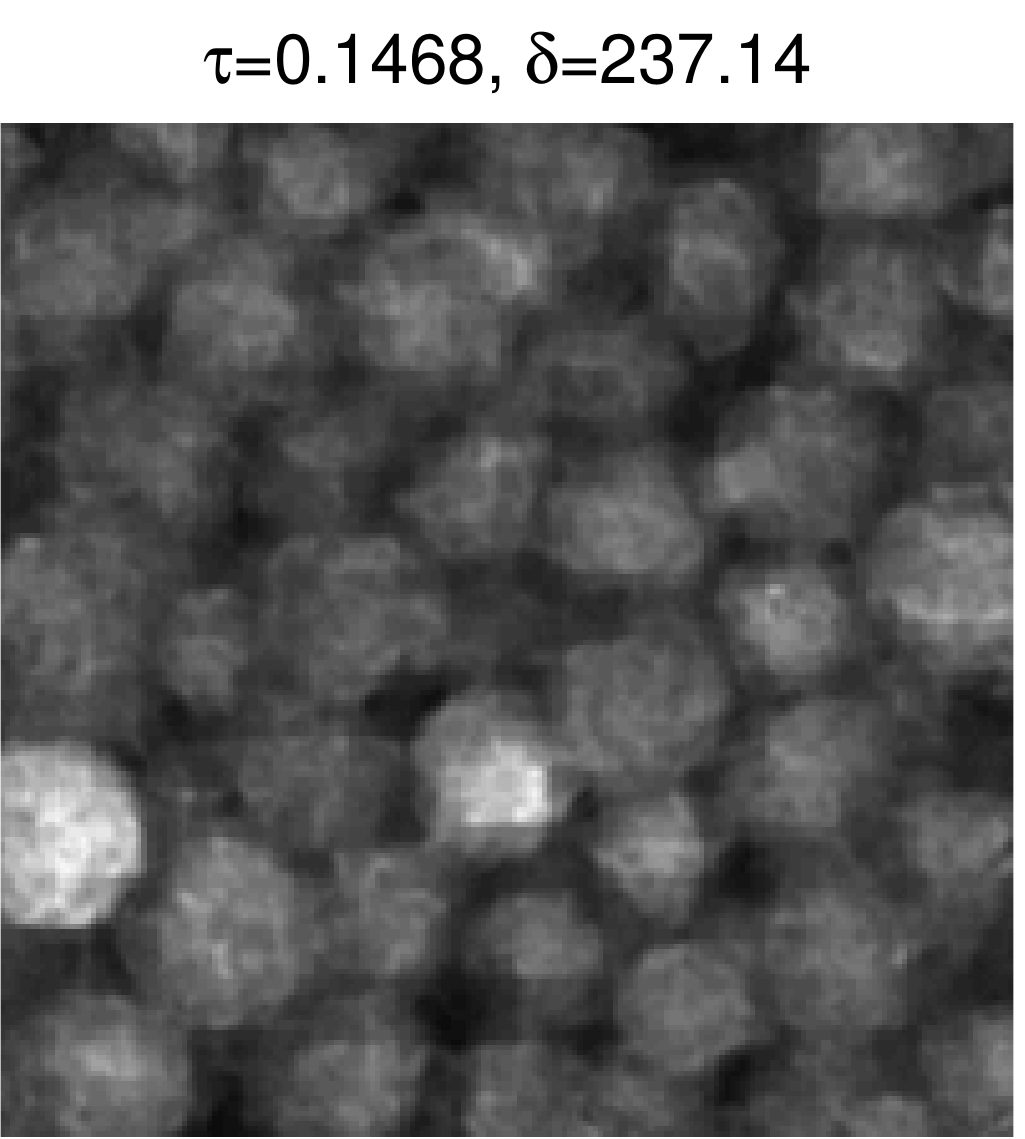}}
  \subfigure{\includegraphics[width=0.32\linewidth]{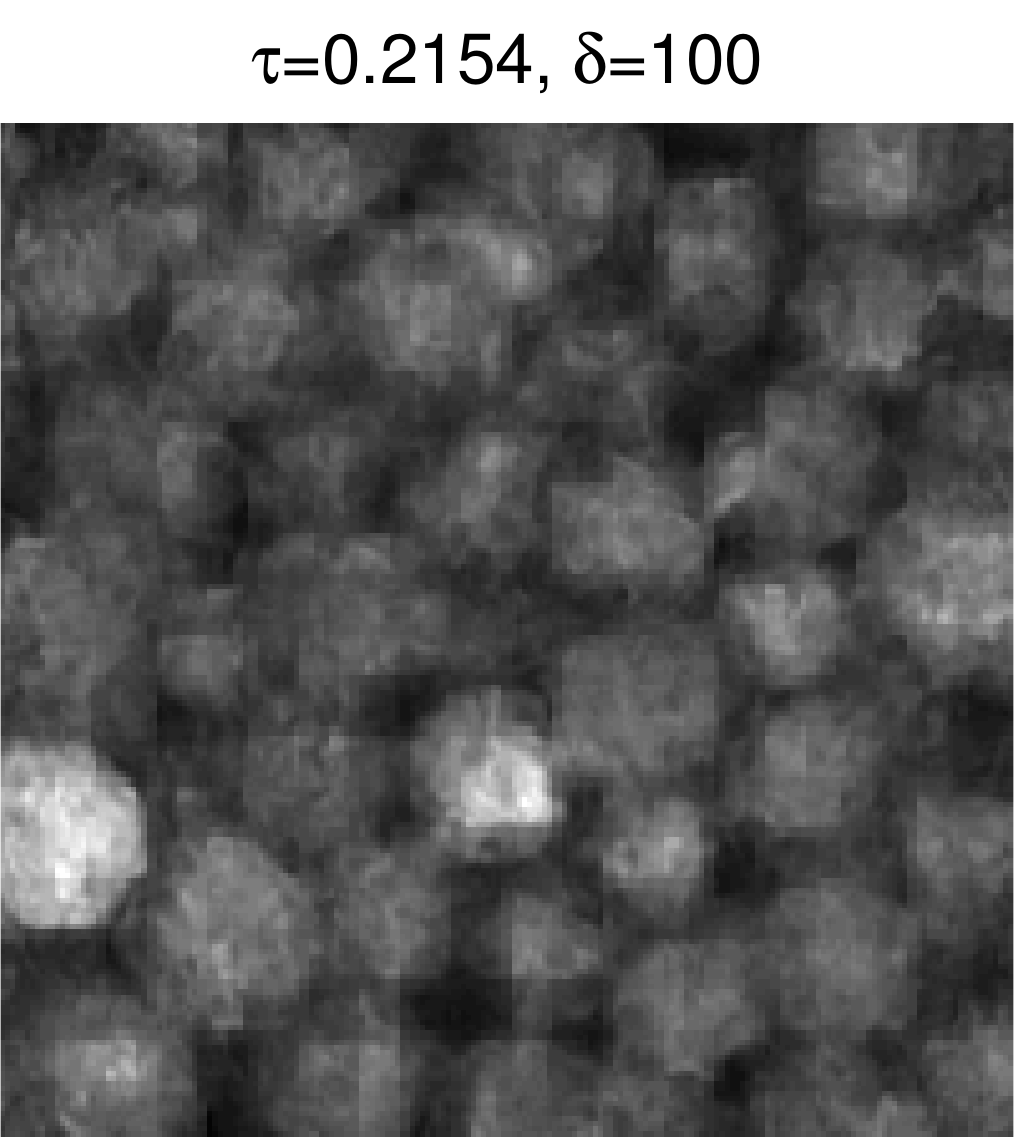}}
  \subfigure{\includegraphics[width=0.32\linewidth]{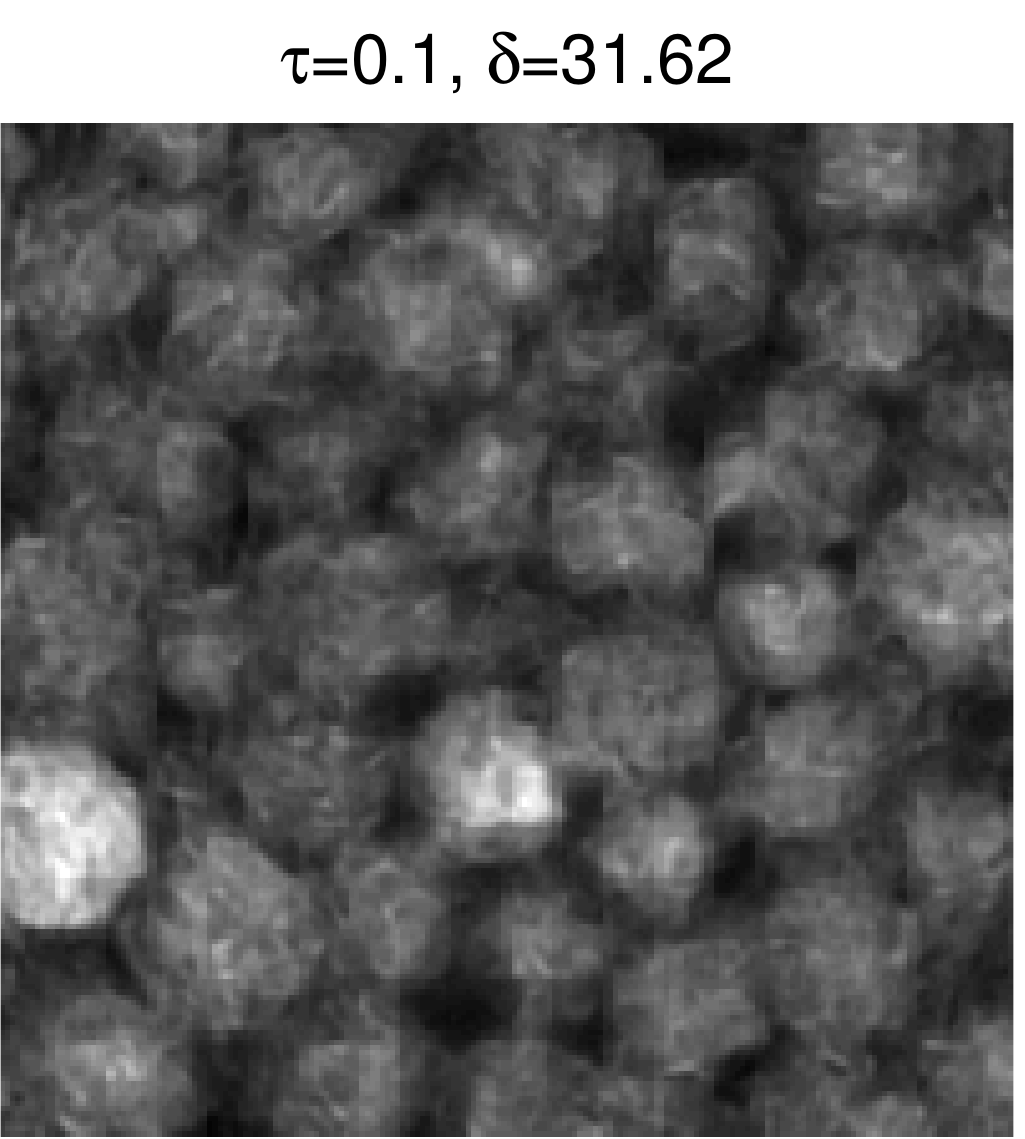}} \\
\vspace{12mm} {\footnotesize \quad \, \quad Matrix alg. \qquad \quad \,  \quad Tensor alg.
  \qquad \quad \,  \quad Tensor alg.} \\[-11mm]
\vspace{12mm} {\footnotesize \hspace{36mm} $\|\mathcal{C}\|_{\mathrm{sum}}$ reg.
  \quad \quad \ $\|\mathcal{C}\|_{\mathrm{sum}}+\|C\|_*$ reg.}
\end{minipage}
\caption{Reconstruction experiments from Section \ref{sec:expr2} with $\lambda=3.1623$.}
\label{fig:ImRes2}
\end{figure}

To further study the performance and robustness of our tensor formulation approach,
we consider problems with more noise in the data, or with
projection angles in a limited range, still using the same test problem.
Knowing that FBP, Tikhonov, and TV give unsatisfactory solutions
for such problems,
we only compare our method with the matrix formulation approach,
and again we consider both regularization terms $\|\mathcal{C}\|_{\mathrm{sum}}$ and
$\|\mathcal{C}\|_{\mathrm{sum}}+\|C\|_*$ in \eqref{eq:trec}.
\begin{itemize}
\item
First we compute a reconstruction with $N_{\mathrm{p}}=50$ projections,
uniform angular sampling in $[0^{\circ},180^{\circ}]$
and with relative noise level $1\%$.
In this scenario we use more projection data than in the previous section.
\item
Next we use $50$ and $25$ projections uniformly distributed
in the limited range $[0^{\circ}, 120^{\circ}]$ and with relative noise level $1\%$.
\item
Finally we use $25$ and $50$ projections with uniform angular sampling
in $[0^{\circ},180^{\circ}]$ and with relative noise level $5\%$, i.e.,
a higher noise level than above.
\end{itemize}
The reconstructions are shown in Fig.~\ref{fig:ImRes2}; they
are similar across the tensor and matrix formulations,
and pronounced artifacts have appeared from the limited angles and the higher noise level.

\begin{table}[htp]
\caption{Comparison of tensor and matrix formulation reconstructions
in the experiments from Section \ref{sec:expr2}.
The methods ``Matrix,'' ``Tensor-1,'' and ``Tensor-2'' refer to the matrix-formulation
algorithm and our new tensor-formulation algorithm with regularization terms
$\|\mathcal{C}\|_{\mathrm{sum}}$ and $\|\mathcal{C}\|_{\mathrm{sum}}+\|C\|_*$.
The bold numbers indicate the lowest iteration number, density and
compression, and the highest SSIM.}
\label{tab:2}
\begin{tabular}{l|llllll}
\hline\noalign{\smallskip}
 Settings & Method & Itr.\# & Density\% & Compr.\% & RE\%  & SSIM  \\
\hline
$N_{\mathrm{p}}=50$ & Matrix & 41204  & 20.70 & 8.80 & 17.70  & 0.6368\\
angles in $[0^{\circ},180^{\circ}]$ & Tensor-1 & 52801 & {\bf 4.46} & {\bf 0.79} & 17.19 & 0.6560 \\
noise $1\%$ & Tensor-2 & {\bf 15676} & 17.39 & 1.84 & {\bf 16.82} & {\bf 0.6688} \\
\hline
$N_{\mathrm{p}}=50$ & Matrix & 48873 & 14.4575 & 9.43 & 22.77 & 0.5695 \\
angles in $[0^{\circ},120^{\circ}]$ &  Tensor-1 & 61106 & {\bf 9.08} & {\bf 0.98} & 22.80 &  0.5818 \\
noise $1\%$ & Tensor-2 & {\bf 16177} &  23.81 & 2.07 &  {\bf 22.49} & {\bf 0.5883} \\
\hline
$N_{\mathrm{p}}=25$ & Matrix & 45775 & 100 &  5.91 & 25.46 & 0.4536 \\
angles in $[0^{\circ},120^{\circ}]$ &  Tensor-1 & 59347 & {\bf 26.00} & {\bf 0.73} & 25.85 & 0.4544 \\
noise $1\%$ & Tensor-2 & {\bf 17053} & 27.49 & 2.29 & {\bf 25.33}  & {\bf 0.4676} \\
\hline
$N_{\mathrm{p}}=50$ & Matrix & 110322 & 50.17 & 8.02 & 22.05 &  0.4910\\
angles in $[0^{\circ},180^{\circ}]$ &  Tensor-1 & 40695 & {\bf 8.97} & {\bf 0.74} & 21.84 & 0.4846 \\
noise $5\%$ & Tensor-2 & {\bf 10392} &  14.64 & 1.72 & {\bf 21.81} & {\bf 0.5107} \\
\hline
$N_{\mathrm{p}}=25$ & Matrix & 72139 & 45.51 & 6.29 & 24.69 & 0.3768 \\
angles in $0^{\circ},180^{\circ}]$ &  Tensor-1 & 37072  & {\bf 8.60} & {\bf 0.64} & 25.12 & 0.3738 \\
noise $5\%$ & Tensor-2 &  {\bf 9076} & 13.28 & 2.4829 &  {\bf 24.67} & {\bf 0.4041}  \\
 \hline
\end{tabular}
\end{table}

Table \ref{tab:2} lists the corresponding relative error, SSIM, density,
and compressibility together with the iteration number.
Comparison of Tables \ref{tab:1} and \ref{tab:2} reveal the same pattern.
Algorithm \ref{alg1} converges faster when imposing the combined regularization term
$\|\mathcal{C}\|_{\mathrm{sum}} + \|C\|_*$, and this choice also slightly improves
the reconstruction in all scenarios.
However, enforcing only the sparsity prior
$\|\mathcal{C}\|_{\mathrm{sum}}$
significantly reduce the representation redundancy, leading
to a very sparse representation comparing to the matrix formulation.
In the scenario with 50 projections and $1\%$ noise, where the regularization and
perturbation errors are less dominating,
the improvement in reconstructions by the tensor algorithm\,---\,compared to
the matrix formulation\,---\,is more pronounced.
Overall, we recommend the use of $\|\mathcal{C}\|_{\mathrm{sum}} + \|C\|_*$ which
leads to the faster algorithm.

\subsubsection{A Larger Test Problem}
\label{sec:expr3}

\begin{figure}%[htp]
\centering
\subfigure{ \includegraphics[height=0.3\linewidth]{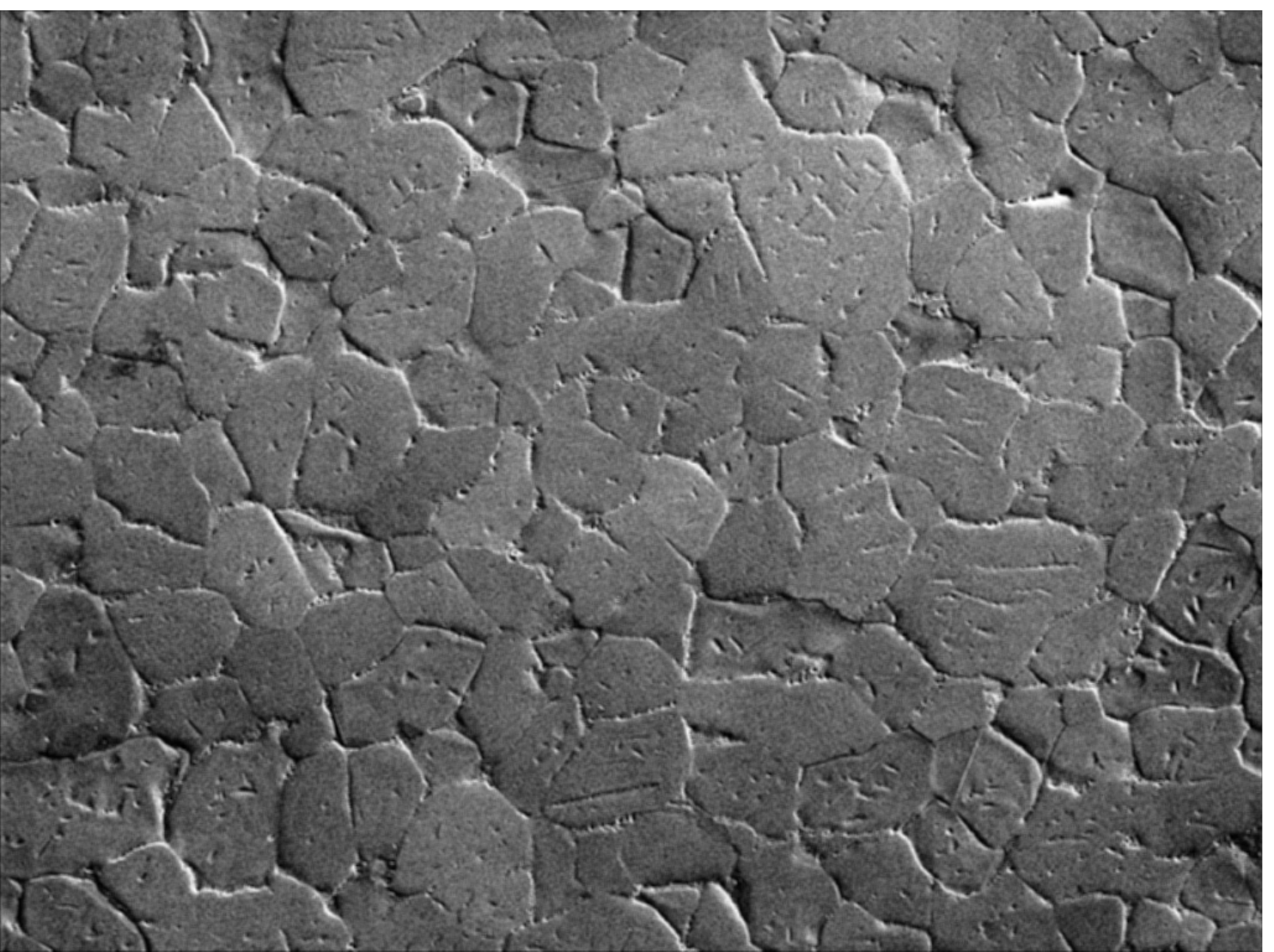}}
\subfigure{ \includegraphics[height=0.3\linewidth]{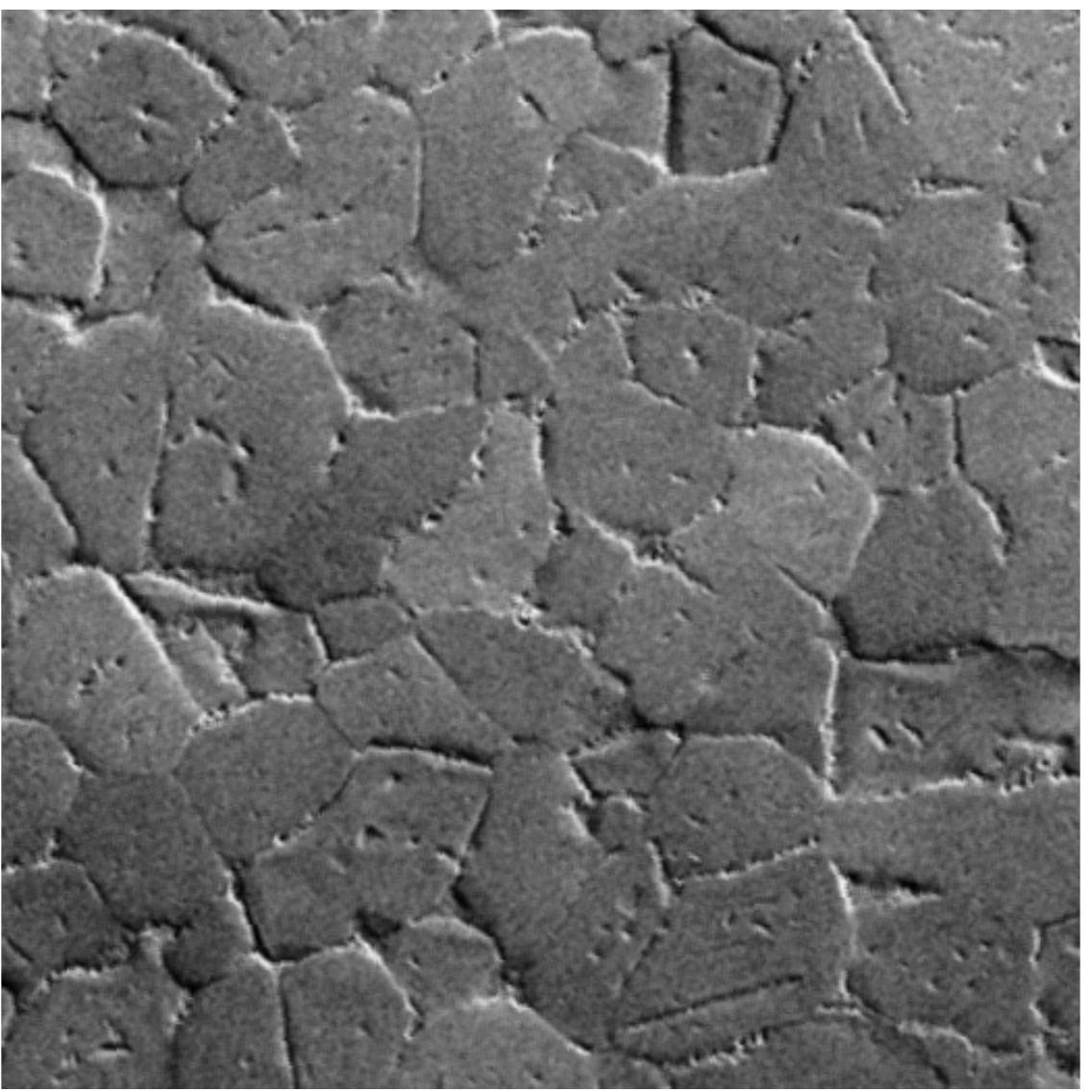}}
\caption{Left:\ the high-resolution image of zirconium grains used to
generate the training patches. Right:\ the exact image of size $520 \times 520$.}
\label{fig:grain}
\end{figure}

Tomography is a common tool in materials science to study the structure of grains
in polycrystalline materials such as metals.
The grain structure is sometimes known a priori in the form of training images.
As test image in this experiment we use a high-resolution image of zirconium grains
(produced by a scanning electron microscope)
of dimension $760 \times 1020$ shown in Fig.~\ref{fig:grain}.

Training patches of size $10 \times 10$ are again extracted from the high-resolution
image to learn matrix and tensor dictionaries size $100 \times 300$ and
$10 \times 300 \times 10$, respectively.
To avoid committing inverse crime, we first rotate the high-resolution image
and then extract the exact image of dimensions
$520\times 520$\,---\,also shown in Fig.~\ref{fig:grain}.

We use a parallel-beam setting with $N_{\mathrm{p}}=50$ projection angles
in $[0^{\circ},180^{\circ}]$ and $N_{\mathrm{r}} = 707$ rays per
projection, and again the matrix is computed by means of the function \texttt{paralleltomo}
from \textsc{AIR Tools} \cite{Hansen}.
We added 1\% white Gaussian noise to the clean data.
This problem is highly underdetermined with $m=36750$ measurements and
$n=270400$ unknowns.

\begin{figure}%[htp]
\centering
\subfigure[Matrix formulation]{ \includegraphics[width=0.3\linewidth]{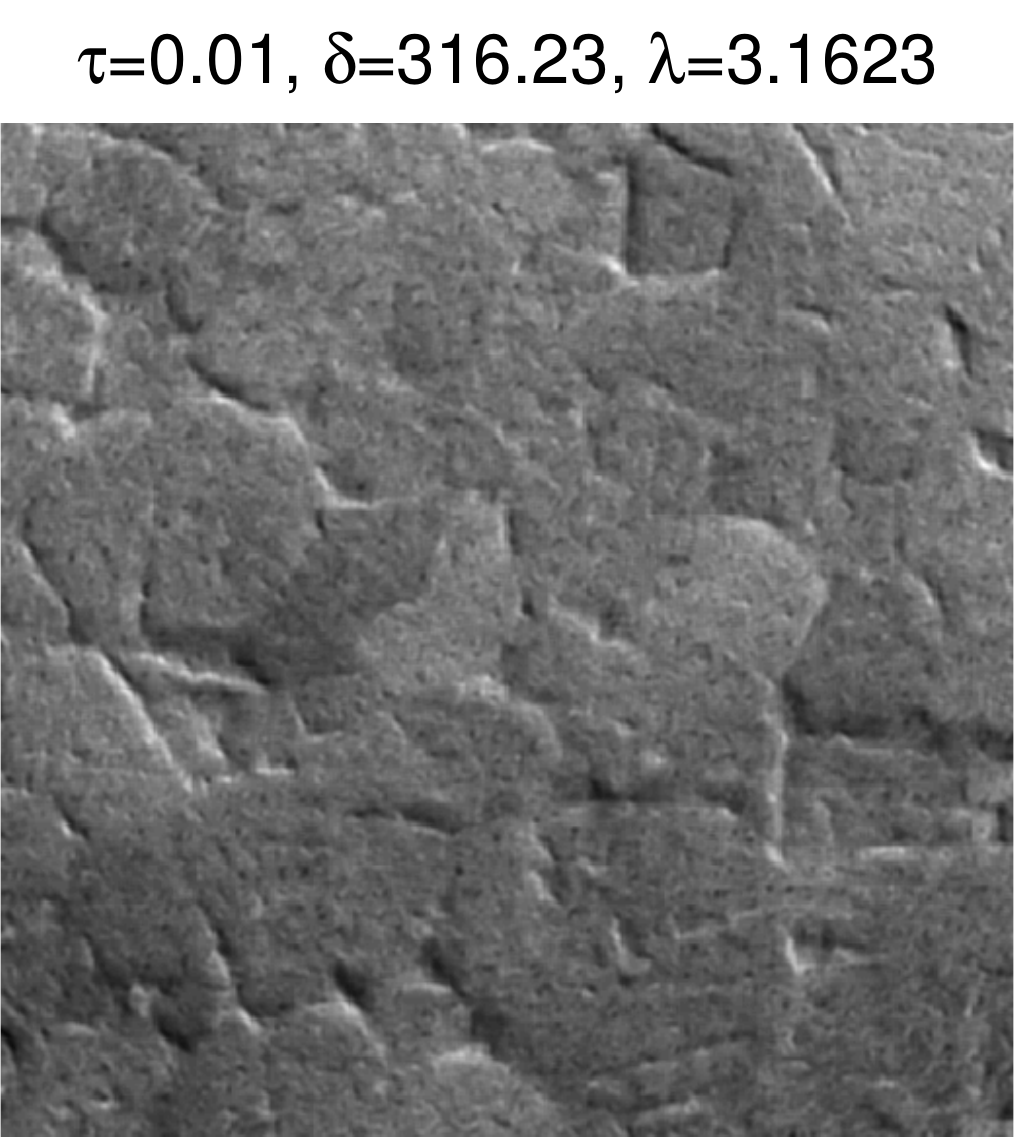}}
\subfigure[Tensor:\ $\|\mathcal{C}\|_{\mathrm{sum}}$]{ \includegraphics[width=0.3\linewidth]{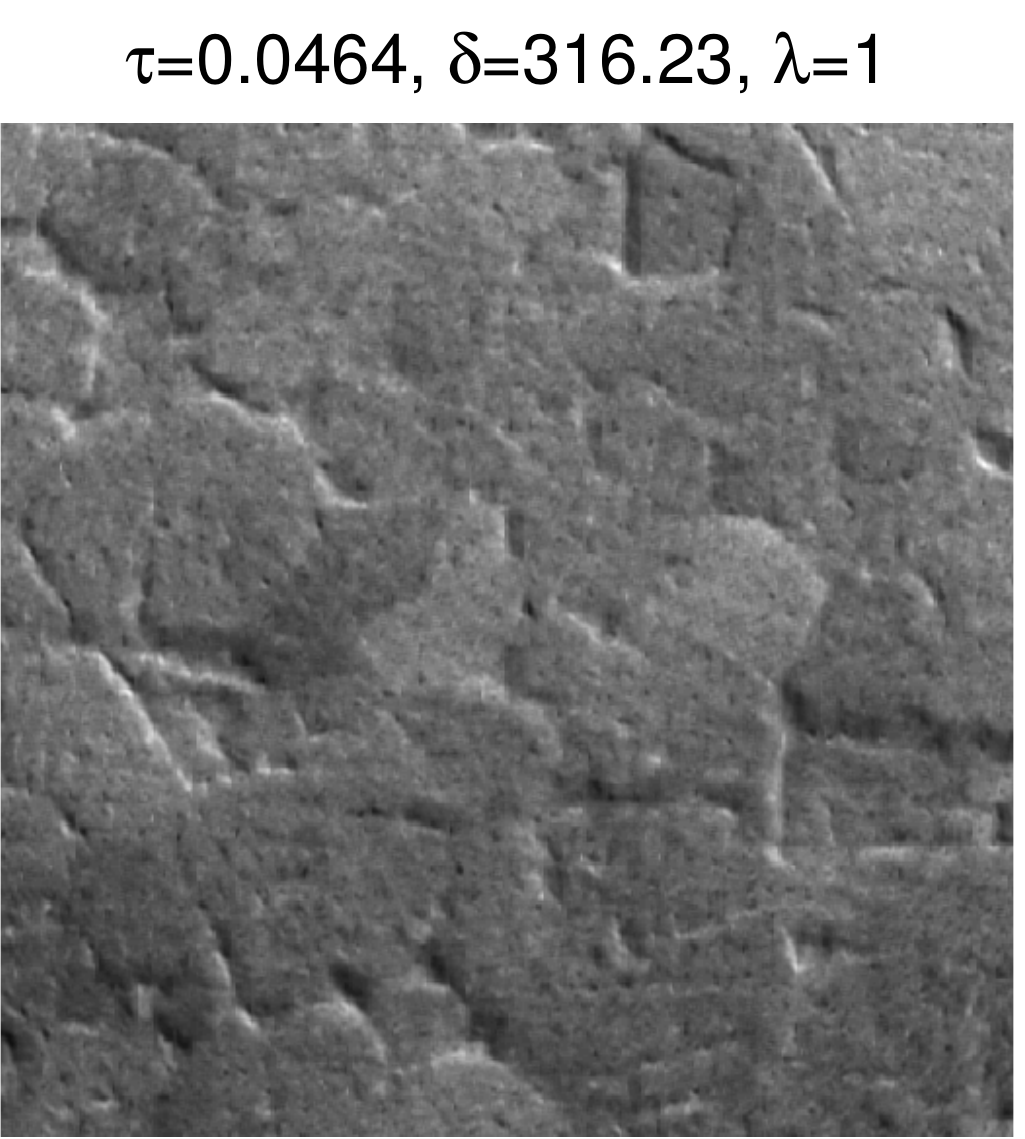}}
\subfigure[Tensor:\ $\|\mathcal{C}\|_{\mathrm{sum}}+\| C \|_{*}$]{ \includegraphics[width=0.3\linewidth]{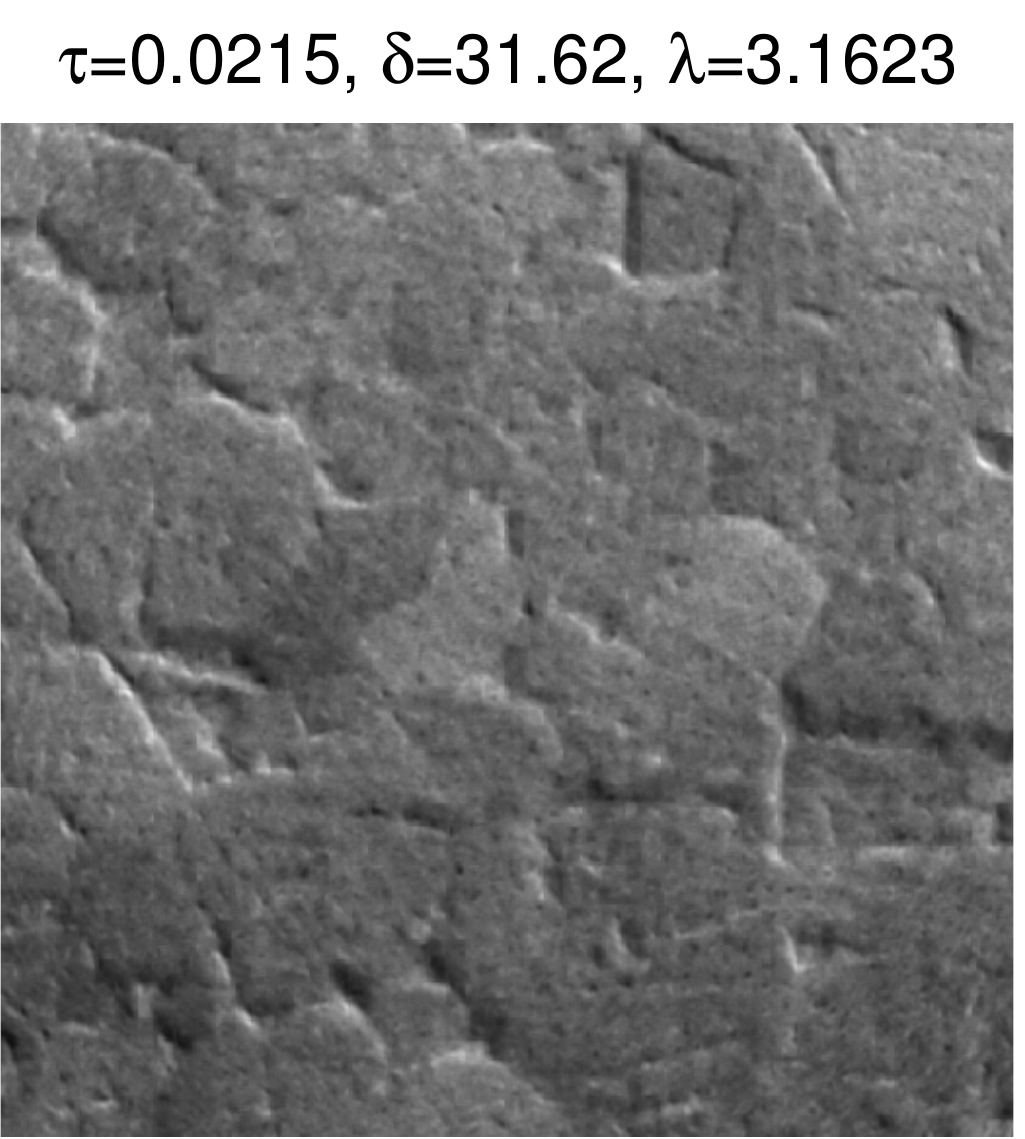}}
\caption{Reconstructions for the realistic test problem, computed with the matrix
formulation (a) and the tensor formulation (b) + (c).}
\label{fig:ImRes3}
\end{figure}

\begin{table}[htp]
\caption{Comparison of reconstruction in the realistic test problem, using the
matrix and tensor formulations.
The bold numbers indicate the lowest iteration number, density, and compression,
and highest SSIM.}
\label{tab:3}
\begin{tabular}{llllll}
\hline\noalign{\smallskip}
Method & Itr.$\#$ & Density$\%$ & Compr.\% & RE\% & SSIM  \\
\hline
Matrix formulation \cite{Soltani} & 73961 & 48.61 & 6.86 & 14.90 & 0.4887 \\
Tensor alg., $\|\mathcal{C}\|_{\mathrm{sum}}$ reg. & 74310 &  {\bf 33.18 } & {\bf 0.76} & 15.23 & 0.4793  \\
Tensor alg., $\|\mathcal{C}\|_{\mathrm{sum}}+\| C \|_*$ reg. & {\bf 24396} &  38.78 & 3.17 & {\bf 14.80} & {\bf 0.5035} \\
\hline
\end{tabular}
\end{table}

Figure \ref{fig:ImRes3} shows the reconstructed images with the matrix and tensor
formulations.
All regularization parameters are chosen empirically to give the smallest
reconstruction errors.
All three reconstructions are similar, since the reconstruction errors are dominated
by the error coming from the regularization of the noisy data.
More data are given in Table \ref{tab:3}.
Imposing the sparsity prior $\|\mathcal{C}\|_{\mathrm{sum}}$ in the tensor formulation
produces the sparsest representation.
The solution is computed in fewer iterations with the
$\|\mathcal{C}\|_{\mathrm{sum}}+\|C\|_*$ regularization term while the
reconstruction has a negligible improvement in terms of RE and SSIM.
We conclude that our tensor algorithm is also well suited for more
realistic tomographic problems.

\section{Conclusion}

We presented the problem of dictionary learning
in a tensor formulation and focused on solving the tomographic image reconstruction
in the context of a t-product tensor-tensor factorization.
We posed the
tensor dictionary learning problem as a non-negative sparse tensor factorization
problem, and we computed a regularized nonnegative reconstruction in the
tensor-space defined by the t-product.
We also gave an algorithm based on the alternating direction method of multipliers
(ADMM) for solving the tensor dictionary learning problem and,
using the tensor dictionary, we formulated a convex optimization problem to
recover the solution's coefficients in the expansion under the t-product.

We presented numerical experiments on the properties of the representation
in the learned tensor dictionary in the context of tomographic reconstruction.
The dictionary-based reconstruction quality is superior to well know
classical regularization schemes, e.g., filtered back projection and total
variation, and
the solution representation in terms of the tensor dictionary is more sparse compared
to a similar matrix dictionary representations \cite{Soltani}.
%Our experiments suggest that additional prior constraints improve representation
%and quality of the reconstruction.
In future work the authors would like to further study the tensor dictionary
representation property using other products from the family of tensor-tensor
products introduced, e.g., in \cite{Kernfeld}.

\section*{Acknowledgments}
The authors acknowledge collaboration with Martin S. Andersen from DTU Compute.
We would like to thank professor Samuli Siltanen from University of Helsinki for
providing the high-resolution image of the peppers
and Dr. Hamidreza Abdolvand from University of Oxford for providing the zirconium image.

\section*{Appendix A: Reconstruction Solution Via TFOCS}

The reconstruction problem \eqref{eq:trec} is a convex, but
$\| \mathcal{C} \|_{\mathrm{sum}}$ and $\|C\|_*$ are not differentiable
which rules out conventional smooth optimization techniques.
The TFOCS software \cite{Becker} provides a general framework for solving
convex optimization problems, and the
core of the method computes the solution to a standard problem of the form
  \begin{equation}
  \label{eq:TFOCS}
    \mbox{minimize} \quad l(A(x)-b)+h(x) ,
  \end{equation}
where the functions $l$ and $h$ are convex, $A$ is a linear operator, and $b$ is a vector;
moreover $l$ is smooth and $h$ is non-smooth.

To solve problem \eqref{eq:trec} by TFOCS, it is reformulated as a constrained
linear least squares problem:
  \begin{equation}
  \label{eq:recreform}
    \min_{\mathcal{C}} \frac{1}{2} \left\| \begin{pmatrix}
    \nicefrac{1}{\sqrt{m}} \, A \\[1mm] \nicefrac{\delta}{c} \, L \end{pmatrix}
    \Pi \mathtt{vec}(\mathcal{D}*\mathcal{C})
    - \begin{pmatrix} b \\ 0 \end{pmatrix} \right\|_2^2 +
    \mu \, \varphi_{\nu}(\mathcal{C})
    \qquad \text{s.t.}  \qquad  \mathcal{C} \geq 0,
\end{equation}
where $c = \sqrt{2( M(M/p-1)+N(N/r-1) )}$.
Referring to \eqref{eq:TFOCS}, $l(\cdot )$ is the squared 2-norm residual and
$h(\cdot ) = \mu \, \varphi_{\nu}(\cdot)$.

The methods used in TFOCS require computation of the proximity operators
of the non-smooth function $h$.
The proximity operator of a convex function is a natural extension of the notion of a
projection operator onto a convex set \cite{Combettes}.

Let $f = \|\mathcal{C}\|_{\mathrm{sum}} = \|C\|_{\mathrm{sum}}$ and
$g=\|C\|_*$ be defined on the set of real-valued matrices
and note that $\mathrm{dom} f \cap \mathrm{dom}\, g \neq \emptyset$.
For $Z \in \mathbb{R}^{m\times n}$ consider the minimization problem
  \begin{equation}
  \label{eq:prox}
    \mbox{minimize}_X \ f(X)+g(X) + \nicefrac{1}{2} \|X-Z\|_{\mathrm{F}}^2
  \end{equation}
whose unique solution is $X = \mathrm{prox}_{f+g}(Z)$.
While the prox operators for $\|C\|_{\mathrm{sum}}$ and $\|C\|_*$ are easily
computed, the prox operator of the sum of two functions is intractable.
Although the TFOCS library includes implementations of a variety of prox
operators\,---\,including norms and indicator functions of many common convex
sets\,---\,implementation of prox operators of the form
$\mathrm{prox}_{f+g}(\cdot)$ is left out.
Hence we compute the prox operator for
$\|\cdot\|_{\mathrm{sum}}+\|\cdot\|_*$ iteratively using a
Dykstra-like proximal algorithm \cite{Combettes},
where prox operators of $\|\cdot\|_{\mathrm{sum}}$ and $\|\cdot\|_*$
are consecutively computed in an iterative scheme.

Let $\tau=\mu/q \geq 0$.
For $f(X)=\tau\|X\|_{\mathrm{sum}}$ and $X \geq 0$, $\mathrm{prox}_f$ is the
one-sided elementwise shrinkage operator
  \[
    \mathrm{prox}_f(X)_{i,j} = \begin{cases}
      0 , &  X_{i,j} \geq \tau \\
      X_{i,j}-\tau , & |X_{i,j}| \leq \tau \\
      0, &  X_{i,j} \leq -\tau \\
    \end{cases}
  \]
The proximity operator of $g(X)=\tau\|X\|_*$ has an analytical expression
via the singular value shrinkage (soft threshold) operator
  \[
    \mathrm{prox}_g (X) = U \mathrm{diag}(\sigma_i-\tau) \, V^T,
  \]
where $X=U \Sigma \, V^T$ is the singular value decomposition of $X$~\cite{Cai}.
The computation of $\tau\| C \|_*$ can be done very efficiently since
$C$ is $sq \times r$ with $r \ll sq$.

The iterative algorithm which computes an approximate solution to
$\mathrm{prox}_{f+g}$ % the  problem \eqref{eq:prox}
is given in Algorithm \ref{alg2}.
Every sequence $X_k$ generated by Algorithm \ref{alg2} converges to the
unique solution $\mathrm{prox}_{f+g}$ of problem \eqref{eq:prox} \cite{Combettes}.

\begin{algorithm}
 \caption{Dykstra-Like Proximal Algorithm}
\label{alg2}
\begin{algorithmic}[h]
 \State {\bf Input}: The matrix $Z$
 \State {\bf Output}: $\mathrm{prox}_{f+g}(Z)$
 \State {\bf Initialization}: Set $X_1=Z$ and set $P_1$ and $Q_1$ to zero matrices
   of appropriate sizes.
 \For{$k~=~1,2,\ldots$} \State{$Y_k~ = ~\mathrm{prox}_g (X_k + P_k)$
  \State $P_{k+1}~ = ~X_k + P_k - Y_k$
  \State $X_{k+1}~ = ~\mathrm{prox}_f(Y_k + Q_k)$
  \State $Q_{k+1}~=~ Y_k + Q_k - X_{k+1}$
  \If {$\| Y_k - X_{k+1} \|_{\mathrm{F}} < 10^{-3}$}
  \State Exit
  \EndIf
 \EndFor}
\end{algorithmic}
\end{algorithm}

\end{document}